\PassOptionsToPackage{table}{xcolor}
\documentclass{article}

\usepackage{iclr2027_conference,times}

\usepackage{amsmath,amsfonts,bm}

\def\eqref#1{equation~\ref{#1}}

\def\1{\bm{1}}

\DeclareMathAlphabet{\mathsfit}{\encodingdefault}{\sfdefault}{m}{sl}
\SetMathAlphabet{\mathsfit}{bold}{\encodingdefault}{\sfdefault}{bx}{n}

\usepackage{hyperref}
\usepackage{url}

\usepackage{amsmath,amssymb,amsfonts}
\usepackage{amsthm}
\usepackage{mathtools}
\usepackage{nicefrac}

\usepackage{booktabs}
\usepackage{multirow}
\usepackage{array}
\usepackage{makecell}
\usepackage{tabularx}
\usepackage{longtable}
\usepackage{arydshln}

\usepackage{graphicx}
\usepackage{wrapfig}
\usepackage{subcaption}
\usepackage{float}

\usepackage{algorithm}
\usepackage{algpseudocode}

\usepackage{microtype}
\usepackage{xcolor}
\usepackage{enumitem}
\usepackage{multicol}
\usepackage{ulem}
\usepackage{CJKutf8}

\usepackage{tcolorbox}
\tcbuselibrary{breakable}

\definecolor{tablegray}{gray}{0.93}
\definecolor{bestbg}{RGB}{211,235,245}
\definecolor{secondbg}{RGB}{250,226,194}

\newcommand{\best}[1]{%
  \begingroup
  \setlength{\fboxsep}{1.0pt}%
  \colorbox{bestbg}{\strut\textbf{#1}}%
  \endgroup
}

\newcommand{\second}[1]{%
  \begingroup
  \setlength{\fboxsep}{1.0pt}%
  \colorbox{secondbg}{\strut\textbf{#1}}%
  \endgroup
}

\newcommand{\bestcell}[1]{\cellcolor{bestbg}\textbf{#1}}
\newcommand{\secondcell}[1]{\cellcolor{secondbg}\textbf{#1}}

\setlist[itemize]{noitemsep,leftmargin=*,topsep=0pt}

\title{Breaking Babel: A Self-Evolving Multi-Agent System for Long-Form Subtitle Translation}

\author{
\begin{tabular}{@{}l@{}}
Haibo Jin$^{1}$\thanks{Work done during an internship at Amazon.} \quad
Xinjie Li$^{2}$\thanks{Project Lead.} \quad
Najmeh Sadoughi$^{2}$ \quad
Yang Liu$^{2}$ \\
Yibo Wang$^{2}$ \quad
Zhu Liu$^{2}$ \quad
Yuzong Liu$^{2}$ \\[3pt]
{\small $^{1}$University of Illinois Urbana-Champaign, USA} \\
{\small $^{2}$Amazon, USA}
\end{tabular}
}

\iclrfinalcopy

\begin{document}

\maketitle

\fancyhead{}
\lhead{Preprint.}

\begin{abstract}
Long-form subtitle translation presents challenges beyond those of conventional machine translation: a single episode may contain hundreds of sentences whose meanings depend on long-range discourse and cultural context spanning the entire episode or even series, while translation quality also requires maintaining consistent terminology and style throughout the series. Existing approaches remain limited. Single-LLM methods operate at the sentence level, lacking long-context understanding and consistent terminology across episodes. Multi-agent methods often rely on static workflows that fail to adapt to scene complexity. In addition, both paradigms ignore the production context, e.g., genre. We propose \textbf{SMART}, a \textbf{S}elf-evolving \textbf{M}ulti-\textbf{A}gent system for long-fo\textbf{R}m subtitle \textbf{T}ranslation. SMART operates in two stages: during \emph{test-time training}, it builds a persistent series-level memory and iteratively translates a subset of sentences through a dynamic graph router and a Mixture-of-Agents translation layer, equipped with tool-calling modules for terminology verification, subtitle constraint validation, and contextual retrieval; a judge-refiner loop scores each candidate translation and back-propagates textual critiques that refine agent prompts and the routing policy, without retraining the underlying LLMs. During \emph{test-time inference}, this evolved configuration translates the remaining sentences of the series. To evaluate long-form subtitle translation at scale, we introduce \textbf{Subtitle Arena}, covering 14 genres, 2--198 episodes per series, production years from 1959 to 2023, and 15 target locales, together with \textbf{SubMQM}, a subtitle-adapted MQM framework with seven dimensions and 19 error categories. SMART achieves the best overall MQM score in all 15 Subtitle Arena directions, reducing the average penalty by 6.9\% over the strongest competing agent system. On a public benchmark, MuSC, SMART obtains the best model result across all 4 language pairs. SMART also achieves the best result in human evaluation with an overall score of 4.50/5. 

\textbf{Epigraph:} \emph{Babel scattered one tongue into many; this work gathers them back.}
\end{abstract}

\section{Introduction}
Large language models (LLMs) have substantially advanced machine translation~\citep{singh2025openai, yang2025qwen3, anthropic2025claudesonnet48} in general-domain settings~\citep{xu2024paradigm, xu2025x}.
Long-form subtitle translation, however, remains more challenging than translating isolated sentences. Subtitles often contain hundreds of sentences~\citep{karakanta2020must}, while their interpretation often depends on discourse spanning speakers, scenes, and episodes, together with visual and cultural context~\citep{wu2025perhaps}. At the same time, translations must preserve terminology and narrative coherence while satisfying strict presentation constraints such as line length and reading speed~\citep{papi2023direct, karakanta2020must}.

Existing approaches mainly follow two paradigms. One line adapts a single LLM through subtitle-specific fine-tuning~\citep{cui2026utterance} or context-aware prompting~\citep{pramodya2025translating} by incorporating neighboring dialogue, genre, or plot summaries. Another line adopts LLM-based multi-agent systems, where specialized roles collaboratively generate and refine translations~\citep{wu-etal-2024-transagents}, sometimes augmented with multimodal context~\citep{lu2025vidove}. Despite their effectiveness, these approaches remain limited when translation extends over long-form narrative content.

In particular, we identify three challenges. \textbf{First, long-range understanding and consistency}. Single LLM methods~\citep{anthropic2025claudesonnet46} primarily reason over local sentences and lack persistent series-level context, causing recurring entities, terminology, and referring expressions to drift across scenes and episodes. \textbf{Second, static coordination workflow.} Multi-agent systems often fix the roles and refinement procedures in advance, although translation difficulty varies considerably across scenes. Simple dialogue may require little coordination, while ambiguous references, culturally grounded expressions, or domain-specific scenes may benefit from additional specialists and verification. \textbf{Third, contextual adaptation.} Appropriate wording depends strongly on production context, including genre, historical setting, and cultural background, yet current systems have limited ability to retrieve and incorporate such signals dynamically.

To address these challenges, we introduce \textbf{SMART}, a \textbf{S}elf-evolving \textbf{M}ulti-\textbf{A}gent system for long-fo\textbf{R}m subtitle \textbf{T}ranslation. SMART contains two stages: \textit{test-time training} and \textit{test-time inference}. During \textit{test-time training}, SMART translates a subset of a series while evolving its agent prompts, routing policy, and persistent series-level memory. A dynamic graph router selects specialized translators and tools for each sentence, while a Mixture-of-Agents layer generates complementary translation candidates. Tool-calling modules provide terminology verification, subtitle-constraint checking, and contextual retrieval for signals such as genre and production background. A judge-refiner loop evaluates candidate translations and converts observed errors into textual feedback that updates the prompts and routing policy without modifying the underlying LLM parameters. During \textit{test-time inference}, the evolved configuration is frozen and applied to the remaining content, while the series memory continues to accumulate.

To evaluate SMART, we construct \textbf{Subtitle Arena} to address limitations of existing subtitle benchmarks, including noisy or misaligned sentences~\citep{cui2026utterance,tiedemann-luo-2026-opensubtitles2024} and limited high-quality multilingual pairing~\citep{refine-ai}. Beyond sentence-level translation quality, Subtitle Arena is designed to evaluate three properties central to long-form subtitle translation: \emph{(i)} discourse-aware translation that uses long-range context to resolve the current sentence, \emph{(ii)} consistent terminology, character references, and style across extended narrative horizons, and \emph{(iii)} compliance with subtitle-specific display constraints. It contains TV series spanning 14 genres, production years from 1959 to 2023, and 2--198 episodes per series across 15 target locales. We further introduce \textbf{SubMQM}, a subtitle-adapted automatic MQM (Multidimensional Quality Metrics) framework covering seven dimensions and 19 fine-grained error categories. Our main contributions are as follows:
\begin{itemize}
\item We propose \textbf{SMART}, a self-evolving multi-agent framework for long-form subtitle translation with a test-time training stage that jointly adapts agent prompts and routing policies, while maintaining persistent series-level memory and dynamically invoking contextual and verification tools, and a test-time inference stage that translates the rest of the content.

\item We introduce \textbf{Subtitle Arena}, a long-form subtitle translation benchmark spanning diverse genres, production eras, episodes, and language pairs, with explicit support for evaluating cross-episode consistency and contextual adaptation. To complement subtitle translation metrics, we further adopt a subtitle-adapted MQM framework, \textbf{SubMQM}, for fine-grained analysis of semantic, linguistic, contextual, and subtitle-specific technical errors.

\item Extensive evaluations show that SMART achieves the best Overall MQM score in all 15 Subtitle Arena directions with a 6.9\% lower average penalty than the strongest agent baseline, and the best model result across all 4 MuSC language pairs. SMART also achieves the best result in human evaluation at 4.50/5.

\end{itemize}

\section{Related Work}
\textbf{Multi-Agent Systems and Translation Agents.}
Multi-agent LLM systems decompose complex tasks across specialized agents~\citep{li2023camel, MAS-9423979, MAS-ijcai2024p890, MAS-cheng2024exploringlargelanguagemodel}, and have increasingly been adopted for machine translation.
Existing systems mainly use role-based collaboration for generation, evaluation, and refinement~\citep{he2024maps, wu-etal-2024-transagents, feng2025tear, wang2025drt, li2025tactic, wang2025maats}, or introduce persistent context for document and subtitle translation~\citep{MAS-mt-wang2025deltaonlinedocumentleveltranslation, lu2025vidove}.
ALPO~\citep{cui2026utterance} instead improves expressive subtitle translation through preference optimization.
Despite these advances, most methods operate with predefined workflows or fixed agent behaviors.

\textbf{Self-Evolving Agents and Prompt Optimization.}
Self-evolving agents improve their workflows~\citep{hu2025automateddesignagenticsystems, zhang2025aflowautomatingagenticworkflow, wang2025evoagentxautomatedframeworkevolving, zhang2025evoflowevolvingdiverseagentic}, reusable skills and experience~\citep{wang2023voyageropenendedembodiedagent, zhao2024expelllmagentsexperiential}, or persistent memory~\citep{suzgun2025dynamiccheatsheettesttimelearning}.
Related prompt-optimization methods iteratively improve instructions using search, reflection, or textual feedback~\citep{zhou2022large, pryzant2023automatic, yuksekgonul2024textgrad, agrawal2025gepa}.
However, these approaches typically optimize performance on future tasks or fixed objectives, rather than adapting during a single long-form task while preserving previously acquired behavior.

\textbf{LLM-as-a-Judge.}
LLMs are increasingly used to evaluate model outputs through scoring, ranking, and natural-language critiques~\citep{zheng2023judging}.
For machine translation, LLM-based evaluators such as GEMBA-MQM~\citep{kocmi2023gemba} and AutoMQM~\citep{fernandes2023devil} provide structured feedback over translation errors and their severity. Such methods primarily use the judge for evaluation or local refinement.

\textbf{Key Differences.}
SMART differs from prior work in three respects. Unlike translation agents with fixed workflows, SMART \emph{evolves at test time}, adapting agent instructions, routing, and contextual knowledge. Unlike general self-evolving agents, SMART targets the \emph{long-form translation task}, improving future sentences while preserving terminology, style, and prior translation competence. Unlike conventional prompt optimization, SMART treats prompts and agent behaviors as \emph{persistent system state} updated through translation-specific feedback.

\section{Methodology}
\subsection{Problem Formulation}

Let an episode-level subtitle be $\mathcal{D}=\{d_i\}_{i=1}^{N}$, where each
sentence $d_i=(x_i,\tau_i,b_i)$ contains source text $x_i$, a timestamp interval
$\tau_i$, and display constraints $b_i$ such as characters per line and reading
speed. Given a target locale $L$, the goal is to generate
$\mathcal{Y}=\{y_i\}_{i=1}^{N}$ that is locally faithful while remaining
globally coherent across episodes of the same series. The translation must
preserve meaning, character narrative voice, terminology, and subtitle-specific display
constraints over long contexts.

At training step $t$, SMART maintains the system state
\begin{equation}
    \mathcal{S}_t=(\mathcal{R},\Pi_t,\rho_t,M_t),
\end{equation}
where $\mathcal{R}$ is a fixed pool of specialized translator roles,
$\Pi_t=\{\pi_{r,t}\}_{r\in\mathcal{R}}$ denotes their prompts, $\rho_t$ is a
routing policy that selects active translators, and $M_t$ is a persistent
series-level memory. Unlike fixed multi-agent pipelines, SMART adapts both how
translators are instructed and which translators are invoked from translation
feedback, without updating the underlying LLM parameters.

\subsection{System Overview}

\begin{figure*}[t]
\centering
\includegraphics[width=\textwidth]{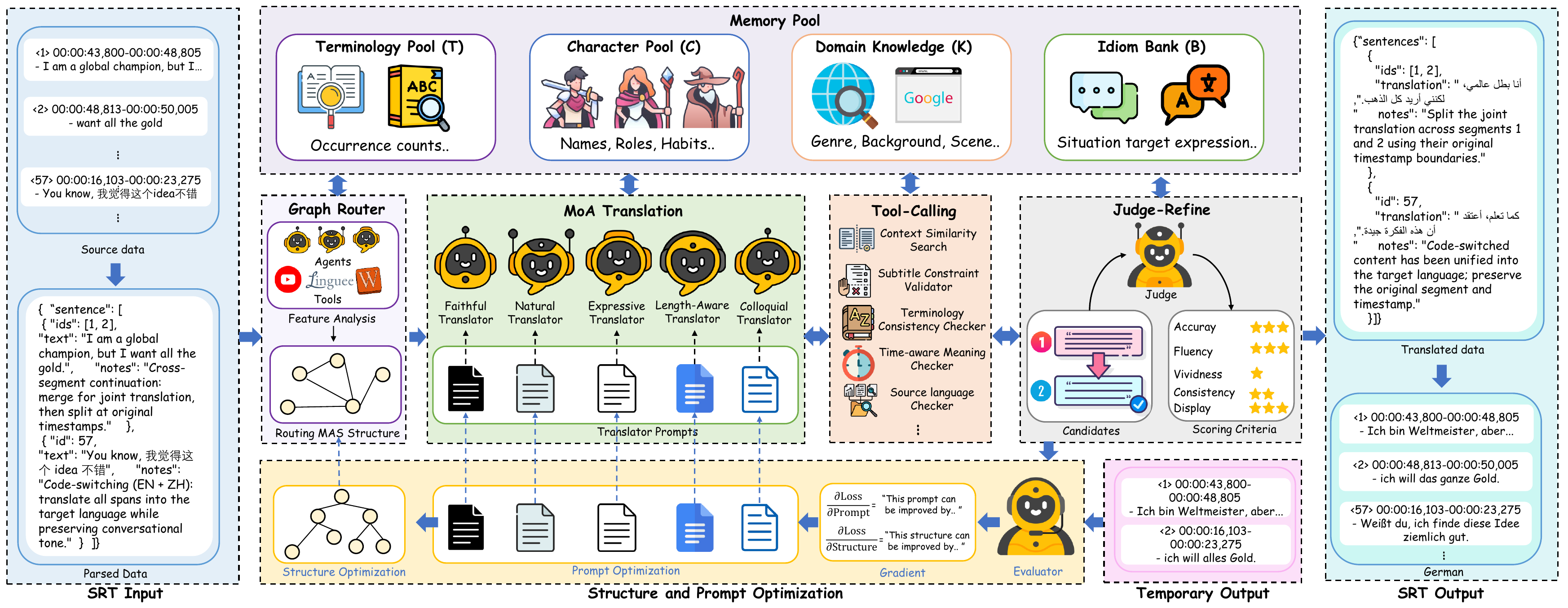}
\vspace{-18pt}
\caption{
Overview of \textbf{SMART}. Given source data in SRT style, SMART parses it into
sentences. For each sentence, a learnable router (\textbf{Graph Router})
activates a subset of translators (\textbf{MoA Translation}); candidates may
invoke tools (\textbf{Tool-Calling}) and are selected and refined by a
judge--refiner loop (\textbf{Judge-Refine}). A persistent \textbf{Memory Pool}
carries terminology, character information, domain knowledge, and idiomatic
expressions across episodes. During test-time training, translation feedback
updates translator prompts and routing.
}
\vspace{-10pt}
\label{pipeline}
\end{figure*}

Fig.~\ref{pipeline} summarizes SMART through three interacting mechanisms.
First, a \textbf{persistent memory pool} maintains series-level information
that should survive beyond individual model calls, including terminology,
character profiles, domain knowledge, and target-language idioms. Second, an
\textbf{adaptive translation graph} routes each subtitle sentence to a subset
of specialized translators, which may invoke tools for contextual retrieval,
terminology control, external knowledge, and subtitle validation. Third, a
\textbf{judge--refiner loop} evaluates the resulting candidates, selects the
most promising translation, and refines it before the accepted result is
written back to memory.

For \textbf{test-time training and inference}, we split the episodes of each series into
$\mathcal{D}_{\mathrm{tr}}$ and a held-out set $\mathcal{D}_{\mathrm{inf}}$
using a $3{:}7$ ratio. Translation feedback from $\mathcal{D}_{\mathrm{tr}}$
is used during training to improve translator prompts and routing decisions. These components
are then frozen for $\mathcal{D}_{\mathrm{inf}}$ during inference, while memory continues to
grow as translation proceeds across episodes.
Appendix~\ref{app:test-time-training-loop} details the optimization protocol,
and Appendix~\ref{app:end-to-end-example} provides an end-to-end example.

\subsection{Persistent Series Memory}
\label{sec:series-memory}

Long-form subtitle translation contains information that is sparse but
persistent: a character, expression, or domain-specific term may be introduced
in one episode and recur much later. SMART therefore maintains a series-level
memory
\begin{equation}
    M=(T,C,K,B),
\end{equation}
where $T$ is a terminology pool, $C$ stores character profiles, $K$ stores
domain and scene knowledge, and $B$ is an idiom bank. Before translating
episode $e$, SMART loads the state $M^{(e)}$ accumulated from earlier episodes.
Newly confirmed information is merged back after translation:
\begin{equation}
    M^{(e+1)} = M^{(e)} \cup \Delta M^{(e)}.
\end{equation}
All translators and tools can query this memory, allowing recurring entities
and prior translation decisions to be retrieved rather than independently
re-derived. The memory therefore provides an explicit long-range state beyond
the context window of any individual model call. Appendix~\ref{app:series-memory-example}
provides a concrete cross-episode example.

SMART additionally constructs a lightweight content profile containing the
series genre, setting, principal characters, and domain background. Ambiguous
jargon, cultural references, and slang may trigger external retrieval. The
retrieved evidence is interpreted in the context of the current scene before
being stored, separating literal external knowledge from the register and
wording appropriate for subtitle translation. Further details are provided in
Appendix~\ref{app:method-details}.

\subsection{Adaptive Multi-Agent Translation}
\label{sec:adaptive-translation}

\textbf{Dynamic routing.}
Rather than invoking every translator for every sentence, SMART selects an
active set $\mathcal{A}_i\subseteq\mathcal{R}$ according to
\begin{equation}
    \mathcal{A}_i = \rho_t\!\left(\phi(d_i,M_t)\right),
\end{equation}
where $\phi(\cdot)$ summarizes properties relevant to translation, including
sentence length, scene tone, and lexical cues such as slang or profanity. The
translator pool contains complementary roles emphasizing semantic
faithfulness, naturalness, expressiveness, subtitle-length control, and
colloquial rendering. The router can therefore allocate additional translation
capacity when a sentence benefits from it instead of applying a fixed topology
to every input.

\textbf{Mixture-of-agents translation with tools.}
Each active translator $r\in\mathcal{A}_i$ produces a candidate under its
current prompt $\pi_{r,t}$:
\begin{equation}
    y_i^{(r)} = r\!\left(d_i,\pi_{r,t},M_t,\mathcal{T}\right),
\end{equation}
where $\mathcal{T}$ denotes the tool set. These tools expose information that is
better retrieved or verified on demand than embedded in a monolithic prompt,
including surrounding context, confirmed terminology, similar translated
sentences, subtitle constraints, domain knowledge, idiom lookup, and
target-language fluency checks. Appendix~\ref{app:tool-reference} documents the
tool interfaces and execution modes.

\textbf{Judge and refinement.}
The judge evaluates each candidate along semantic accuracy, fluency, style and
register, long-range consistency, and subtitle display constraints:
\begin{equation}
    c_i^{(r)},\delta_i^{(r)}
    = J\!\left(y_i^{(r)},d_i,M_t\right),
\end{equation}
where $c_i^{(r)}$ is the score and $\delta_i^{(r)}$ the corresponding critique.
SMART selects the highest-scoring candidate and refines it using this feedback:
\begin{equation}
    r^\star=\arg\max_{r\in\mathcal{A}_i} c_i^{(r)}, \qquad
    \hat{y}_i =
    R\!\left(y_i^{(r^\star)},\delta_i^{(r^\star)},M_t\right).
\end{equation}
The accepted translation updates the episode history and persistent memory.
After each episode, an episode-level pass revisits the subtitles to repair
residual consistency errors that are difficult to detect from a single sentence.
Appendix~\ref{app:end-to-end-example} illustrates the complete sentence-level
decision path.

\subsection{Self-Evolution at Test-Time Training}
\label{sec:self-evolution}

SMART evolves its translation process from feedback collected on
$\mathcal{D}_{\mathrm{tr}}$. After each training batch
$\mathcal{B}_t$, an evaluator aggregates candidate scores and textual critiques
into a structured feedback signal:
\begin{equation}
    \Delta_t =
    E\!\left(
        \left\{
            c_i^{(r)}, \delta_i^{(r)}
        \right\}_{i \in \mathcal{B}_t,\; r \in \mathcal{A}_i}
    \right).
\end{equation}
This feedback updates both translator prompts and the routing policy:
\begin{equation}
    \Pi_{t+1} = F^{\Pi}(\Pi_t; \Delta_t), \qquad
    \rho_{t+1} = F^{\rho}(\rho_t; \Delta_t).
\end{equation}
$F^{\Pi}$ revises translator prompts to address recurring failure modes, while
$F^{\rho}$ adjusts the router toward translator configurations that receive
higher scores on similar inputs. Both updates are expressed in natural
language and occur entirely at test time; the underlying LLM parameters remain
fixed. Appendix~\ref{app:prompt-update-example} and
Appendix~\ref{app:routing-update-example} provide examples of the two update
processes.

Both are constrained rewrites, not free-form edits: $F^{\Pi}$ revises an
underperforming agent's prompt while preserving its role and workflow, and
$F^{\rho}$ rewrites a table from routing categories to agent and tool
subsets under a floor of two agents per category. Both apply once per
training epoch from batch-aggregated feedback rather than per sentence.
Appendix~\ref{app:evolution-operators} gives an example.

After $T$ training steps, SMART freezes $(\Pi_T,\rho_T)$ and applies the
resulting configuration to $\mathcal{D}_{\mathrm{inf}}$. Self-evolution thus
changes the translation policy only during test-time training, whereas the
memory $M$ continues updating during inference to preserve cross-sentence and
cross-episode consistency. This separation allows SMART to adapt its
translation strategy without leaking held-out feedback into the frozen
inference configuration. Appendix~\ref{app:test-time-training-loop} gives the
complete optimization loop.

\section{Subtitle Arena: Benchmark and Evaluation}
\subsection{Benchmark Construction}
\label{sec:benchmark-construction}

\textbf{Benchmark scope.}
Existing subtitle benchmarks are not designed to fully capture the challenges of long-form viewing, including discourse preservation, terminology and character consistency, and subtitle-specific display constraints. We therefore introduce \textbf{Subtitle Arena}, a series-centric benchmark containing $192$ television series, $6{,}267$ English source episodes, and $70{,}664$ aligned bilingual episode pairs across $15$ target locales.

As shown in Table~\ref{tab:subtitle-arena-statistics}, Subtitle Arena provides broad coverage across $15$ target locales and $14$ television genres. Six locales contain more than $5{,}000$ aligned episodes, and all locales retain more than $3{,}200$ episode pairs. Its series-level organization supports evaluation over continuous narrative contexts, while the diversity of languages and genres tests robustness across different discourse, stylistic, and locale-specific conditions. Benchmark statistics are provided in Appendix~\ref{app:subtitle-arena-visualization}.

\begin{wraptable}{r}{0.65\textwidth}
\centering
\caption{
Statistics of \textbf{Subtitle Arena}.
\textbf{Top:} aligned bilingual episode coverage for each target locale, measured against $6{,}267$ English source episodes.
\textbf{Bottom:} distribution of the $192$ television series across $14$ genres.
}
\vspace{-10pt}
\label{tab:subtitle-arena-statistics}

\scriptsize
\setlength{\tabcolsep}{3.2pt}
\renewcommand{\arraystretch}{1.06}

\resizebox{\linewidth}{!}{%
\begin{tabular}{lcccccccc}
\toprule

\rowcolor{tablegray}
\multicolumn{9}{c}{\textbf{Aligned Episode Coverage by Target Locale}} \\
\midrule

\textbf{Locale}
& zh-CN
& pt-BR
& es-ES
& es-MX
& fr-FR
& ro-RO
& tr-TR
& pt-PT \\
\textbf{Episodes}
& 6,267
& 5,407
& 5,144
& 5,144
& 5,079
& 5,002
& 4,956
& 4,853 \\

\addlinespace[2pt]

\textbf{Locale}
& de-DE
& it-IT
& nl-NL
& sv-SE
& da-DK
& no-NO
& ko-KR
& \textbf{Total} \\
\textbf{Episodes}
& 4,697
& 4,518
& 4,518
& 4,286
& 4,273
& 3,277
& 3,243
& \textbf{70,664} \\

\midrule

\rowcolor{tablegray}
\multicolumn{9}{c}{\textbf{Series Distribution by Genre}} \\
\midrule

\textbf{Genre}
& Comedy
& Action
& Crime
& Animation
& Fantasy
& Drama
& Sci-Fi
& \\
\textbf{Series}
& 20
& 20
& 20
& 18
& 15
& 15
& 15
& \\

\textbf{Genre}
& Documentary
& Romance
& Mystery
& Horror
& Adventure
& Thriller
& Reality
& \textbf{Total} \\
\textbf{Series}
& 12
& 12
& 10
& 10
& 10
& 10
& 5
& \textbf{192} \\

\bottomrule
\end{tabular}%
}

\vspace{-10pt}
\end{wraptable}

\textbf{Dataset construction.}
Subtitle Arena is constructed from OpenSubtitles2024~\citep{tiedemann-luo-2026-opensubtitles2024}, which contains subtitles released before 2024. Rather than treating aligned sentences as independent translation examples, we reorganize the data at the episode and series levels. Specifically, we recover bilingual correspondences and associate subtitle files using IMDb identifiers, season indices, and episode indices. For each subtitle cue, we preserve the original start and end timecodes, normalize malformed formatting, and remove empty, one-sided, or unparsable alignments. We then group aligned episodes from the same television series to preserve long-range narrative structure. Full details on data processing, locale mapping, and filtering are provided in Appendix~\ref{app:subtitle-arena-details}.

\textbf{Purpose.}
Subtitle Arena is designed to test three properties that are underrepresented in conventional MT evaluation: \emph{(i)} discourse-aware translation, where long-range context is needed to interpret and translate the current sentence; \emph{(ii)} consistency of terminology, character references, and style across extended narrative horizons; and \emph{(iii)} compliance with subtitle-specific display requirements~\citep{papi2023direct,wilken-etal-2022-suber}. Its series-level organization further allows us to test whether translation behavior remains robust across different genres, scenes, and locale-specific conventions.

\subsection{LLM-as-a-Judge Evaluation with SubMQM}

\begin{wraptable}{r}{0.65\columnwidth}
\vspace{-12pt}
\centering
\caption{\textbf{Overview of SubMQM.} Each error type is assigned a penalty
in $\{0,5,10\}$; lower is better. The complete 19-type rubric is provided
in Appendix~\ref{app:mqm-protocol}.}
\label{tab:submqm-summary}
\vspace{-8pt}
\scriptsize
\resizebox{\linewidth}{!}{%
\begin{tabular}{llcc}
\toprule
\rowcolor{tablegray}
\textbf{Dimension} &
\textbf{Representative Errors} &
\textbf{\#} &
\textbf{Weight} \\
\midrule

Accuracy
& Mistranslation, omission, overtranslation
& 3 & 0.30 \\

Terminology
& Name and term inconsistency
& 2 & 0.20 \\

Fluency
& Coherence, naturalness, vividness
& 3 & 0.20 \\

Audience Appropriateness
& Profanity, formality
& 2 & 0.12 \\

Linguistic Conventions
& Punctuation, capitalization, grammar, spacing
& 4 & 0.08 \\

Technical
& Line breaking, CPL, lines per box
& 3 & 0.06 \\

Locale Conventions
& Localization, language detection
& 2 & 0.04 \\

\midrule
\rowcolor{tablegray}
\textbf{Total}
& \textbf{19 error types}
& \textbf{19}
& \textbf{1.00} \\
\bottomrule
\end{tabular}%
}
\vspace{-12pt}
\end{wraptable}

For the LLM-as-a-judge evaluation, we develop a new rubric-based
evaluation metric, \textbf{SubMQM}, a subtitle-adapted Multidimensional
Quality Metrics (MQM) protocol. For each hypothesis, an LLM evaluator
assigns error penalties along seven dimensions---Terminology, Accuracy,
Fluency, Linguistic Conventions, Technical, Locale Conventions, and
Audience Appropriateness---which are further decomposed into 19 error
types. Each error type receives a penalty in $\{0,5,10\}$ for no, minor,
and severe error, respectively; dimension-level scores aggregate the
corresponding error types, and Overall places greater weight on semantic
fidelity. All SubMQM results are therefore reported as penalties, where
lower is better. We use the same evaluator and rubric for all systems.
The complete rubric, weighting scheme, and alignment procedure are
provided in Appendix~\ref{app:mqm-protocol}.

\section{Experiments}
\subsection{Experimental Setup}
\label{sec:experimental-setup}

\textbf{Baselines.} We compare against three classes of systems. \emph{Online} refers to community-authored subtitles distributed with the corresponding episodes in OpenSubtitles. Because these subtitles are collected in the wild and vary in quality, we do not treat them as a controlled human upper bound. \emph{Single-call LLMs} include Gemma 3 4B, DeepSeek-V3.2, Claude Sonnet 4.6, Claude Opus 4.8, and GPT-5.5. These models receive the same local context and subtitle-formatting instructions as SMART, but translate each sentence once without routing, persistent memory, tools, candidate selection, or refinement. \emph{Agentic baselines} include TransAgent~\citep{wu-etal-2024-transagents} and DRT~\citep{wang2025drt}, both reimplemented with Claude Sonnet 4.6 for a controlled comparison with SMART. On the public MuSC benchmark~\citep{cui2026utterance}, we additionally report the systems evaluated in the original benchmark, including the fine-tuned ALPO variant of Qwen2.5-14B.

\textbf{Implementation.}
Unless otherwise specified, all SMART roles use Claude Sonnet 4.6. For each series, the first $30\%$ of the available content is used for test-time training and the remaining $70\%$ for held-out inference. Prompt and routing updates are permitted only during training and are frozen thereafter; persistent series memory continues to accumulate because it represents task state. The episode-level consistency check operates over overlapping windows of size $5$, $10$, and $20$ with $50\%$ overlap. Section~\ref{sec:backbone-evaluator} tests weaker translation backbones and an alternative evaluator.

\subsection{Main Results}
\label{sec:main-results}

\begin{table*}[ht]
\centering
\caption{
SubMQM results on representative bidirectional language pairs from Subtitle Arena. Each cell reports both directions in the order shown in the first column (left $|$ right). Dimension scores are averaged over their constituent error types. All values are penalties (lower is better). The best and second-best values are highlighted in blue and orange, respectively.
}
\vspace{-10pt}
\label{tab:subtitle_arena_main}
\scriptsize
\setlength{\tabcolsep}{3.0pt}
\renewcommand{\arraystretch}{1.08}

\resizebox{\textwidth}{!}{%
\begin{tabular}{llcccccccc}
\toprule
\rowcolor{tablegray}
\textbf{Language Direction} &
\textbf{Method} &
\textbf{Term.} &
\textbf{Acc.} &
\textbf{Flu.} &
\textbf{Ling.} &
\textbf{Tech.} &
\textbf{Locale} &
\textbf{Audience} &
\textbf{Overall} \\
\midrule

\multirow{9}{*}{\centering
en$\rightarrow$zh $|$ zh$\rightarrow$en}
& Online
& 0.35 $|$ 0.42
& 4.87 $|$ 4.23
& 4.07 $|$ 3.28
& 1.98 $|$ 1.47
& 0.20 $|$ 0.29
& 1.40 $|$ 0.11
& \best{0.05} $|$ 0.22
& 2.58 $|$ 2.17 \\

& Gemma 3 4B
& 0.49 $|$ 0.52
& 2.76 $|$ 3.10
& 1.97 $|$ 2.16
& 0.59 $|$ 0.81
& 0.79 $|$ 1.00
& 0.15 $|$ 0.16
& 0.35 $|$ 0.39
& 1.46 $|$ 1.64 \\

& DeepSeek-V3.2
& 0.28 $|$ 0.31
& 1.86 $|$ 2.03
& 1.25 $|$ 1.43
& 0.29 $|$ 0.42
& 0.49 $|$ 0.88
& \best{0.03} $|$ 0.08
& 0.13 $|$ 0.18
& 0.93 $|$ 1.07 \\

& Claude 4.6
& 0.26 $|$ 0.27
& 1.63 $|$ 1.84
& 1.16 $|$ 1.28
& 0.26 $|$ 0.34
& 0.42 $|$ 0.83
& \second{0.03} $|$ 0.05
& 0.16 $|$ 0.15
& 0.84 $|$ 0.96 \\

& Claude 4.8
& 0.25 $|$ 0.20
& 1.56 $|$ 1.39
& 1.09 $|$ 1.00
& 0.20 $|$ 0.23
& 0.32 $|$ 0.68
& 0.04 $|$ 0.03
& 0.13 $|$ 0.11
& 0.78 $|$ 0.73 \\

& GPT-5.5
& 0.20 $|$ 0.18
& 1.16 $|$ 1.20
& 1.13 $|$ 0.98
& 0.19 $|$ 0.19
& 0.22 $|$ 0.87
& \second{0.03} $|$ 0.03
& 0.14 $|$ 0.13
& 0.66 $|$ 0.67 \\

\cdashline{2-10}

& DRT
& 0.22 $|$ 0.16
& 1.11 $|$ 1.01
& \second{1.04} $|$ \second{0.86}
& 0.12 $|$ 0.15
& 0.12 $|$ 0.30
& 0.04 $|$ \second{0.03}
& \second{0.11} $|$ 0.11
& 0.61 $|$ 0.55 \\

& TransAgent
& \second{0.16} $|$ \second{0.14}
& \second{0.85} $|$ \second{0.92}
& 1.06 $|$ \best{0.79}
& \second{0.08} $|$ \second{0.13}
& \second{0.05} $|$ \second{0.25}
& 0.04 $|$ \second{0.03}
& 0.14 $|$ \second{0.10}
& \second{0.52} $|$ \second{0.50} \\

\rowcolor{gray!4}
& \textbf{SMART}
& \best{0.15} $|$ \best{0.13}
& \best{0.80} $|$ \best{0.80}
& \best{0.95} $|$ \best{0.79}
& \best{0.07} $|$ \best{0.10}
& \best{0.00} $|$ \best{0.08}
& 0.04 $|$ \best{0.02}
& 0.12 $|$ \best{0.09}
& \best{0.48} $|$ \best{0.44} \\

\midrule

\multirow{9}{*}{\centering
en$\rightarrow$de $|$ de$\rightarrow$en}
& Online
& 0.51 $|$ 0.39
& 6.69 $|$ 3.69
& 2.76 $|$ 2.45
& 0.61 $|$ 1.01
& \best{0.05} $|$ 0.23
& 0.91 $|$ 0.29
& \best{0.16} $|$ 0.19
& 2.77 $|$ 1.80 \\

& Gemma 3 4B
& 0.70 $|$ 0.46
& 4.18 $|$ 2.86
& 2.74 $|$ 1.92
& 0.86 $|$ 0.67
& 1.77 $|$ 0.90
& 0.48 $|$ 0.19
& 0.58 $|$ 0.32
& 2.21 $|$ 1.49 \\

& DeepSeek-V3.2
& 0.44 $|$ 0.29
& 2.90 $|$ 1.88
& 1.92 $|$ 1.28
& 0.41 $|$ 0.34
& 1.38 $|$ 0.82
& 0.22 $|$ 0.21
& 0.31 $|$ 0.52
& 1.50 $|$ 1.02 \\

& Claude 4.6
& 0.41 $|$ 0.25
& 2.56 $|$ 1.71
& 1.81 $|$ 1.16
& 0.41 $|$ 0.28
& \second{1.18} $|$ 0.76
& 0.19 $|$ 0.08
& 0.31 $|$ 0.13
& 1.36 $|$ 0.88 \\

& Claude 4.8
& 0.40 $|$ 0.18
& 2.31 $|$ 1.29
& 1.54 $|$ 0.92
& 0.37 $|$ 0.19
& 1.59 $|$ 0.61
& 0.15 $|$ 0.06
& \second{0.28} $|$ 0.10
& 1.25 $|$ 0.67 \\

& GPT-5.5
& 0.36 $|$ 0.16
& 2.07 $|$ 1.11
& 1.71 $|$ 0.89
& 0.32 $|$ 0.16
& 1.58 $|$ 0.78
& 0.11 $|$ 0.05
& 0.38 $|$ 0.11
& 1.21 $|$ 0.62 \\

\cdashline{2-10}

& DRT
& \best{0.24} $|$ 0.15
& 1.62 $|$ 0.94
& 1.62 $|$ 0.78
& 0.28 $|$ 0.13
& 1.59 $|$ 0.27
& 0.08 $|$ \second{0.04}
& 0.35 $|$ 0.09
& 1.02 $|$ 0.50 \\

& TransAgent
& 0.26 $|$ \second{0.13}
& \second{1.51} $|$ \second{0.85}
& \second{1.46} $|$ \second{0.71}
& \second{0.27} $|$ \second{0.11}
& 1.31 $|$ \second{0.21}
& \second{0.05} $|$ \best{0.04}
& 0.40 $|$ \second{0.08}
& \second{0.95} $|$ \second{0.45} \\

\rowcolor{gray!4}
& \textbf{SMART}
& \second{0.25} $|$ \best{0.12}
& \best{1.24} $|$ \best{0.74}
& \best{1.46} $|$ \best{0.71}
& \best{0.22} $|$ \best{0.08}
& 1.55 $|$ \best{0.07}
& \best{0.03} $|$ \best{0.04}
& 0.41 $|$ \best{0.07}
& \best{0.87} $|$ \best{0.41} \\

\midrule

\multirow{9}{*}{\centering
en$\rightarrow$ko $|$ ko$\rightarrow$en}
& Online
& 0.41 $|$ 0.44
& 2.91 $|$ 4.44
& 1.87 $|$ 3.40
& 0.63 $|$ 1.56
& 0.27 $|$ 0.34
& \second{0.04} $|$ 0.12
& 0.70 $|$ 0.24
& 1.48 $|$ 2.28 \\

& Gemma 3 4B
& 0.68 $|$ 0.54
& 3.23 $|$ 3.30
& 2.43 $|$ 2.29
& 0.64 $|$ 0.87
& 0.80 $|$ 1.04
& 0.22 $|$ 0.17
& 1.04 $|$ 0.42
& 1.82 $|$ 1.74 \\

& DeepSeek-V3.2
& 0.42 $|$ 0.33
& 2.09 $|$ 2.16
& 1.63 $|$ 1.51
& 0.31 $|$ 0.44
& 0.47 $|$ 0.94
& \second{0.04} $|$ 0.08
& \second{0.66} $|$ 0.19
& 1.17 $|$ 1.13 \\

& Claude 4.6
& 0.40 $|$ 0.29
& 2.03 $|$ 1.93
& \best{1.32} $|$ 1.34
& 0.29 $|$ 0.36
& 0.44 $|$ 0.87
& \second{0.04} $|$ 0.07
& \best{0.61} $|$ 0.16
& 1.07 $|$ 1.01 \\

& Claude 4.8
& 0.35 $|$ 0.22
& 1.80 $|$ 1.46
& 1.56 $|$ 1.05
& 0.27 $|$ 0.25
& 0.35 $|$ 0.71
& \second{0.04} $|$ 0.05
& 0.72 $|$ 0.13
& 1.05 $|$ 0.77 \\

& GPT-5.5
& 0.37 $|$ 0.19
& 1.85 $|$ 1.25
& 1.47 $|$ 1.03
& 0.22 $|$ 0.21
& 0.25 $|$ 0.91
& \best{0.04} $|$ 0.05
& 0.68 $|$ 0.14
& 1.04 $|$ 0.71 \\

\cdashline{2-10}

& DRT
& \best{0.28} $|$ 0.17
& 1.71 $|$ 1.05
& 1.50 $|$ \second{0.90}
& 0.22 $|$ 0.16
& 0.20 $|$ 0.31
& \second{0.04} $|$ \second{0.04}
& 0.75 $|$ 0.11
& 0.99 $|$ 0.57 \\

& TransAgent
& 0.32 $|$ \second{0.15}
& \second{1.63} $|$ \second{0.96}
& 1.57 $|$ \best{0.82}
& \second{0.19} $|$ \second{0.14}
& \second{0.13} $|$ \second{0.26}
& 0.05 $|$ \best{0.04}
& 0.67 $|$ \second{0.11}
& \second{0.97} $|$ \second{0.52} \\

\rowcolor{gray!4}
& \textbf{SMART}
& \second{0.29} $|$ \best{0.14}
& \best{1.55} $|$ \best{0.84}
& \second{1.43} $|$ \best{0.82}
& \best{0.17} $|$ \best{0.11}
& \best{0.10} $|$ \best{0.10}
& \second{0.04} $|$ \best{0.04}
& 0.79 $|$ \best{0.10}
& \best{0.92} $|$ \best{0.47} \\

\midrule

\multirow{9}{*}{\centering
en$\rightarrow$it $|$ it$\rightarrow$en}
& Online
& 0.54 $|$ 0.34
& 6.36 $|$ 3.40
& 3.04 $|$ 2.30
& 2.21 $|$ 0.92
& \best{0.13} $|$ \second{0.20}
& 0.68 $|$ 0.22
& \best{0.07} $|$ 0.16
& 2.85 $|$ 1.66 \\

& Gemma 3 4B
& 0.70 $|$ 0.42
& 4.25 $|$ 2.70
& 3.25 $|$ 1.80
& 1.01 $|$ 0.62
& 1.69 $|$ 0.85
& 0.44 $|$ 0.15
& 0.62 $|$ 0.28
& 2.34 $|$ 1.39 \\

& DeepSeek-V3.2
& 0.43 $|$ 0.26
& 2.94 $|$ 1.75
& 2.46 $|$ 1.20
& 0.55 $|$ 0.31
& \second{1.29} $|$ 0.77
& 0.20 $|$ 0.09
& 0.29 $|$ 0.14
& 1.62 $|$ 0.91 \\

& Claude 4.6
& 0.38 $|$ 0.22
& 2.54 $|$ 1.59
& 2.42 $|$ 1.07
& 0.53 $|$ 0.25
& 1.48 $|$ 0.72
& 0.20 $|$ 0.07
& \second{0.27} $|$ 0.12
& 1.49 $|$ 0.82 \\

& Claude 4.8
& 0.37 $|$ 0.16
& 2.28 $|$ 1.19
& 2.43 $|$ 0.85
& 0.43 $|$ 0.17
& 1.36 $|$ 0.58
& 0.19 $|$ 0.05
& \second{0.27} $|$ 0.09
& 1.39 $|$ 0.62 \\

& GPT-5.5
& 0.32 $|$ 0.15
& 2.13 $|$ 1.03
& 2.43 $|$ 0.83
& 0.39 $|$ 0.14
& 1.82 $|$ 0.73
& 0.15 $|$ 0.05
& 0.31 $|$ 0.10
& 1.37 $|$ 0.57 \\

\cdashline{2-10}
& DRT
& \second{0.27} $|$ 0.13
& \second{1.66} $|$ 0.86
& 2.23 $|$ 0.73
& 0.37 $|$ 0.11
& 1.89 $|$ 0.25
& \second{0.13} $|$ \second{0.04}
& 0.34 $|$ 0.08
& 1.19 $|$ 0.46 \\

& TransAgent
& \best{0.21} $|$ \second{0.12}
& 1.73 $|$ \second{0.78}
& \best{2.03} $|$ \second{0.66}
& \second{0.29} $|$ \second{0.10}
& 2.28 $|$ \second{0.20}
& \second{0.13} $|$ \best{0.03}
& 0.33 $|$ \second{0.07}
& \second{1.17} $|$ \second{0.42} \\

\rowcolor{gray!4}
& \textbf{SMART}
& \best{0.21} $|$ \best{0.11}
& \best{1.49} $|$ \best{0.69}
& \second{2.15} $|$ \best{0.65}
& \best{0.28} $|$ \best{0.08}
& 2.37 $|$ \best{0.07}
& \best{0.11} $|$ \best{0.03}
& 0.35 $|$ \best{0.06}
& \best{1.13} $|$ \best{0.38} \\

\bottomrule
\end{tabular}%
}

\vspace{3pt}
\begin{minipage}{0.99\textwidth}
\scriptsize
\textbf{Abbreviations.}
Term. = Terminology;
Acc. = Accuracy;
Flu. = Fluency;
Ling. = Linguistic Conventions;
Tech. = Technical.
Within each cell, the left and right values correspond to the left and
right translation directions in the first column, respectively.
Complete fine-grained results for all 19 SubMQM error types and all 30 translation
directions are provided in Appendix~\ref{app:subtitle-arena-out-en} and
Appendix~\ref{app:subtitle-arena-into-en}: English$\rightarrow$locale results are
reported in Tables~\ref{tab:subtitle_arena_out_en_semantic} and
\ref{tab:subtitle_arena_out_en_form}, while locale$\rightarrow$English results are
reported in Tables~\ref{tab:subtitle_arena_into_en_semantic} and
\ref{tab:subtitle_arena_into_en_form}.
\end{minipage}
\vspace{-12pt}
\end{table*}

\textbf{Results on Subtitle Arena.}
Table~\ref{tab:subtitle_arena_main} summarizes representative bidirectional results on Subtitle Arena. Complete fine-grained results for all 30 translation directions and all 19 SubMQM error types are provided in Appendix~\ref{app:subtitle-arena-out-en} and Appendix~\ref{app:subtitle-arena-into-en}. Across the 15 English$\rightarrow$locale directions, SMART obtains the lowest Overall SubMQM penalty in every direction and reduces the mean Overall penalty from $1.20$ (TransAgent) to $1.11$, a relative reduction of $6.9\%$. Despite using Claude Sonnet 4.6 as its default backbone, SMART also improves over stronger single-call translators, including GPT-5.5 ($1.44$) and Claude Opus 4.8 ($1.55$) on average.

The improvements are distributed across semantic and subtitle-specific dimensions rather than being driven by one error type. Relative to TransAgent, SMART lowers the mean Accuracy penalty from $1.83$ to $1.67$ and the mean Technical penalty from $1.83$ to $1.72$, while also reducing Terminology, Fluency, Linguistic Conventions, Locale Conventions, and Audience Appropriateness. The reverse locale$\rightarrow$English evaluation shows the same overall trend: SMART reduces the mean Overall penalty from $0.41$ to $0.37$, a $10.0\%$ relative reduction, and achieves the lowest Overall score in all 15 directions. These results indicate that the gains are not only specific to generating diverse target languages but also persist when every system generates English.

\begin{wraptable}{r}{0.65\textwidth}
\vspace{-12pt}
\centering
\caption{
MuSC results across four directions. Higher is better. Best and second-best are highlighted in blue-gray and warm beige.
}
\vspace{-10pt}
\label{tab:musc_results}

\scriptsize
\setlength{\tabcolsep}{3.2pt}
\renewcommand{\arraystretch}{1}

\resizebox{\linewidth}{!}{%
\begin{tabular}{llcccccccccccc}
\toprule

\rowcolor{tablegray}
& &
\multicolumn{3}{c}{\textbf{en$\rightarrow$zh}} &
\multicolumn{3}{c}{\textbf{ko$\rightarrow$zh}} &
\multicolumn{3}{c}{\textbf{zh$\rightarrow$en}} &
\multicolumn{3}{c}{\textbf{zh$\rightarrow$th}} \\

\rowcolor{tablegray}
\multirow{-2}{*}{\textbf{Model}} &
\multirow{-2}{*}{\textbf{Training}} &
\textbf{Acc.} & \textbf{Nat.} & \textbf{Viv.} &
\textbf{Acc.} & \textbf{Nat.} & \textbf{Viv.} &
\textbf{Acc.} & \textbf{Nat.} & \textbf{Viv.} &
\textbf{Acc.} & \textbf{Nat.} & \textbf{Viv.} \\

\midrule

Gold Reference & Human
& 83.6 & 82.6 & 71.5
& 78.0 & 77.8 & 65.8
& 83.0 & 80.3 & 73.3
& 76.6 & 75.1 & 66.3 \\

\midrule

VideoDubber & --
& 46.9 & 51.9 & 49.7
& 39.6 & 45.2 & 48.2
& 53.6 & 54.8 & 50.1
& 34.1 & 34.9 & 41.5 \\

NLLB-3.3B & --
& 61.4 & 54.0 & 43.7
& 33.1 & 26.1 & 25.4
& 29.1 & 21.7 & 20.8
& 42.6 & 33.9 & 40.5 \\

MADLAD-10B & --
& 59.7 & 55.5 & 46.3
& 44.9 & 42.9 & 46.7
& 45.1 & 38.9 & 37.6
& 47.9 & 50.8 & 51.0 \\

Google Translate & --
& 84.2 & 79.7 & 54.4
& 54.9 & 52.8 & 52.0
& 79.8 & 66.3 & 50.2
& 55.2 & 56.2 & 54.5 \\

\cdashline{1-14}

GPT-4o & ICL (C)
& 89.3 & 82.3 & 59.8
& 80.0 & 79.9 & 58.1
& 88.5 & 83.0 & 64.6
& 88.0 & 84.4 & 67.9 \\

Qwen-Max & ICL (C)
& 91.9 & 84.4 & 61.3
& 83.7 & 82.5 & 61.8
& \secondcell{90.0} & 85.0 & 66.8
& 91.3 & \secondcell{85.8} & 69.1 \\

DeepSeek-V3.1 & ICL (C)
& 91.2 & 85.3 & 63.5
& 83.1 & 82.2 & 57.2
& 89.5 & 84.1 & 63.0
& 89.9 & 84.6 & 67.1 \\

DeepSeek-R1 & ICL (R)
& 90.5 & 85.7 & 70.8
& 79.8 & 81.6 & 65.6
& 88.5 & 85.6 & 73.5
& 87.6 & 84.0 & 71.0 \\

GPT-5 & ICL (R)
& \secondcell{92.4}
& \secondcell{87.0}
& 71.1
& \secondcell{84.5}
& 82.6
& 65.0
& 89.1
& 86.1
& 75.2
& 88.7
& 83.9
& 73.0 \\

\cdashline{1-14}

Qwen2.5-14B & SFT
& 86.4 & 82.0 & 59.1
& 80.9 & 76.1 & 53.9
& 85.2 & 80.1 & 54.8
& 87.3 & 82.6 & 66.0 \\

Qwen2.5-14B & ALPO
& 90.6
& 84.3
& \secondcell{76.6}
& 84.3
& \secondcell{83.3}
& \secondcell{70.5}
& 88.3
& \secondcell{86.8}
& \secondcell{81.7}
& \secondcell{91.9}
& 84.7
& \secondcell{74.2} \\

\rowcolor{gray!4}
\textbf{SMART} & --
& \bestcell{94.3}
& \bestcell{88.2}
& \bestcell{80.4}
& \bestcell{86.8}
& \bestcell{85.9}
& \bestcell{78.4}
& \bestcell{92.3}
& \bestcell{88.6}
& \bestcell{83.7}
& \bestcell{94.2}
& \bestcell{87.0}
& \bestcell{80.5} \\

\bottomrule
\end{tabular}%
}

\vspace{2pt}
\begin{minipage}{\linewidth}
\scriptsize
\textbf{Abbreviations.}
Acc. = Accuracy; Nat. = Naturalness; Viv. = Vividness.
SFT = supervised fine-tuning; ICL = in-context learning;
(C) = chat model; (R) = reasoning model.
\end{minipage}

\vspace{-10pt}
\end{wraptable}

\textbf{Results on the MuSC dataset.}
We further evaluate SMART on MuSC to test whether its gains transfer beyond Subtitle Arena. As shown in Table~\ref{tab:musc_results}, SMART achieves the best model result on all Accuracy, Naturalness, and Vividness measurements across the four translation directions. Compared to the strongest non-SMART result in each column, SMART improves by an average of approximately $3.0$ points, with gains ranging from $1.2$ to $7.9$ points. The advantage remains clear against both strong reasoning models and the fine-tuned ALPO baseline. The largest gains appear in Vividness, where SMART improves over the strongest prior system by $3.8$, $7.9$, $2.0$, and $6.3$ points on en$\rightarrow$zh, ko$\rightarrow$zh, zh$\rightarrow$en, and zh$\rightarrow$th, respectively. This result is particularly important for subtitle translation, where high-quality outputs require not only semantic fidelity but also contextual and stylistic adaptation over long-form narrative content.

\begin{table*}[ht]
\centering
\vspace{-10pt}
\caption{
Backbone and judge analysis on 4 bidirectional pairs from Subtitle Arena.
Each cell gives both directions in the first column's order (left $|$ right).
All values are penalties (lower is better). Blue-gray and warm-beige mark
the best and second-best values.
}
\vspace{-10pt}
\label{tab:backbone_judge_main}
\scriptsize
\setlength{\tabcolsep}{3.0pt}
\renewcommand{\arraystretch}{1.06}
\resizebox{\textwidth}{!}{%
\begin{tabular}{llcccccccc}
\toprule
\rowcolor{tablegray}
\textbf{Language Direction} &
\textbf{Method / Setting} &
\textbf{Term.} &
\textbf{Acc.} &
\textbf{Flu.} &
\textbf{Ling.} &
\textbf{Tech.} &
\textbf{Locale} &
\textbf{Audience} &
\textbf{Overall} \\
\midrule

\multirow{6}{*}{\centering en$\rightarrow$zh $|$ zh$\rightarrow$en} & Gemma 3 4B & 0.49 $|$ 0.52 & 2.76 $|$ 3.10 & 1.97 $|$ 2.16 & 0.59 $|$ 0.81 & 0.79 $|$ 1.00 & 0.15 $|$ 0.16 & 0.35 $|$ 0.39 & 1.46 $|$ 1.64 \\
 & DeepSeek-V3.2 & 0.28 $|$ 0.31 & 1.86 $|$ 2.03 & 1.25 $|$ 1.43 & 0.29 $|$ 0.42 & 0.49 $|$ 0.88 & \best{0.03} $|$ 0.08 & \second{0.13} $|$ 0.18 & 0.93 $|$ 1.07 \\
\cdashline{2-10}
 & SMART (Gemma 3 4B) & 0.23 $|$ 0.19 & 1.14 $|$ 1.12 & 1.27 $|$ 1.06 & 0.11 $|$ 0.16 & 0.03 $|$ 0.11 & 0.07 $|$ 0.05 & 0.19 $|$ 0.12 & 0.67 $|$ 0.62 \\
 & SMART (DeepSeek-V3.2) & \second{0.18} $|$ 0.15 & \second{0.89} $|$ 0.92 & 1.04 $|$ 0.91 & \second{0.08} $|$ \second{0.12} & \second{0.01} $|$ \second{0.09} & 0.05 $|$ \second{0.03} & 0.14 $|$ \second{0.10} & 0.53 $|$ \second{0.51} \\
 & SMART (Judge: GPT-5.5) & \best{0.15} $|$ \best{0.12} & \best{0.80} $|$ \best{0.76} & \best{0.90} $|$ \second{0.82} & \best{0.07} $|$ \best{0.10} & \best{0.00} $|$ \best{0.08} & \second{0.04} $|$ \best{0.02} & \best{0.12} $|$ \best{0.09} & \best{0.47} $|$ \best{0.44} \\
\rowcolor{gray!4}  & \textbf{SMART (Claude 4.6)} & \best{0.15} $|$ \second{0.13} & \best{0.80} $|$ \second{0.80} & \second{0.95} $|$ \best{0.79} & \best{0.07} $|$ \best{0.10} & \best{0.00} $|$ \best{0.08} & \second{0.04} $|$ \best{0.02} & \best{0.12} $|$ \best{0.09} & \second{0.48} $|$ \best{0.44} \\
\midrule
\multirow{6}{*}{\centering en$\rightarrow$de $|$ de$\rightarrow$en} & Gemma 3 4B & 0.70 $|$ 0.46 & 4.18 $|$ 2.86 & 2.74 $|$ 1.92 & 0.86 $|$ 0.67 & 1.77 $|$ 0.90 & 0.48 $|$ 0.19 & 0.58 $|$ 0.32 & 2.21 $|$ 1.49 \\
 & DeepSeek-V3.2 & 0.44 $|$ 0.29 & 2.90 $|$ 1.88 & 1.92 $|$ 1.28 & 0.41 $|$ 0.34 & \best{1.38} $|$ 0.82 & 0.22 $|$ 0.21 & \best{0.31} $|$ 0.52 & 1.50 $|$ 1.02 \\
\cdashline{2-10}
 & SMART (Gemma 3 4B) & 0.33 $|$ 0.16 & 1.68 $|$ 1.10 & 2.18 $|$ 1.03 & 0.33 $|$ 0.13 & 2.15 $|$ 0.10 & 0.06 $|$ 0.07 & 0.60 $|$ 0.10 & 1.23 $|$ 0.60 \\
 & SMART (DeepSeek-V3.2) & \second{0.27} $|$ 0.14 & 1.41 $|$ 0.88 & 1.68 $|$ 0.80 & \second{0.26} $|$ 0.10 & 1.78 $|$ \second{0.08} & \second{0.04} $|$ \second{0.05} & 0.45 $|$ \second{0.08} & \second{0.99} $|$ \second{0.47} \\
 & SMART (Judge: GPT-5.5) & \best{0.25} $|$ \best{0.11} & \best{1.22} $|$ \second{0.78} & \second{1.51} $|$ \best{0.68} & \best{0.22} $|$ \second{0.09} & \second{1.51} $|$ \best{0.07} & \best{0.03} $|$ \best{0.04} & \second{0.39} $|$ \second{0.08} & \best{0.87} $|$ \best{0.41} \\
\rowcolor{gray!4}  & \textbf{SMART (Claude 4.6)} & \best{0.25} $|$ \second{0.12} & \second{1.24} $|$ \best{0.74} & \best{1.46} $|$ \second{0.71} & \best{0.22} $|$ \best{0.08} & 1.55 $|$ \best{0.07} & \best{0.03} $|$ \best{0.04} & 0.41 $|$ \best{0.07} & \best{0.87} $|$ \best{0.41} \\
\midrule
\multirow{6}{*}{\centering en$\rightarrow$ko $|$ ko$\rightarrow$en} & Gemma 3 4B & 0.68 $|$ 0.55 & 3.23 $|$ 3.30 & 2.43 $|$ 2.29 & 0.64 $|$ 0.87 & 0.80 $|$ 1.04 & 0.22 $|$ 0.17 & 1.04 $|$ 0.42 & 1.82 $|$ 1.74 \\
 & DeepSeek-V3.2 & 0.42 $|$ 0.33 & 2.09 $|$ 2.16 & \second{1.63} $|$ 1.51 & 0.31 $|$ 0.44 & 0.47 $|$ 0.94 & \best{0.04} $|$ 0.08 & \best{0.66} $|$ 0.19 & 1.17 $|$ 1.13 \\
\cdashline{2-10}
 & SMART (Gemma 3 4B) & 0.42 $|$ 0.22 & 2.03 $|$ 1.16 & 1.96 $|$ 1.23 & 0.25 $|$ 0.17 & 0.14 $|$ 0.15 & 0.07 $|$ 0.07 & 1.29 $|$ 0.15 & 1.26 $|$ 0.68 \\
 & SMART (DeepSeek-V3.2) & 0.33 $|$ \second{0.17} & 1.68 $|$ \second{0.98} & 1.65 $|$ 0.98 & 0.19 $|$ \second{0.13} & \second{0.12} $|$ \second{0.11} & \second{0.05} $|$ \second{0.05} & 0.95 $|$ \second{0.12} & 1.03 $|$ \second{0.55} \\
 & SMART (Judge: GPT-5.5) & \second{0.30} $|$ \best{0.14} & \best{1.53} $|$ \best{0.84} & \best{1.43} $|$ \second{0.84} & \second{0.18} $|$ \best{0.11} & \best{0.10} $|$ \best{0.10} & \best{0.04} $|$ \best{0.04} & \second{0.74} $|$ \best{0.10} & \best{0.91} $|$ \best{0.47} \\
\rowcolor{gray!4}  & \textbf{SMART (Claude 4.6)} & \best{0.29} $|$ \best{0.14} & \second{1.55} $|$ \best{0.84} & \best{1.43} $|$ \best{0.82} & \best{0.17} $|$ \best{0.11} & \best{0.10} $|$ \best{0.10} & \best{0.04} $|$ \best{0.04} & 0.79 $|$ \best{0.10} & \second{0.92} $|$ \best{0.47} \\
\midrule
\multirow{6}{*}{\centering en$\rightarrow$it $|$ it$\rightarrow$en} & Gemma 3 4B & 0.70 $|$ 0.42 & 4.25 $|$ 2.70 & 3.25 $|$ 1.80 & 1.01 $|$ 0.62 & \second{1.69} $|$ 0.85 & 0.44 $|$ 0.15 & 0.62 $|$ 0.28 & 2.34 $|$ 1.39 \\
 & DeepSeek-V3.2 & 0.43 $|$ 0.26 & 2.94 $|$ 1.75 & 2.46 $|$ 1.20 & 0.55 $|$ 0.31 & \best{1.29} $|$ 0.77 & 0.20 $|$ 0.09 & \best{0.29} $|$ 0.14 & 1.62 $|$ 0.91 \\
\cdashline{2-10}
 & SMART (Gemma 3 4B) & 0.33 $|$ 0.15 & 1.97 $|$ 1.01 & 3.03 $|$ 0.96 & 0.40 $|$ 0.12 & 3.45 $|$ 0.12 & 0.18 $|$ 0.06 & 0.47 $|$ 0.10 & 1.56 $|$ 0.55 \\
 & SMART (DeepSeek-V3.2) & 0.25 $|$ 0.12 & 1.69 $|$ \second{0.82} & 2.50 $|$ 0.76 & 0.33 $|$ \second{0.09} & 2.70 $|$ \second{0.09} & \second{0.14} $|$ \second{0.04} & 0.39 $|$ \second{0.08} & \second{1.30} $|$ \second{0.44} \\
 & SMART (Judge: GPT-5.5) & \second{0.23} $|$ \best{0.10} & \best{1.45} $|$ \best{0.69} & \second{2.19} $|$ \second{0.67} & \second{0.29} $|$ \best{0.08} & 2.42 $|$ \best{0.07} & \best{0.11} $|$ \best{0.03} & \second{0.35} $|$ \best{0.06} & \best{1.13} $|$ \best{0.38} \\
\rowcolor{gray!4}  & \textbf{SMART (Claude 4.6)} & \best{0.21} $|$ \second{0.11} & \second{1.49} $|$ \best{0.69} & \best{2.15} $|$ \best{0.65} & \best{0.28} $|$ \best{0.08} & 2.37 $|$ \best{0.07} & \best{0.11} $|$ \best{0.03} & \second{0.35} $|$ \best{0.06} & \best{1.13} $|$ \best{0.38} \\
\bottomrule
\end{tabular}%
}

\vspace{3pt}
\begin{minipage}{0.99\textwidth}
\scriptsize
\textbf{Abbreviations.}
Term. = Terminology; Acc. = Accuracy; Flu. = Fluency;
Ling. = Linguistic Conventions; Tech. = Technical.
Within each cell, the left and right values correspond to the left and right
translation directions shown in the first column, respectively.
SMART (Claude 4.6) is the default setting; SMART (Judge: GPT-5.5) changes
only the judge model.
Complete fine-grained results over all 15 bidirectional language pairs are
reported in Appendix~\ref{app:backbone-judge-full}: English$\rightarrow$locale results
are in Tables~\ref{tab:backbone_out_en_semantic} and
\ref{tab:backbone_out_en_form}, and locale$\rightarrow$English results are in
Tables~\ref{tab:backbone_into_en_semantic} and
\ref{tab:backbone_into_en_form}.
\end{minipage}
\vspace{-12pt}
\end{table*}

\textbf{Multimodal Expansion.}
We extend SMART with video and audio tools powered by Qwen3-Omni~\citep{xu2025qwen3}. Across six representative directions, SMART + MM reduces the average SubMQM penalty from 0.63 to 0.57 and outperforms the multimodal subtitle systems ViDove~\citep{lu2025vidove} and Hermes~\citep{cui2026hermes}. Full results are provided in Appendix~\ref{app:multimodal}.

\textbf{Temporal Generalization.} On 200 TV series released in 2025--2026, including titles beyond the reported knowledge cutoff of Claude Sonnet 4.6, SMART outperforms TransAgent across all six evaluated directions. This suggests that its gains extend to temporally shifted content rather than being confined to Subtitle Arena. Details are in Appendix~\ref{sec:temporal-generalization}.

\textbf{Qualitative Analysis.} Rendered examples further illustrate SMART's advantages on context-sensitive translation errors. Details are in  Appendix~\ref{app:rendered-frames}.

\subsection{Backbone and Evaluator Robustness}
\label{sec:backbone-evaluator}

Table~\ref{tab:backbone_judge_main} separates the effect of SMART's scaffold from backbone strength on representative bidirectional pairs. For en$\rightarrow$zh, SMART reduces the Overall penalty from $1.46$ to $0.67$ with Gemma 3 4B and from $0.93$ to $0.53$ with DeepSeek-V3.2; similar reductions hold in the reverse direction and across the other pairs. These results indicate that SMART's gains are not solely due to using a stronger base model. We further replace the default Claude Sonnet 4.6 judge with GPT-5.5 while keeping translation outputs fixed. Overall penalties change only marginally, e.g., $0.48$ versus $0.47$ on en$\rightarrow$zh and $0.87$ versus $0.87$ on en$\rightarrow$de, suggesting that the evaluation is robust to the choice of judge. Full backbone and judge results across all 15 bidirectional language pairs are in Appendix~\ref{app:backbone-judge-full}.

\subsection{Ablation Study}
\label{sec:ablation}

\begin{wraptable}{r}{0.65\textwidth}
\vspace{-12pt}
\centering
\caption{
Component ablation of SMART across six directions. Values are Overall SubMQM penalties (lower is better); parentheses show increases over the full system.
}
\label{tab:ablation}
\vspace{-10pt}
\setlength{\tabcolsep}{2.3pt}
\renewcommand{\arraystretch}{1}

\resizebox{\linewidth}{!}{%
\begin{tabular}{lcccccc}
\toprule

\rowcolor{tablegray}
\textbf{Setting} &
\textbf{en$\rightarrow$zh} &
\textbf{en$\rightarrow$de} &
\textbf{en$\rightarrow$ko} &
\textbf{en$\rightarrow$it} &
\textbf{en$\rightarrow$es} &
\textbf{en$\rightarrow$fr} \\
\midrule

\rowcolor{gray!4}
\textbf{Full system}
& \bestcell{0.48}
& \bestcell{0.88}
& \bestcell{0.93}
& \bestcell{1.13}
& \bestcell{1.18}
& \bestcell{1.19} \\

\midrule

w/o Dynamic Router
& \secondcell{0.70} (+0.22)
& \secondcell{1.12} (+0.24)
& \secondcell{1.18} (+0.25)
& 1.47 (+0.34)
& 1.46 (+0.28)
& 1.48 (+0.29) \\

w/o Self-Evolution
& 0.80 (+0.32)
& 1.22 (+0.34)
& 1.28 (+0.35)
& 1.50 (+0.37)
& 1.55 (+0.37)
& 1.57 (+0.38) \\

w/o Memory
& 1.43 (+0.95)
& 1.78 (+0.90)
& 1.69 (+0.76)
& 1.67 (+0.54)
& 1.74 (+0.56)
& 2.13 (+0.94) \\

w/o Contextual Retrieval \& Idiom Bank
& 1.24 (+0.76)
& 1.56 (+0.68)
& 1.31 (+0.38)
& 1.49 (+0.36)
& 1.44 (+0.26)
& 2.08 (+0.89) \\

w/o Sliding Window
& 0.93 (+0.45)
& 1.41 (+0.53)
& 1.52 (+0.59)
& 1.52 (+0.39)
& 1.58 (+0.40)
& 1.64 (+0.45) \\

w/o MoA
& 1.21 (+0.73)
& 1.66 (+0.78)
& 1.66 (+0.73)
& 1.63 (+0.50)
& 1.61 (+0.43)
& 1.70 (+0.51) \\

w/ Combined Prompt
& 1.06 (+0.58)
& 1.53 (+0.65)
& 1.47 (+0.54)
& \secondcell{1.32} (+0.19)
& \secondcell{1.41} (+0.23)
& \secondcell{1.44} (+0.25) \\

\bottomrule
\end{tabular}%
}

\vspace{-12pt}
\end{wraptable}

Table~\ref{tab:ablation} isolates the contribution of SMART's major components
across six representative directions. Series-level memory has the largest
effect, increasing the Overall penalty by $0.78$ on average when removed,
followed by MoA ($+0.61$), contextual retrieval and the idiom bank ($+0.56$),
and sliding-window consistency ($+0.47$). These results highlight the importance
of maintaining long-range context, retrieving relevant knowledge, and
reconciling multiple specialized translation hypotheses.

Dynamic routing and self-evolution provide consistent gains, with
penalty increases of $0.27$ and $0.36$ when removed. Replacing MoA with a
single translator using a \emph{Combined Prompt} increases the penalty by
$0.41$, indicating that MoA benefits from independently generated specialist
hypotheses rather than combining their instructions in one prompt.
Appendix~\ref{app:ablation-settings} defines all ablation settings.

\subsection{Human Evaluation}
\label{sec:human-evaluation}

\begin{wraptable}{r}{0.44\linewidth}
\vspace{-12pt}
\centering
\scriptsize
\caption{
Human evaluation by average rank ($5$ best, $1$ worst). Best and second-best are highlighted in blue-gray and warm beige.
}
\vspace{-10pt}
\label{tab:human_eval}
\setlength{\tabcolsep}{2.2pt}
\renewcommand{\arraystretch}{1}

\resizebox{\linewidth}{!}{%
\begin{tabular}{lccccc}
\toprule
\rowcolor{tablegray}
\textbf{Method} &
\textbf{Fidelity} &
\textbf{Consistency} &
\textbf{Language} &
\textbf{Subtitle} &
\textbf{Overall} \\
\midrule

Online
& 1.10
& 1.10
& 1.05
& 1.10
& 1.05 \\

Claude Sonnet 4.6
& \secondcell{3.65}
& 3.35
& \secondcell{3.80}
& \secondcell{3.90}
& \secondcell{3.75} \\

GPT-5.5
& 2.45
& 2.30
& 2.45
& 2.30
& 2.30 \\

TransAgent
& 3.35
& \secondcell{3.70}
& 3.30
& 3.25
& 3.40 \\

\rowcolor{gray!4}
\textbf{SMART}
& \bestcell{4.45}
& \bestcell{4.55}
& \bestcell{4.40}
& \bestcell{4.45}
& \bestcell{4.50} \\

\bottomrule
\end{tabular}%
}

\vspace{2pt}
\parbox{\linewidth}{\scriptsize
\textbf{Criteria.}
Fidelity = meaning preservation;
Consistency = terminology and cross-segment consistency;
Language = fluency, naturalness, and style;
Subtitle = readability and subtitle-specific presentation.
}
\vspace{-15pt}
\end{wraptable}

To complement SubMQM with viewer-facing assessment, we conduct a blind human
study on $20$ continuous subtitle passages comprising $267$ aligned lines from
$13$ television series. Twenty annotators, given the corresponding video and
English source, rank five anonymized translations from Online, Claude Sonnet
4.6, GPT-5.5, TransAgent, and SMART on four criteria: Fidelity, Consistency,
Language Quality, and Subtitle Quality. Candidate order is randomized, and
ranks run from $5$ (best) to $1$ (worst). Instructions and the evaluation
interface are in Appendix~\ref{app:human-evaluation-details}.

As shown in Table~\ref{tab:human_eval}, SMART ranks first on all four criteria and achieves the highest Overall score of $4.50$, with its best result on Consistency ($4.55$). Given the study's limited scale and coarse criteria, we use human evaluation to validate viewer-facing quality and retain SubMQM for fine-grained error analysis.

\subsection{Qualitative Visualization}
\label{sec:visualization}

\begin{CJK}{UTF8}{gkai}

\begin{figure}[ht]
    \centering
    \includegraphics[width=\textwidth]{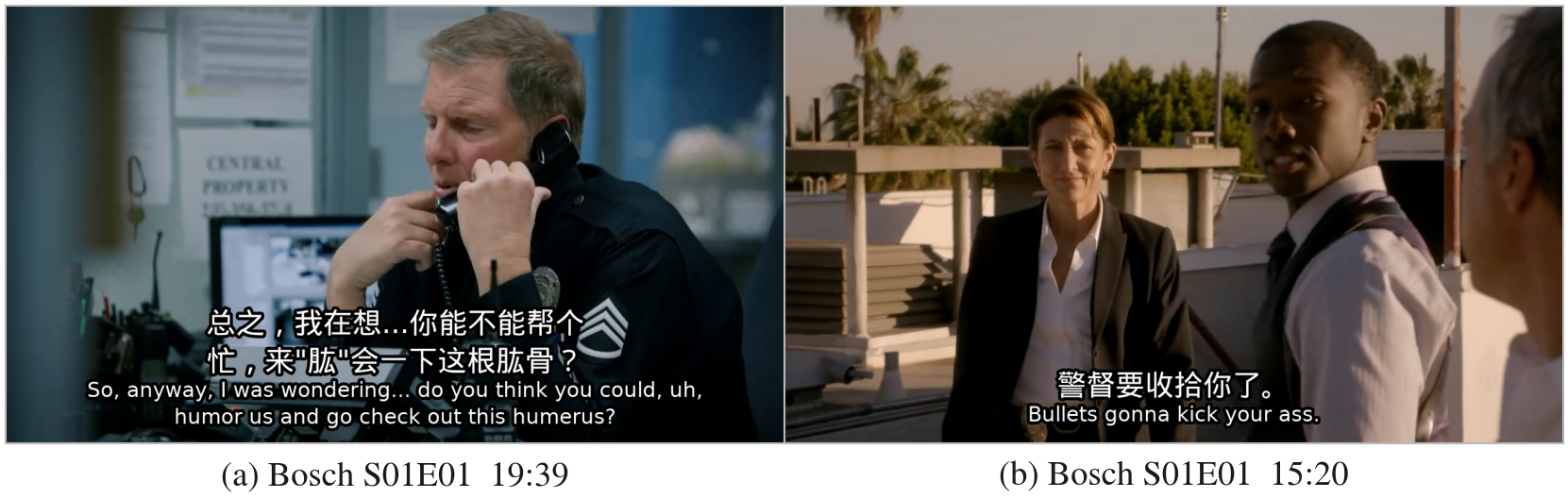}
    \caption{
    \textbf{Qualitative examples of context-aware subtitle translation.}
    (a) SMART preserves the wordplay between ``humor'' and ``humerus'' through
    a localized Chinese rendering involving ``肱骨''.
    (b) External web knowledge identifies ``Bullets'' as the nickname of
    Lt.~Grace Billets, allowing SMART to translate the utterance as a reference
    to the officer rather than the literal noun ``bullets''.
    }
    \label{fig:wordplay}
\end{figure}

Fig.~\ref{fig:wordplay} visualizes two cases where SMART goes beyond
literal segment-level translation by exploiting linguistic reasoning and
external knowledge.

\textbf{Wordplay preservation.}
In Fig.~\ref{fig:wordplay}(a), the source line ``humor us and go check
out this humerus'' plays on the phonetic similarity between \emph{humor} and
\emph{humerus}. A literal translation would lose this effect. SMART instead
reconstructs the joke in Chinese as ``帮个忙，来‘肱’会一下这根肱骨？'', introducing
a localized play around ``肱'' and ``肱骨'' while preserving the underlying
request. This illustrates that the system can adapt non-literal expressions
rather than translating their surface forms independently.

\textbf{Knowledge-grounded entity disambiguation.}
Figure~\ref{fig:wordplay}(b) shows a case where the line ``Bullets gonna
kick your ass'' is ambiguous without series-specific knowledge: \emph{Bullets}
can naturally be interpreted as the common noun ``bullets.'' Through web search,
SMART identifies ``Bullets'' as the nickname of Lt.~Grace Billets, a commanding
officer in the LAPD Hollywood Division. It consequently renders the line as
``警督要收拾你了'', referring to the character rather than literally translating
``Bullets'' as ``子弹''. This example demonstrates how external knowledge can
resolve entity references that are difficult to recover from the local subtitle
segment alone.

\end{CJK}


\section{Conclusion}
We introduced \textbf{SMART}, a self-evolving multi-agent system for long-form subtitle translation that combines persistent series-level memory, adaptive routing, tool-augmented translation, and test-time evolution without updating model parameters. We also introduced \textbf{Subtitle Arena} and \textbf{SubMQM} for evaluating long-form subtitle translation across languages and quality dimensions. Experiments show that SMART consistently improves over strong baselines, suggesting that long-form translation benefits from treating translation as a stateful and adaptive process rather than a sequence of independent sentence-level decisions.

\newpage
\appendix
\section{Extended Related Work}
\label{app:related-work}

\textbf{Multi-Agent Systems and Translation Agents.}
Multi-agent LLM systems distribute reasoning across agents with specialized roles, enabling decomposition, cross-agent verification, and iterative refinement~\citep{li2023camel, MAS-9423979, MAS-ijcai2024p890, MAS-cheng2024exploringlargelanguagemodel}.
In machine translation, existing approaches can broadly be grouped into role-based collaboration and context-aware translation.
MAPS~\citep{he2024maps}, TransAgents~\citep{wu-etal-2024-transagents}, TEaR~\citep{feng2025tear}, and DRT~\citep{wang2025drt} decompose translation into complementary processes such as knowledge elicitation, generation, estimation, criticism, and refinement.
TACTIC~\citep{li2025tactic} and MAATS~\citep{wang2025maats} further specialize agents for contextual reasoning, external knowledge, and individual MQM error categories.
For long-context translation, DelTA~\citep{MAS-mt-wang2025deltaonlinedocumentleveltranslation} maintains multi-level memory for document consistency, while ViDove~\citep{lu2025vidove} incorporates visual, auditory, and historical context for video subtitle translation.
ALPO~\citep{cui2026utterance} takes a different direction and improves expressive subtitle translation through adaptive preference optimization.
Although these methods substantially improve translation quality and contextual consistency, their workflows, agent roles, or learned policies are generally fixed once deployed.

\textbf{Self-Evolving LLM Agents.}
Self-evolving agents seek to improve agent behavior through continued interaction.
One line of work evolves the structure and execution flow of agent systems, including ADAS~\citep{hu2025automateddesignagenticsystems}, AFlow~\citep{zhang2025aflowautomatingagenticworkflow}, EvoAgentX~\citep{wang2025evoagentxautomatedframeworkevolving}, EvoFlow~\citep{zhang2025evoflowevolvingdiverseagentic}, and SEW~\citep{liu2026sewselfevolvingagenticworkflows}.
Other methods accumulate reusable skills, tools, or experience~\citep{wang2023voyageropenendedembodiedagent, zhao2024expelllmagentsexperiential, nguyen2025dynasaurlargelanguageagents, zhang2026memskilllearningevolvingmemory, acikgoz2026toolr0selfevolvingllmagents, chen2026skillcraftllmagentslearn}, improve model reasoning through self-generated supervision~\citep{zelikman2022starbootstrappingreasoningreasoning, zeng2025bstarmonitoringbalancingexploration, tian2024selfimprovementllmsimaginationsearching}, or evolve persistent memory at inference time~\citep{suzgun2025dynamiccheatsheettesttimelearning, wei2025evomemorybenchmarkingllmagent}.
Together, these approaches establish workflows, capabilities, models, and memory as complementary targets for continual adaptation~\citep{gao2025survey, fang2025comprehensivesurveyselfevolvingai}.
However, prior work mainly evaluates whether adaptation improves subsequent tasks, with less attention to whether newly acquired behaviors interfere with previously learned ones.
For long-form translation, this distinction is important because improvements on later sentences should not destabilize terminology, register, or translation conventions established earlier.

\textbf{Prompt Optimization.}
Automated prompt optimization adapts LLM behavior without updating model parameters.
Earlier methods formulate prompt construction as discrete search or reinforcement learning~\citep{shin2020autoprompt, deng2022rlprompt}, while later approaches use LLMs themselves to generate, mutate, and select prompts~\citep{zhou2022large, guo2023connecting, fernando2023promptbreeder, khattab2023dspy}.
Feedback-driven methods are particularly related to SMART.
APO~\citep{pryzant2023automatic} interprets critiques as textual gradients, TextGrad~\citep{yuksekgonul2024textgrad} propagates textual feedback through computational graphs, and REVOLVE~\citep{zhang2024revolve}, SIPDO~\citep{yu2025sipdo}, and GEPA~\citep{agrawal2025gepa} iteratively refine prompts through reflection and generated feedback.
These methods demonstrate that natural-language feedback can effectively improve instructions, but typically optimize a prompt against a fixed task or objective.
SMART instead maintains agent instructions as persistent state that evolves throughout an ongoing translation process.

\textbf{LLM-as-a-Judge.}
LLMs have increasingly been adopted as scalable evaluators of model-generated outputs.
General LLM-as-a-judge methods use strong language models to score, rank, or critique responses, while also exposing evaluation biases such as position or verbosity preferences~\citep{zheng2023judging}.
This paradigm is especially useful for machine translation, where translation errors cannot always be summarized by a single lexical similarity score.
GEMBA-MQM~\citep{kocmi2023gemba} and AutoMQM~\citep{fernandes2023devil} use LLMs to provide structured MQM-style judgments over error spans, categories, and severities.
Such feedback offers richer diagnostic information than scalar quality scores and can support targeted correction.
Existing work, however, primarily treats the judge as an evaluation endpoint or as feedback for local refinement.
SMART additionally converts translation-specific judgments into persistent updates of the translation system, coupling evaluation with continual test-time training.

\begin{CJK}{UTF8}{gkai}
\section{SMART Implementation Details}
\label{app:method-details}

\subsection{Subtitle Parsing and Context Construction}
Each SRT block is parsed into its index, start and end timestamps, and source text. SMART augments the current block with neighboring source lines, scene metadata, and the series content profile. Consecutive blocks that appear to form a single interrupted sentence are grouped into one sentence and are split back to their original timestamp boundaries after translation. This avoids forcing the model to translate each block independently when its semantics are sentence-level.

\subsection{Memory Construction and Content Profiling}
The terminology pool stores source terms together with confirmed target renderings and occurrence statistics. Character memory contains names, target-language renderings when confidently established, roles. Domain memory stores series- and scene-level facts required to interpret technical language, cultural references, and ambiguous slang. The idiom bank stores target-language expressions indexed by communicative situation and genre.

At the beginning of a series, SMART first builds an initial profile of genre, setting, principal characters, and likely domain terminology. During translation, uncertain expressions can trigger targeted retrieval by web search. Search results are passed through a contextual interpretation step before being exposed to translators, so retrieved literal meanings are not automatically treated as appropriate subtitle renderings. When a series already has persistent memory, existing information is reused and only missing entries are queried.

\subsection{Routing Policy and Translator Roles}\label{routing}
The router uses features including source length, subtitle duration, scene tone, emotional markers, slang/profanity indicators, and display-budget pressure. Inputs are mapped to a small number of routing categories, and each category activates a subset of the translator pool. The default path uses the faithful and natural translators; length-constrained sentences may additionally activate the length-aware translator, emotionally marked dialogue may activate the expressive translator, and slang-heavy dialogue may activate the colloquial translator. During training, routing outcomes are logged together with judge scores and become evidence for updating $\rho_t$.

The translator roles are:
\begin{itemize}
    \item \textbf{Faithful translator}: prioritizes semantic accuracy and preservation of established entities and domain terms.
    \item \textbf{Natural translator}: prioritizes target-language fluency and register-appropriate phrasing.
    \item \textbf{Expressive translator}: prioritizes character narrative voice, emotion, and dramatic force when these are salient.
    \item \textbf{Length-aware translator}: explicitly conditions on subtitle duration and display budgets.
    \item \textbf{Colloquial translator}: handles slang, profanity, and conversational language without over-literal or over-sanitized rendering.
\end{itemize}

\subsection{Tool Interfaces}
\label{app:tool-reference}
SMART exposes eight callable tools. \texttt{terminology\_lookup} retrieves established renderings, while \texttt{terminology\_register} proposes new entries with conflict detection. \texttt{memory\_search} retrieves similar prior subtitle units and their translations. \texttt{get\_context} returns surrounding dialogue, existing translations, scene information, and domain notes. \texttt{constraint\_check} measures line length and reading-speed violations. \texttt{web\_search} retrieves external evidence for jargon, cultural references, and unclear slang. \texttt{idiom\_lookup} returns situation-appropriate target-language idioms from the series bank or a fallback query. \texttt{fluency\_check} flags translationese, over-colloquialization, and register mismatch. Tool calls and outputs are recorded during training and can inform subsequent routing updates.

Table~\ref{tab:tool-reference} lists the purpose and execution of each. All tools run locally except \texttt{web\_search}, which issues an external query and passes the raw result through the contextual interpretation step of Appendix~\ref{app:web-search-example} before returning it. The router fixes which subset of tools each activated translator may call, and this assignment is itself updated during self-evolution (Appendix~\ref{app:routing-update-example}).

\begin{table*}[t]
\centering
\caption{Tool modules available to SMART translators. Execution is local unless otherwise noted.}
\label{tab:tool-reference}
\small
\setlength{\tabcolsep}{5pt}
\renewcommand{\arraystretch}{1.15}
\begin{tabular}{
    >{\raggedright\arraybackslash}p{0.28\textwidth}
    >{\raggedright\arraybackslash}p{0.28\textwidth}
    >{\raggedright\arraybackslash}p{\dimexpr0.44\textwidth-6\tabcolsep\relax}
}
\toprule
\rowcolor{tablegray}
\textbf{Tool} & \textbf{Purpose} & \textbf{Execution} \\
\midrule
\texttt{terminology\_lookup} & Retrieve the established target rendering of a recurring name or term. & Match against the shared terminology table and return the confirmed translation or partial matches. \\
\texttt{terminology\_register} & Fix the target rendering of a new name or term. & Write to the terminology table with conflict detection to prevent overwriting an existing entry. \\
\texttt{memory\_search} & Find previously translated sentences with similar source content. & Rank entries from the current episode's translation memory by source overlap to support consistency and style. \\
\texttt{get\_context} & Retrieve surrounding dialogue, scene information, and domain notes. & Read up to eight preceding and succeeding sentences together with the associated scene metadata. \\
\texttt{constraint\_check} & Verify subtitle reading-speed and line-length constraints. & Compute characters per second and per-line character counts, and report specific violations. \\
\texttt{web\_search} & Resolve jargon, cultural references, or ambiguous slang. & \emph{External} search whose raw results are interpreted to separate literal meaning from the register appropriate to the scene. \\
\texttt{idiom\_lookup} & Retrieve target-language idioms and colloquial expressions. & Query a content-specific idiom bank, with a dynamic lookup when the requested category is unavailable. \\
\texttt{fluency\_check} & Detect translationese, excessive colloquialization, and register mismatch. & Evaluate the target-language draft and return localized revision suggestions. \\
\bottomrule
\end{tabular}
\end{table*}

\subsection{Judge--Refiner and Rescue Step}
The judge evaluates candidate translations using semantic accuracy, fluency, register, expressiveness when relevant, terminology consistency, and subtitle constraints, and reports an overall score on a $0$--$10$ scale. The top-scoring candidate is passed to the refiner together with its critique. If that score falls below the rescue threshold $\theta=8.0$, SMART performs a rescue pass by re-running the MoA translator with the judge critique appended and comparing the new output against the existing candidates. The accepted translation updates the local translation history and, when appropriate, the persistent series memory.

\subsection{Episode-Level Consistency Pass}\label{Episode}
After sentence-level translation, SMART performs an episode-level consistency pass over overlapping windows of increasing size. Each window is conditioned on locked preceding translations to reduce drift. The editor checks recurring terminology, dialogue coherence, pronoun and reference consistency, and residual translationese. Later rounds are skipped when an earlier round makes no changes.

\subsection{Test-Time Training Loop}
\label{app:test-time-training-loop}
Each series is partitioned chronologically using a $3{:}7$ training/inference ratio. Prompt and routing updates are allowed only on the training prefix. The evolved prompts and router are frozen on the held-out portion, while translation memory and series memory continue to update because they represent task state rather than learned model parameters.

The self-evolution procedure described in the main text proceeds over epochs of the training subset $\mathcal{D}_{\text{tr}}$. Each epoch runs the same per-sentence pipeline as inference, then evaluates the collected scores and updates the prompts and routing policy. The learnable state is the set of agent prompts, the routing policy, and the tool assignments, all expressed as text or configuration rather than model weights. The five phases of one epoch are as follows.

\begin{center}
\begin{minipage}{0.96\linewidth}
\footnotesize\ttfamily
\begin{verbatim}
Phase 1  TRANSLATE
  translate the training files with the current config
  collect per segment: judge score, agent used, tools called, category

Phase 2  EVALUATE
  aggregate into statistics:
    per agent    : mean/min/max score, tool usage counts
    per category : which agent scores highest
    worst cases  : lowest-scoring sentences with critiques

Phase 3  PROMPT OPTIMIZATION  (F^Pi)
  for each agent with mean score below a threshold:
    feed current prompt + statistics + worst cases to the evaluator
    the evaluator rewrites the prompt to address the failure modes

Phase 4  STRUCTURE OPTIMIZATION  (F^rho)
  feed routing policy + per-category performance to the evaluator
  update: drop underperforming agents from a category,
          reassign tools, keep at least two agents per category

Phase 5  SAVE CONFIG
  persist evolved prompts, routing policy, tool assignments, stats
  this config is loaded unchanged at inference time
\end{verbatim}
\end{minipage}
\end{center}

The loop terminates after a fixed number of epochs, and the configuration from the final epoch is the $(\Pi_T, \rho_T)$ applied during test-time inference. Because every update is a text rewrite driven by aggregated judge feedback, the procedure requires no gradient computation and leaves the underlying LLM unchanged.

\subsection{Instantiation of $F^{\Pi}$ and $F^{\rho}$}
\label{app:evolution-operators}

Both operators are single-proposal rewrites over constrained spaces, applied once per training epoch from statistics aggregated over the entire training prefix. 
The series memory $M$ is initialized once for each series before the first training epoch and is carried across epochs, so prompt and routing updates build on the same persistent task state rather than reconstructing series knowledge from scratch. Algorithm~\ref{alg:self-evolution} states the full loop; the paragraphs below give the parts that the pseudocode abbreviates.

\begin{algorithm}[t]
\caption{Self-evolution at test-time training for one series. Lines~\ref{line:fpi-start}--\ref{line:fpi-end} are $F^{\Pi}$; line~\ref{line:frho} is $F^{\rho}$. Both act once per epoch on statistics aggregated over the whole training prefix.}
\label{alg:self-evolution}
\small
\begin{algorithmic}[1]
\Require training prefix $\mathcal{D}_{\mathrm{tr}}$ (first $30\%$ of the series); translator pool $\mathcal{R}$; tool set $\mathcal{T}$; judge $J$; refiner $R$; evaluator $E$
\Require initial prompts $\Pi_0$, initial routing table $\rho_0$, epochs $T$, agent threshold $\tau_{\text{agent}}$, rescue threshold $\theta$, worst-case budget $k$
\Ensure frozen configuration $(\Pi_T, \rho_T)$
\State $M \gets \emptyset$ \Comment{series memory}
\For{$t = 0$ \textbf{to} $T-1$}
  \State $\mathcal{L} \gets \emptyset$ \Comment{Phase 1: translate}
  \For{each episode $e \in \mathcal{D}_{\mathrm{tr}}$}
    \For{each sentence $d_i \in e$}
      \State $\mathcal{A}_i \gets \rho_t\!\left(\phi(d_i, M)\right)$ \Comment{category lookup, no learned selector}
      \State $y_i^{(r)} \gets r\!\left(d_i, \pi_{r,t}, M, \mathcal{T}\right)$ for each $r \in \mathcal{A}_i$
      \State $\left(c_i^{(r)}, \delta_i^{(r)}\right) \gets J\!\left(y_i^{(r)}, d_i, M\right)$ for each $r \in \mathcal{A}_i$
      \State $r^\star \gets \arg\max_{r \in \mathcal{A}_i} c_i^{(r)}$
      \If{$c_i^{(r^\star)} < \theta$}
        \State rerun the faithful translator with $\delta_i^{(r^\star)}$ appended; keep the better candidate
      \EndIf
      \State $\hat{y}_i \gets R\!\left(y_i^{(r^\star)}, \delta_i^{(r^\star)}, M\right)$; \; $M \gets \textsc{UpdateMemory}(M, \hat{y}_i)$
      \State append $\left(\text{category}(d_i), \mathcal{A}_i, r^\star, \{c_i^{(r)}, \delta_i^{(r)}\}, \text{tools called}\right)$ to $\mathcal{L}$
    \EndFor
    \State episode-level consistency pass over overlapping windows
  \EndFor
  \State $\left(S_{\text{agent}}, S_{\text{cat}}, W\right) \gets E(\mathcal{L})$ \Comment{Phase 2: $\Delta_t$; $W$ holds $k$ worst sentences per agent}
  \ForAll{$r \in \mathcal{R}$} \Comment{Phase 3: $F^{\Pi}$}
    \label{line:fpi-start}
    \If{$\mathrm{mean}\!\left(S_{\text{agent}}[r]\right) < \tau_{\text{agent}}$}
      \State $\pi_{r,t+1} \gets \textsc{Rewrite}\!\left(\pi_{r,t}, S_{\text{agent}}[r], W_r\right)$
      \State \quad \textbf{s.t.} role statement and workflow steps preserved
    \Else
      \State $\pi_{r,t+1} \gets \pi_{r,t}$ \Comment{at or above threshold: untouched}
      \label{line:fpi-end}
    \EndIf
  \EndFor
  \State $\rho_{t+1} \gets \textsc{Rewrite}\!\left(\rho_t, S_{\text{cat}}\right)$ \Comment{Phase 4: $F^{\rho}$}
    \label{line:frho}
  \State \quad \textbf{s.t.} $\left|\rho_{t+1}(c)\right| \ge 2$ for every category $c$; category set fixed
  \State persist $\left(\Pi_{t+1}, \rho_{t+1}, S_{\text{agent}}, S_{\text{cat}}\right)$ \Comment{Phase 5}
\EndFor
\State \Return $(\Pi_T, \rho_T)$ \Comment{final epoch; no acceptance test, rollback, or early stopping}
\end{algorithmic}
\end{algorithm}

\textbf{$F^{\Pi}$: prompt rewrite.} The evaluator uses the aggregated per-agent statistics and low-scoring cases from Phase~2 to identify agents with recurring failure modes; prompts for agents not flagged by this feedback are left unchanged. For each flagged agent, the operator receives the current prompt, its aggregate statistics, and representative low-scoring sentences together with their critiques, and produces one rewritten prompt. The rewrite is constrained: it must preserve the agent's role statement and the workflow skeleton, and may add or amend rules but may not redefine the role or delete a workflow step. The critiques describe failure modes rather than supplying corrected outputs (Appendix~\ref{app:prompt-update-example}), so the rewrite cannot encode sentence-specific fixes.

\textbf{$F^{\rho}$: table rewrite.} Its input is the current routing table, the per-category record of which agents achieved the highest judge scores, and tool-usage counts; its output is a new table. Two constraints bound it: every category retains at least two agents, and the category set is fixed, so the operator reassigns agents and tools within a fixed schema rather than inventing categories. The reachable space is therefore finite and enumerable.

A proposal satisfying the structural constraints above is adopted; there is no separate validation-gated acceptance test。
The loop runs a fixed number of epochs and the final epoch's configuration becomes $(\Pi_T, \rho_T)$. We run self-evolution for three epochs over $\mathcal{D}_{\mathrm{tr}}$.
After the third epoch, the resulting prompts, routing policy, and tool assignments are frozen and used for test-time inference on
$\mathcal{D}_{\mathrm{inf}}$.


\section{Worked Examples}
\label{app:worked-examples}

The examples below trace SMART on episodes of the crime drama \emph{Bosch}, translating from English to Chinese. They are illustrative of one language direction; the same mechanisms apply to every target language in \textbf{Subtitle Arena}. We show a full sentence traversal of the per-sentence pipeline, then isolate the two components, the web-search contextual retrieval path and the cross-episode series memory.

\subsection{End-to-End Sentence Example}
\label{app:end-to-end-example}

Consider a sentence from a scene in which one detective needles another about a stalled case. The router extracts surface features, classifies the sentence, and activates the corresponding subset of translators; each activated translator runs its own tool loop before emitting a candidate; the judge scores all candidates and the refiner finalizes the output.

\begin{center}
\begin{minipage}{0.96\linewidth}
\footnotesize\ttfamily
\begin{verbatim}
Input sentence
  index    : 128
  time     : 00:12:04,320 --> 00:12:06,880   (duration 2.56s)
  source   : Bosch, you're chasing your own tail here.

Step 1  Routing
  features : {word_count: 8, duration: 2.56, has_slang: false,
              has_emotion: false, scene_tone: "tense"}
  category : expressive         (scene tone is tense)
  agents   : [faithful, natural, expressive]

Step 2  Agent execution (each agent runs its own tool loop)
  [faithful]
    tool_use  terminology_lookup {"term": "Bosch"}
    result    {found: true, target: 博斯, category: name}
    tool_use  constraint_check {"text": 博斯，你这是在原地打转。,
                                "duration": 2.56}
    result    {valid: true, cjk_chars: 11, max_chars: 38, cps: 4.3}
    output    博斯，你这是在原地打转。
  [natural]
    output    博斯，你这么查是在白费劲。
  [expressive]
    tool_use  idiom_lookup {"category": "futile_effort"}
    result    {idioms: [徒劳无功, 缘木求鱼, 竹篮打水], ...}
    output    博斯，你这是缘木求鱼。

Step 3  Judge (scores abbreviated)
  1 [faithful]   accuracy 9 fluency 8 register 8 ... overall 8.2
  2 [natural]    accuracy 9 fluency 9 register 9 ... overall 8.8
  3 [expressive] accuracy 7 fluency 9 register 7 ... overall 7.9
     critique(3): idiom overstates the register for a terse jab
  winner : [natural] 8.8

Step 4  Refiner + state update
  score 8.8 >= threshold, output kept as is
  final  : 博斯，你这么查是在白费劲。
  memory : translation memory += (source, final)
           terminology "Bosch"->"博斯" count += 1
\end{verbatim}
\end{minipage}
\end{center}

The example shows the two properties the router is designed for: only the relevant translators are activated (the colloquial and length-aware translators are not needed here), and the terminology tool supplies the confirmed rendering of the recurring name so that the candidate does not re-derive it.

\subsection{Web Search and Contextual Retrieval Example}
\label{app:web-search-example}

Police procedurals contain jargon whose literal reading differs from its conversational intent. When a translator encounters such a term, it calls \texttt{web\_search}; the raw result is passed through a contextual interpretation step that separates the literal domain meaning from the register the scene calls for, and the outcome is written into the scene domain notes so that later sentences and the judge see the same context.

\begin{center}
\begin{minipage}{0.96\linewidth}
\footnotesize\ttfamily
\begin{verbatim}
Segment
  source : We got a 187 in Hollywood, roll on it.
  scene  : squad room, tone = urgent

web_search {"query": "What does 187 mean in US police radio code?",
            "context": "detective assigning a case, urgent tone"}

raw result
  California Penal Code section 187 defines the crime of murder;
  "a 187" is police radio shorthand for a homicide.

=== translation guidance ===
  literal : 187 is the penal-code number for homicide/murder.
  context : squad-room dispatch, urgent; use the natural term a
            Chinese audience expects, not the code number.
  advice  : render as 凶杀案/命案, not the literal number "187".

scene.domain_notes += "[187] homicide radio code; render as 凶杀案"

final translation
  好莱坞发生一起凶杀案，马上去。
\end{verbatim}
\end{minipage}
\end{center}

Without retrieval, a literal system emits the number 187, which is meaningless to the audience. The contextual interpretation step is what prevents the opposite failure of injecting the technical penal-code register into casual speech; it reports the literal meaning and the register separately, and the domain note it writes is reused by every subsequent sentence in the scene.

\subsection{Series Memory Example}
\label{app:series-memory-example}

The series memory persists terminology, character profiles, domain facts, and the idiom bank across episodes of the same series. A name fixed in an early episode is retrieved directly in later episodes, however far apart the occurrences are, which is the mechanism behind the cross-episode consistency the benchmark measures.

\begin{center}
\begin{minipage}{0.96\linewidth}
\footnotesize\ttfamily
\begin{verbatim}
Episode S01E01  (memory starts empty)
  content profiling registers principal characters:
    "Harry Bosch" -> 哈里·博斯
    "J. Edgar"    -> 杰瑞·埃德加
  domain facts: LAPD Hollywood Division homicide unit
  at episode end, series memory is persisted:
    terminology : "Harry Bosch" -> {target: 哈里·博斯, count: 14}
                  "J. Edgar"    -> {target: 杰瑞·埃德加, count: 6}
    characters  : Harry Bosch (detective), J. Edgar, ...
    idiom_bank  : crime_investigation, interrogation, ...

Episode S01E05  (series memory loaded before translation)
  inject_into_executor : 23 terms preloaded into terminology table
  inject_into_profile  : characters, domain facts, idiom bank reused
                         (idiom bank rebuild skipped)

  segment source : Edgar, get Bosch on the phone.
    terminology_lookup "Bosch"  -> found, target = 博斯
    terminology_lookup "Edgar"  -> found, target = 埃德加
    final : 埃德加，给博斯打电话。

  merge back at episode end:
    new terms confirmed this episode are added to memory
    counts of existing terms accumulated (Harry Bosch: 14 -> 27)
\end{verbatim}
\end{minipage}
\end{center}

Because the mapping \texttt{Harry Bosch} → \texttt{哈里·博斯} is loaded from memory rather than re-inferred, the fifth episode reuses the exact rendering fixed in the first, and the accumulated occurrence counts let the judge treat the established translation as the consistency reference for the remainder of the series.

\subsection{Agent Reasoning Trace}
\label{app:agent-trace}

The examples above summarize which tools were called; here we show the interleaved reasoning that produces those calls. Each translator follows a reason-act loop: it states what it is uncertain about, calls a tool to resolve that uncertainty, reads the result, and decides the next step, until it commits to a translation. The trace below is from the \emph{natural} translator on a sentence whose difficulty is not the vocabulary but the register: a line that reads as a factual statement but is meant as a sarcastic jab.

\begin{center}
\begin{minipage}{0.96\linewidth}
\footnotesize\ttfamily
\begin{verbatim}
Segment
  source   : Great, another dead end. Love this job.
  scene    : stakeout gone wrong, tone = frustrated/sarcastic

Agent: natural  (reasoning interleaved with tool calls)

  thought : Two sentences. "dead end" is an investigation term
            but here it is figurative. "Love this job" is not
            literal; the scene tone is frustrated, so it is
            sarcasm. A literal 我爱这份工作 would read as sincere
            and invert the meaning. I first confirm the tone.
  action  : get_context {"direction": "before", "count": 2}
  result  : prev lines show the lead just fell through;
            scene.tone = "frustrated/sarcastic"

  thought : Tone confirmed as sarcastic. I should not translate
            "Love this job" straight. I want a Chinese phrasing
            that carries the same dry sarcasm rather than a
            literal statement. Check for a fitting expression.
  action  : idiom_lookup {"category": "sarcasm/complaint"}
  result  : {colloquial: [这活儿真"好", 又白跑一趟, 干这行真"值"], ...}

  thought : "又白跑一趟" matches "another dead end" naturally.
            For "Love this job" I keep the sarcasm with a dry
            这工作真"带劲", the quotes marking irony. Now verify
            it reads naturally and is not over-colloquial.
  action  : fluency_check {"text": "又白跑一趟。这工作真"带劲"。",
                           "tone": "sarcastic"}
  result  : {fluent: true, register: ok,
             note: irony quotes read well, keep as is}

  thought : Fluency check passes. Length is short, no constraint
            risk. Commit.
  output  : 又白跑一趟。这工作真"带劲"。
\end{verbatim}
\end{minipage}
\end{center}

The trace shows why the reasoning matters: the hard decision is not lexical but pragmatic, and the agent reaches the sarcastic reading by first confirming the scene tone through \texttt{get\_context}, then choosing an ironic phrasing over the literal one and validating its register with \texttt{fluency\_check}. A system that translated the surface form would emit a sincere \texttt{我爱这份工作}, reversing the speaker's intent. This interleaved reason-act record is also what the judge and the self-evolution stage consume: the recorded thoughts and tool results make it possible to attribute a low score to a specific decision rather than to the sentence as a whole.

\subsection{Prompt Update Example}
\label{app:prompt-update-example}

During test-time training the evaluator aggregates judge scores over a batch, identifies recurring deficiencies attributable to a specific agent, and rewrites that agent's prompt through the operator $F^{\Pi}$ defined in the main text. The update is expressed entirely in natural language; no LLM weights change. The example below shows the faithful agent's prompt before and after one such update, together with the failure pattern that drove it.

\begin{center}
\begin{minipage}{0.96\linewidth}
\footnotesize\ttfamily
\begin{verbatim}
Prompt at step t  (faithful agent, expert)
  PRIORITY: semantic accuracy; preserve exact meaning.
  WORKFLOW:
   1. terminology_lookup for names/terms
   2. translate preserving full meaning
   3. constraint_check with translation and duration
   4. terminology_register for new names

Failure pattern found by the evaluator over the batch
   - interjections/onomatopoeia translated literally
     ("Oh, boy." -> 哦，孩子。),  mean score 4.2
   - short ambiguous lines translated without get_context
   - emotional register flattened in colloquial sentences

Textual critique delta_t (fed to F^Pi)
   "add rules for interjections, short-line context, and
    register preservation; keep the accuracy priority."

Prompt at step t+1  (after F^Pi, added rules)
  KEY RULES (new):
   - interjections (oh, huh, ugh): translate the FUNCTION,
     not the literal sound
   - short lines (<= 5 chars): call get_context FIRST
   - profanity/slang: match INTENSITY, do not sanitize
  WORKFLOW: (unchanged steps, get_context added at step 1)

Same input, before and after the update
  source : Oh, boy. Here we go again.   (scene tone = weary)
  before : 哦，孩子。我们又来了。           judge 4.1
  after  : 唉，又来了。                    judge 8.6
\end{verbatim}
\end{minipage}
\end{center}

The critique names the failure modes rather than supplying corrected outputs, so the rewritten prompt generalizes to unseen sentences with the same difficulty rather than memorizing specific fixes. The before/after pair is a sentence the updated prompt was never shown: the literal reading of the interjection and the redundant subject both disappear once the interjection rule is in force.

\subsection{Routing Policy Update Example}
\label{app:routing-update-example}

In parallel with the prompt update, the router policy is updated through $F^{\rho}$ from the per-category statistics of which agents achieved the highest judge scores. The example shows the policy and one agent's tool assignment before and after an update driven by the recorded scores.

\begin{center}
\begin{minipage}{0.96\linewidth}
\footnotesize\ttfamily
\begin{verbatim}
Routing policy at step t
  colloquial : [faithful, natural, colloquial]
  expressive : [faithful, natural, expressive]
  short      : [faithful, natural]
  long       : [faithful, natural, length_aware]
  default    : [faithful, natural]

Per-category statistics over the batch
  - faithful averages 6.2 on "colloquial" (too literal for slang)
  - natural averages 8.1 on "expressive", above expressive (7.9)

Routing policy at step t+1  (after F^rho)
  colloquial : [natural, colloquial]     (faithful dropped)
  expressive : [faithful, natural, expressive]
  short      : [faithful, expressive]    (emotion helps short lines)
  long       : [faithful, natural, length_aware]
  default    : [faithful, natural]

Tool assignment update (colloquial agent)
  + memory_search     (reuse prior slang renderings)
  + web_search        (resolve unfamiliar slang meaning)

Same input, before and after the update
  source : He's been jerking us around all day.  (category: colloquial)
  before : [faithful]   他一整天都在让我们绕圈子。    judge 6.2
  after  : [colloquial] 他耍了我们一整天。           judge 8.7
\end{verbatim}
\end{minipage}
\end{center}

The policy update removes an agent from a category where it consistently underperformed and promotes one that scored higher, while the tool-assignment update biases each agent toward the tools that helped on similar inputs. Both changes are recorded so that the evolved configuration $(\Pi_T, \rho_T)$ can be applied unchanged during inference. In the before/after pair, the winning candidate changes because the category no longer activates the agent that scored $6.2$ on it, not because any agent was retrained. 
\section{Subtitle Arena: Additional Details}
\label{app:subtitle-arena-details}

\subsection{Source Corpus and Episode Reconstruction}
Subtitle Arena is derived from the OpenSubtitles collection~\citep{tiedemann-luo-2026-opensubtitles2024}. For each target locale we read the English--Target sentence alignments released with that collection, so English is the pivot side of every bitext; both translation directions are then obtained by exchanging the roles of the two sides rather than by collecting a separate corpus for each direction. Each subtitle document is associated with an IMDb identifier and season/episode indices. We use the tuple $(\mathrm{IMDb},\mathrm{season},\mathrm{episode})$ as the canonical episode key and group the English and target-language versions of the same television episode under this identifier. This reconstruction converts sentence-level parallel data into episode-level subtitle streams while preserving the original narrative boundary.

\subsection{SRT Reconstruction and Filtering}
For each aligned sentence pair we recover the timestamp span and surface form on both sides from their respective subtitle documents and emit standard SRT blocks in presentation order, so that either side can serve as the source stream. Timestamp strings are normalized to the canonical \texttt{hh:mm:ss,ms} format. Empty or one-sided alignment groups are discarded, as are episodes whose target subtitle stream is missing or cannot be parsed.

Subtitle files may divide a single sentence across multiple adjacent display blocks. We therefore merge continuation blocks when reconstructing aligned sentences while retaining the corresponding temporal boundaries in the episode representation. 

\subsection{Locale Normalization}
Subtitle Arena reports $15$ target locales: Chinese (zh\_CN), Korean (ko\_KR), European and Brazilian Portuguese (pt\_PT, pt\_BR), Italian (it\_IT), Norwegian (no\_NO), Danish (da\_DK), Dutch (nl\_NL), Romanian (ro\_RO), Swedish (sv\_SE), Turkish (tr\_TR), European and Mexican Spanish (es\_ES, es\_MX), German (de\_DE), and French (fr\_FR). OpenSubtitles does not expose a distinct subtitle stream for every regional variety represented in our evaluation. 

\subsection{Benchmark Statistics and Translation Directions}
The final benchmark contains $192$ television series, $6{,}267$ English source episodes, and $70{,}664$ aligned bilingual episode pairs. Individual series contain between $2$ and $198$ episodes, with an average of $32.6$ episodes per series, and span production years from $1959$ to $2023$. The benchmark covers $14$ genres; Appendix~\ref{app:subtitle-arena-visualization} visualizes per-locale episode coverage and genre diversity.

Because every episode pair aligns English with one target locale, the $15$ locales yield $30$ evaluation directions: $15$ English$\rightarrow$locale and $15$ locale$\rightarrow$English. We evaluate all of them, and report English$\rightarrow$locale results in Appendix~\ref{app:subtitle-arena-out-en} and locale$\rightarrow$English results in Appendix~\ref{app:subtitle-arena-into-en}.

\subsection{Relation to Existing Subtitle Benchmarks}
Subtitle Arena complements prior subtitle and multimodal translation resources rather than replacing them. MuST-Cinema~\citep{karakanta2020must} targets speech-to-subtitle translation of TED talks, while BigVideo~\citep{kang-etal-2023-bigvideo} provides video--subtitle supervision for multimodal MT. OpenSubtitles2024~\citep{tiedemann-luo-2026-opensubtitles2024} offers broad multilingual parallel subtitle data at the sentence level, and SubScene~\citep{refine-ai} further broadens multilingual subtitle coverage. Closest to our setting is MuSC~\citep{cui2026utterance}, a multidirectional subtitle corpus built from streaming programs and used to train and evaluate expressive subtitle translation; it is organized as multi-line utterance sentences rather than as complete episodes, and its evaluation targets vividness at the segment level. Table~\ref{tab:benchmark-comparison} summarizes the comparison. In contrast to all of these, Subtitle Arena organizes bilingual subtitle data explicitly around complete episodes and television series, making persistent discourse context and long-horizon consistency first-class evaluation targets, and pairs them with an error-typed subtitle evaluation protocol (Appendix~\ref{app:mqm-protocol}).

\begin{table*}[t]
\centering
\caption{
Subtitle Arena against existing subtitle translation resources.
\emph{Unit} is the largest span the resource is organized around;
\emph{Series context} indicates whether context persists across episodes of the same title;
\emph{Display constraints} indicates whether subtitle timing and line-length limits are part of the evaluation.
}
\label{tab:benchmark-comparison}
\small
\setlength{\tabcolsep}{4pt}
\renewcommand{\arraystretch}{1.15}
\resizebox{\textwidth}{!}{%
\begin{tabular}{
    >{\raggedright\arraybackslash}p{0.17\textwidth}
    >{\raggedright\arraybackslash}p{0.13\textwidth}
    >{\raggedright\arraybackslash}p{0.16\textwidth}
    >{\raggedright\arraybackslash}p{0.15\textwidth}
    c c
    >{\raggedright\arraybackslash}p{0.13\textwidth}
}
\toprule
\rowcolor{tablegray}
\textbf{Resource} & \textbf{Unit} & \textbf{Domain} & \textbf{Directions} & \makecell{\textbf{Series}\\\textbf{context}} & \makecell{\textbf{Display}\\\textbf{constraints}} & \textbf{Evaluation} \\
\midrule
MuST-Cinema & Talk & TED talks (speech) & en$\rightarrow$7 languages & $\times$ & \checkmark & BLEU \\
BigVideo & Clip & Web video & en$\rightarrow$zh & $\times$ & $\times$ & BLEU \\
OpenSubtitles2024 & Sentence & Film and television & Many & $\times$ & $\times$ & Corpus only \\
SubScene & Sentence & Film and television & Many & $\times$ & $\times$ & Corpus only \\
MuSC & Utterance segment & Streaming programs & 6 directions & $\times$ & $\times$ & Accuracy, naturalness, vividness \\
\midrule
\rowcolor{gray!4}
\textbf{Subtitle Arena} & \textbf{Series} & \textbf{Television} & \textbf{30 directions} (15 locales, both ways) & \checkmark & \checkmark & \textbf{SubMQM} (7 dimensions, 19 error types) \\
\bottomrule
\end{tabular}}
\end{table*}


\subsection{Subtitle Arena Visualization}
\label{app:subtitle-arena-visualization}

\begin{figure*}[t]
    \centering
    \includegraphics[width=\textwidth]{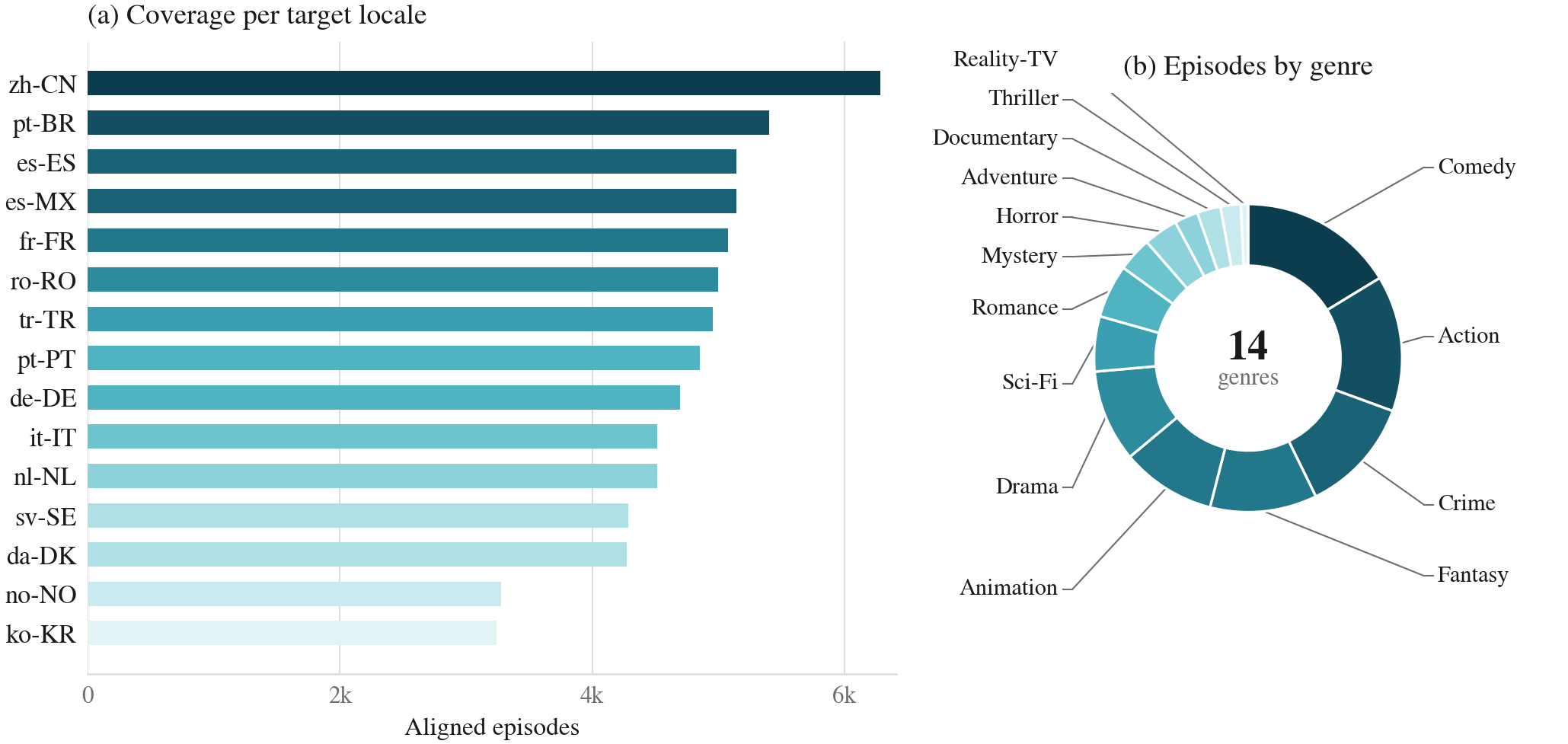}
    \caption{
    Visualization of \textbf{Subtitle Arena} statistics.
    \textbf{Left:} aligned bilingual episode coverage for each target locale, measured against $6{,}267$ English source episodes.
    \textbf{Right:} distribution of the $192$ television series across $14$ genres.
    }
    \label{fig:subtitle-arena-statistics}
\end{figure*}

Figure~\ref{fig:subtitle-arena-statistics} provides a visualization of the language coverage and genre composition summarized in Table~\ref{tab:subtitle-arena-statistics}.

\section{MQM Evaluation Protocol}
\label{app:mqm-protocol}

We propose SubMQM by adapting Multidimensional Quality Metrics (MQM) to subtitle translation. Given a source subtitle and its translation, an LLM judge evaluates a sliding window of consecutive sentences, for each of the 19 subcategories in Table~\ref{tab:mqm-rubric}, a discrete error penalty in $\{0, 5, 10\}$: $0$ indicates no error in that subcategory within the window, $5$ a minor issue that does not impede comprehension, and $10$ a severe error that changes meaning, misleads the viewer, or violates a hard subtitle constraint. Subcategories that are not applicable to a window (e.g., no profanity, no units) default to $0$. All scores are penalties, so lower is better.

\textbf{Aggregation.} A dimension penalty is the mean of its subcategory penalties, averaged over all windows of an episode. The Overall penalty is a weighted average of the seven dimensions,

$\text{Overall} = \sum_{d} w_d \, s_d, \quad \textstyle\sum_d w_d = 1,$

with weights that emphasize semantic fidelity over surface form: Accuracy $0.3$, Terminology $0.2$, Fluency $0.2$, Audience Appropriateness $0.12$, Linguistic Conventions $0.08$, Technical $0.06$, and Locale Conventions $0.04$. A lower Overall indicates fewer and less severe errors.

\textbf{Alignment.} Model outputs preserve the source block segmentation and are scored block-by-block. Human reference subtitles are segmented independently of the source, so we align them to the source by timestamp overlap and score at the passage level, preventing benign re-segmentation from being penalized as omission or mistranslation.

\begin{table*}[t]
\centering
\caption{
Subtitle-adapted MQM rubric used by SubMQM.
The rubric contains seven dimensions and 19 error types.
Each error type receives a penalty in $\{0,5,10\}$; lower is better.
}
\vspace{-10pt}
\label{tab:mqm-rubric}
\small
\setlength{\tabcolsep}{4pt}
\renewcommand{\arraystretch}{1.08}
\begin{tabular}{
    >{\raggedright\arraybackslash}p{0.15\textwidth}
    >{\raggedright\arraybackslash}p{0.19\textwidth}
    >{\raggedright\arraybackslash}p{\dimexpr0.66\textwidth-6\tabcolsep\relax}
}
\toprule
\rowcolor{tablegray}
\textbf{Dimension} & \textbf{Error Type} & \textbf{Description} \\
\midrule
\textbf{Terminology} & Name Inconsistency & Inconsistent translations or spellings of recurring proper names, such as characters, places, and entities. \\
& Term Inconsistency & Inconsistent translations of recurring domain-specific or contextual terms and phrases. \\
\addlinespace[2pt]
\textbf{Accuracy} & Mistranslation & Translation that incorrectly changes the meaning of the source. \\
& Undertranslation & Source content that should be translated is omitted. \\
& Overtranslation & Unsupported information or specificity is introduced into the translation. \\
\addlinespace[2pt]
\textbf{Fluency} & Coherence & The translation lacks continuity with the surrounding scene or discourse. \\
& Naturalness & Awkward or translation-like phrasing that does not resemble native usage. \\
& Vividness & Emotional nuance, humor, wordplay, or stylistic expressiveness is weakened. \\
\addlinespace[2pt]
\textbf{Linguistic Conventions} & Mispunctuation & Missing, incorrect, or improperly formatted punctuation. \\
& Miscapitalization & Incorrect capitalization of sentence-initial words, proper nouns, or other forms. \\
& Grammar & Grammatical errors in agreement, morphology, syntax, or related constructions. \\
& Spacing Error & Missing, extra, or duplicated spaces around words or punctuation. \\
\addlinespace[2pt]
\textbf{Technical} & Incorrect Line Breaking & Line breaks improperly separate tightly coupled linguistic units. \\
& Exceeding Characters per Line & A subtitle line exceeds the predefined character-per-line constraint. \\
& Exceeding Lines per Box & A subtitle event exceeds the predefined maximum number of lines. \\
\addlinespace[2pt]
\textbf{Locale Conventions} & Localization Error & Units, currencies, dates, or culturally specific expressions are improperly localized. \\
& Language Detection Error & The detected language does not match the expected source or target language. \\
\addlinespace[2pt]
\textbf{Audience Appropriateness} & Profanity & Profanity is unjustifiably strengthened, weakened, introduced, or removed. \\
& Formality Error & The register or level of formality is inappropriate or inconsistent with the scene. \\
\bottomrule
\end{tabular}
\vspace{-10pt}
\end{table*}

\section{Additional Experimental Results}
\label{app:additional-experiments}

\subsection{English-to-Locale Results}
\label{app:subtitle-arena-out-en}

Tables~\ref{tab:subtitle_arena_out_en_semantic} and \ref{tab:subtitle_arena_out_en_form} provide the complete fine-grained English$\rightarrow$locale results corresponding to the representative summary in Table~\ref{tab:subtitle_arena_main}. Across the 15 target locales, SMART achieves the lowest Overall SubMQM penalty in every direction. Its mean Overall penalty is $1.11$, compared with $1.20$ for TransAgent, corresponding to a $6.9\%$ relative reduction. The reduction is not confined to one category: the mean Accuracy penalty decreases from $1.83$ to $1.67$, the mean Fluency penalty from $1.88$ to $1.79$, and the mean Technical penalty from $1.83$ to $1.72$.

The fine-grained breakdown further shows reductions in recurring long-form errors such as mistranslation, undertranslation, overtranslation, terminology inconsistency, and line-formatting violations. These improvements are consistent with the components targeted by SMART: persistent series memory maintains cross-episode information, contextual retrieval supplies relevant long-range evidence, and the judge--refiner loop revisits translations when local hypotheses conflict with accumulated context or subtitle constraints.

\begingroup
\scriptsize
\setlength{\tabcolsep}{3.0pt}
\renewcommand{\arraystretch}{0.96}
\setlength\LTleft{0pt}
\setlength\LTright{0pt}
\begin{longtable}{llccccccccc}
\caption{Fine-grained SubMQM results for translation from English into 15 target locales: Terminology, Accuracy, and Fluency. All values are penalties (lower is better). Best and second-best values within each language direction and metric are highlighted with blue-gray and warm-beige backgrounds, respectively. These results provide the full breakdown corresponding to Table~\ref{tab:subtitle_arena_main}.}
\vspace{-10pt}
\label{tab:subtitle_arena_out_en_semantic}\\
\toprule
\rowcolor{tablegray}
\textbf{Dir.} &
\textbf{Method} &
\multicolumn{2}{c}{\textbf{Terminology}} &
\multicolumn{3}{c}{\textbf{Accuracy}} &
\multicolumn{3}{c}{\textbf{Fluency}} &
\textbf{Overall} \\
\rowcolor{tablegray}
& &
NameInc & TermInc &
MisTrans & UndTrans & OvrTrans &
Coher & Natur & Vivid &
\\
\midrule
\endfirsthead
\multicolumn{11}{c}{\tablename~\thetable\ (continued)}\\
\toprule
\rowcolor{tablegray}
\textbf{Dir.} &
\textbf{Method} &
\multicolumn{2}{c}{\textbf{Terminology}} &
\multicolumn{3}{c}{\textbf{Accuracy}} &
\multicolumn{3}{c}{\textbf{Fluency}} &
\textbf{Overall} \\
\rowcolor{tablegray}
& &
NameInc & TermInc &
MisTrans & UndTrans & OvrTrans &
Coher & Natur & Vivid &
\\
\midrule
\endhead
\midrule
\multicolumn{11}{r}{\scriptsize Continued on next page}\\
\endfoot
\bottomrule
\multicolumn{11}{p{\textwidth}}{\scriptsize
\textbf{Abbreviations.}
\textbf{Terminology} (NameInc = Name Inconsistency; TermInc = Term Inconsistency).
\textbf{Accuracy} (MisTrans = Mistranslation; UndTrans = Undertranslation; OvrTrans = Overtranslation).
\textbf{Fluency} (Coher = Coherence; Natur = Naturalness; Vivid = Vividness).
} \\
\endlastfoot
\multirow{9}{*}{\centering en$\rightarrow$zh} & Online & 0.40 & 0.30 & 4.40 & 5.60 & 4.60 & 6.30 & 5.00 & 0.90 & 2.58 \\
 & Gemma 3 4B & 0.52 & 0.46 & 3.45 & 2.68 & 2.15 & 2.65 & 2.12 & 1.15 & 1.46 \\
 & DeepSeek-V3.2 & 0.31 & 0.25 & 2.52 & 1.76 & 1.30 & 1.88 & 1.29 & \secondcell{0.58} & 0.93 \\
 & Claude 4.6 & 0.26 & 0.25 & 2.31 & 1.43 & 1.15 & 2.03 & \secondcell{0.90} & \bestcell{0.56} & 0.84 \\
 & Claude 4.8 & 0.27 & 0.22 & 2.13 & 1.31 & 1.23 & 1.40 & 1.08 & 0.78 & 0.78 \\
 & GPT-5.5 & \secondcell{0.21} & 0.18 & 1.50 & 1.03 & 0.94 & 1.53 & 1.15 & 0.72 & 0.66 \\
\cdashline{2-11}
 & DRT & 0.26 & 0.18 & 1.41 & 0.95 & 0.96 & 1.33 & \bestcell{0.84} & 0.94 & 0.61 \\
 & TransAgent & \bestcell{0.18} & \secondcell{0.13} & \secondcell{0.92} & \bestcell{0.71} & \secondcell{0.92} & \bestcell{1.05} & 1.21 & 0.91 & \secondcell{0.52} \\
\end{longtable}
\endgroup

\begingroup
\tiny
\setlength{\tabcolsep}{1.2pt}
\renewcommand{\arraystretch}{1}
\setlength\LTleft{0pt}
\setlength\LTright{0pt}
\begin{longtable}{llcccccccccccc}
\caption{Fine-grained SubMQM results for translation from English into 15 target locales: Linguistic Conventions, Technical, Locale Conventions, and Audience Appropriateness. All values are penalties (lower is better). Best and second-best values within each language direction and metric are highlighted with blue-gray and warm-beige backgrounds, respectively. These results provide the full breakdown corresponding to Table~\ref{tab:subtitle_arena_main}.}

\label{tab:subtitle_arena_out_en_form}\\
\toprule
\rowcolor{tablegray}
\textbf{Dir.} &
\textbf{Method} &
\multicolumn{4}{c}{\textbf{Linguistic Conventions}} &
\multicolumn{3}{c}{\textbf{Technical}} &
\multicolumn{2}{c}{\textbf{Locale Conventions}} &
\multicolumn{2}{c}{\textbf{Audience Appropriateness}} &
\textbf{Overall} \\
\rowcolor{tablegray}
& &
MisPunc & MisCap & Gram & Space &
LineBrk & CharLim & LineLim &
LocErr & LangDet &
Profan & Formal &
\\
\midrule
\endfirsthead
\multicolumn{14}{c}{\tablename~\thetable\ (continued)}\\
\toprule
\rowcolor{tablegray}
\textbf{Dir.} &
\textbf{Method} &
\multicolumn{4}{c}{\textbf{Linguistic Conventions}} &
\multicolumn{3}{c}{\textbf{Technical}} &
\multicolumn{2}{c}{\textbf{Locale Conventions}} &
\multicolumn{2}{c}{\textbf{Audience Appropriateness}} &
\textbf{Overall} \\
\rowcolor{tablegray}
& &
MisPunc & MisCap & Gram & Space &
LineBrk & CharLim & LineLim &
LocErr & LangDet &
Profan & Formal &
\\
\midrule
\endhead
\midrule
\multicolumn{14}{r}{\scriptsize Continued on next page}\\
\endfoot
\bottomrule
\multicolumn{14}{p{\textwidth}}{\scriptsize
\textbf{Abbreviations.}
\textbf{Linguistic Conventions} (MisPunc = Mispunctuation; MisCap = Miscapitalization; Gram = Grammar; Space = Spacing Error).
\textbf{Technical} (LineBrk = Incorrect Line Breaking; CharLim = Exceeding Characters per Line; LineLim = Exceeding Lines per Box).
\textbf{Locale Conventions} (LocErr = Localization Error; LangDet = Language Detection Error).
\textbf{Audience Appropriateness} (Profan = Profanity; Formal = Formality Error).
} \\
\endlastfoot
\multirow{9}{*}{\centering en$\rightarrow$zh} & Online & 4.30 & \bestcell{0.00} & 0.50 & 3.10 & 0.10 & 0.50 & \bestcell{0.00} & \bestcell{0.00} & 2.80 & \bestcell{0.10} & \bestcell{0.00} & 2.58 \\
 & Gemma 3 4B & 1.05 & \bestcell{0.00} & 0.58 & 0.72 & 1.18 & 0.84 & 0.35 & 0.18 & 0.12 & 0.38 & 0.31 & 1.46 \\
 & DeepSeek-V3.2 & 0.61 & \bestcell{0.00} & 0.19 & 0.34 & 0.82 & 0.52 & 0.14 & \secondcell{0.04} & \bestcell{0.01} & 0.15 & 0.10 & 0.93 \\
 & Claude 4.6 & 0.55 & \bestcell{0.00} & 0.16 & 0.31 & 0.70 & 0.44 & 0.12 & \secondcell{0.04} & \secondcell{0.02} & 0.20 & 0.11 & 0.84 \\
 & Claude 4.8 & 0.41 & \bestcell{0.00} & 0.13 & 0.26 & 0.53 & 0.34 & 0.09 & 0.05 & \secondcell{0.02} & 0.15 & 0.10 & 0.78 \\
 & GPT-5.5 & 0.37 & \bestcell{0.00} & 0.12 & 0.27 & 0.37 & 0.23 & 0.06 & \secondcell{0.04} & \secondcell{0.02} & 0.18 & 0.10 & 0.66 \\
\cdashline{2-14}
 & DRT & 0.24 & \bestcell{0.00} & 0.08 & \secondcell{0.16} & 0.20 & 0.13 & 0.04 & 0.06 & \secondcell{0.02} & \secondcell{0.11} & 0.10 & 0.61 \\
 & TransAgent & \secondcell{0.15} & \bestcell{0.00} & \secondcell{0.06} & \bestcell{0.13} & \secondcell{0.08} & \secondcell{0.05} & \secondcell{0.01} & \secondcell{0.04} & 0.03 & 0.19 & \secondcell{0.08} & \secondcell{0.52} \\
\end{longtable}
\endgroup

\subsection{Locale-to-English Results}
\label{app:subtitle-arena-into-en}

Tables~\ref{tab:subtitle_arena_into_en_semantic} and \ref{tab:subtitle_arena_into_en_form} provide the complete locale$\rightarrow$English results corresponding to Table~\ref{tab:subtitle_arena_main}. SMART again achieves the lowest Overall SubMQM penalty in all 15 directions, with a mean Overall penalty of $0.37$ compared with $0.41$ for TransAgent, a $10.0\%$ relative reduction. The mean Accuracy penalty decreases from $0.75$ to $0.67$, while the mean Technical penalty decreases from $0.13$ to $0.07$; Terminology, Fluency, Linguistic Conventions, Locale Conventions, and Audience Appropriateness also improve on average.

The error-type breakdown is not uniformly better in every individual category: for example, Vividness and Profanity show small mixed changes in some directions. The Overall improvement instead comes from consistent reductions across the higher-weight semantic and consistency errors, particularly mistranslation, undertranslation, overtranslation, coherence, terminology inconsistency, and subtitle-formatting violations. Because every system generates English in this setting, these results also show that SMART's gains are not specific to target-language morphology or typography.

\begingroup
\scriptsize
\setlength{\tabcolsep}{3.0pt}
\renewcommand{\arraystretch}{1}
\setlength\LTleft{0pt}
\setlength\LTright{0pt}
\begin{longtable}{llccccccccc}
\caption{Fine-grained SubMQM results for translation from 15 source locales into English: Terminology, Accuracy, and Fluency. All values are penalties (lower is better). Best and second-best values within each language direction and metric are highlighted with blue-gray and warm-beige backgrounds, respectively. These results provide the full breakdown corresponding to Table~\ref{tab:subtitle_arena_main}.}\label{tab:subtitle_arena_into_en_semantic}\\
\toprule
\rowcolor{tablegray}
\textbf{Dir.} &
\textbf{Method} &
\multicolumn{2}{c}{\textbf{Terminology}} &
\multicolumn{3}{c}{\textbf{Accuracy}} &
\multicolumn{3}{c}{\textbf{Fluency}} &
\textbf{Overall} \\
\rowcolor{tablegray}
& &
NameInc & TermInc &
MisTrans & UndTrans & OvrTrans &
Coher & Natur & Vivid &
\\
\midrule
\endfirsthead
\multicolumn{11}{c}{\tablename~\thetable\ (continued)}\\
\toprule
\rowcolor{tablegray}
\textbf{Dir.} &
\textbf{Method} &
\multicolumn{2}{c}{\textbf{Terminology}} &
\multicolumn{3}{c}{\textbf{Accuracy}} &
\multicolumn{3}{c}{\textbf{Fluency}} &
\textbf{Overall} \\
\rowcolor{tablegray}
& &
NameInc & TermInc &
MisTrans & UndTrans & OvrTrans &
Coher & Natur & Vivid &
\\
\midrule
\endhead
\midrule
\multicolumn{11}{r}{\scriptsize Continued on next page}\\
\endfoot
\bottomrule
\multicolumn{11}{p{\textwidth}}{\scriptsize
\textbf{Abbreviations.}
\textbf{Terminology} (NameInc = Name Inconsistency; TermInc = Term Inconsistency).
\textbf{Accuracy} (MisTrans = Mistranslation; UndTrans = Undertranslation; OvrTrans = Overtranslation).
\textbf{Fluency} (Coher = Coherence; Natur = Naturalness; Vivid = Vividness).
} \\
\endlastfoot
\multirow{9}{*}{\centering zh$\rightarrow$en} & Online & 0.47 & 0.36 & 4.92 & 4.05 & 3.71 & 4.56 & 3.79 & 1.49 & 2.17 \\
 & Gemma 3 4B & 0.58 & 0.45 & 3.91 & 2.89 & 2.51 & 2.91 & 2.24 & 1.34 & 1.64 \\
 & DeepSeek-V3.2 & 0.36 & 0.26 & 2.74 & 1.86 & 1.50 & 2.11 & 1.39 & 0.78 & 1.07 \\
 & Claude 4.6 & 0.32 & 0.22 & 2.49 & 1.68 & 1.34 & 1.89 & 1.26 & 0.69 & 0.96 \\
 & Claude 4.8 & 0.24 & 0.16 & 1.86 & 1.24 & 1.08 & 1.49 & 0.98 & 0.54 & 0.73 \\
 & GPT-5.5 & 0.21 & 0.14 & 1.59 & 1.06 & 0.94 & 1.32 & 1.04 & 0.58 & 0.67 \\
\cdashline{2-11}
 & DRT & 0.19 & 0.13 & 1.28 & 0.91 & 0.84 & 1.19 & 0.88 & \secondcell{0.52} & 0.55 \\
 & TransAgent & \secondcell{0.17} & \secondcell{0.11} & \secondcell{1.16} & \secondcell{0.82} & \secondcell{0.77} & \secondcell{1.10} & \secondcell{0.80} & \bestcell{0.46} & \secondcell{0.50} \\
\end{longtable}
\endgroup

\begingroup
\tiny
\setlength{\tabcolsep}{1.2pt}
\renewcommand{\arraystretch}{1}
\setlength\LTleft{0pt}
\setlength\LTright{0pt}
\begin{longtable}{llcccccccccccc}
\caption{Fine-grained SubMQM results for translation from 15 source locales into English: Linguistic Conventions, Technical, Locale Conventions, and Audience Appropriateness. All values are penalties (lower is better). Best and second-best values within each language direction and metric are highlighted with blue-gray and warm-beige backgrounds, respectively. These results provide the full breakdown corresponding to Table~\ref{tab:subtitle_arena_main}.}\label{tab:subtitle_arena_into_en_form}\\
\toprule
\rowcolor{tablegray}
\textbf{Dir.} &
\textbf{Method} &
\multicolumn{4}{c}{\textbf{Linguistic Conventions}} &
\multicolumn{3}{c}{\textbf{Technical}} &
\multicolumn{2}{c}{\textbf{Locale Conventions}} &
\multicolumn{2}{c}{\textbf{Audience Appropriateness}} &
\textbf{Overall} \\
\rowcolor{tablegray}
& &
MisPunc & MisCap & Gram & Space &
LineBrk & CharLim & LineLim &
LocErr & LangDet &
Profan & Formal &
\\
\midrule
\endfirsthead
\multicolumn{14}{c}{\tablename~\thetable\ (continued)}\\
\toprule
\rowcolor{tablegray}
\textbf{Dir.} &
\textbf{Method} &
\multicolumn{4}{c}{\textbf{Linguistic Conventions}} &
\multicolumn{3}{c}{\textbf{Technical}} &
\multicolumn{2}{c}{\textbf{Locale Conventions}} &
\multicolumn{2}{c}{\textbf{Audience Appropriateness}} &
\textbf{Overall} \\
\rowcolor{tablegray}
& &
MisPunc & MisCap & Gram & Space &
LineBrk & CharLim & LineLim &
LocErr & LangDet &
Profan & Formal &
\\
\midrule
\endhead
\midrule
\multicolumn{14}{r}{\scriptsize Continued on next page}\\
\endfoot
\bottomrule
\multicolumn{14}{p{\textwidth}}{\scriptsize
\textbf{Abbreviations.}
\textbf{Linguistic Conventions} (MisPunc = Mispunctuation; MisCap = Miscapitalization; Gram = Grammar; Space = Spacing Error).
\textbf{Technical} (LineBrk = Incorrect Line Breaking; CharLim = Exceeding Characters per Line; LineLim = Exceeding Lines per Box).
\textbf{Locale Conventions} (LocErr = Localization Error; LangDet = Language Detection Error).
\textbf{Audience Appropriateness} (Profan = Profanity; Formal = Formality Error).
} \\
\endlastfoot
\multirow{9}{*}{\centering zh$\rightarrow$en} & Online & 2.16 & 1.19 & 1.71 & 0.81 & 0.49 & 0.31 & \secondcell{0.06} & 0.18 & 0.03 & 0.28 & 0.15 & 2.17 \\
 & Gemma 3 4B & 1.21 & 0.56 & 0.88 & 0.59 & 1.51 & 1.04 & 0.44 & 0.26 & 0.05 & 0.46 & 0.31 & 1.64 \\
 & DeepSeek-V3.2 & 0.64 & 0.32 & 0.46 & 0.24 & 1.41 & 0.99 & 0.24 & 0.15 & \secondcell{0.01} & 0.25 & 0.10 & 1.07 \\
 & Claude 4.6 & 0.59 & 0.18 & 0.39 & 0.18 & 1.31 & 0.89 & 0.29 & 0.06 & 0.04 & 0.15 & 0.15 & 0.96 \\
 & Claude 4.8 & 0.42 & 0.11 & 0.28 & 0.12 & 1.11 & 0.72 & 0.22 & 0.03 & 0.03 & \secondcell{0.11} & 0.12 & 0.73 \\
 & GPT-5.5 & 0.34 & 0.08 & 0.24 & 0.10 & 1.38 & 0.92 & 0.32 & 0.06 & \bestcell{0.00} & 0.18 & 0.07 & 0.67 \\
\cdashline{2-14}
 & DRT & 0.28 & 0.05 & 0.20 & 0.07 & 0.51 & 0.26 & 0.13 & \secondcell{0.02} & 0.03 & 0.16 & \secondcell{0.05} & 0.55 \\
 & TransAgent & \secondcell{0.24} & \secondcell{0.04} & \secondcell{0.18} & \secondcell{0.06} & \secondcell{0.44} & \secondcell{0.20} & 0.11 & 0.05 & \bestcell{0.00} & \bestcell{0.10} & 0.09 & \secondcell{0.50} \\
\end{longtable}
\endgroup

\subsection{Backbone and Judge Analysis}
\label{app:backbone-judge-full}

Tables~\ref{tab:backbone_out_en_semantic} and \ref{tab:backbone_out_en_form} report the complete English$\rightarrow$locale backbone and judge analysis, while Tables~\ref{tab:backbone_into_en_semantic} and \ref{tab:backbone_into_en_form} report the corresponding locale$\rightarrow$English results. These tables are the full fine-grained counterpart of the representative results summarized in Table~\ref{tab:backbone_judge_main}. Across directions, applying SMART to Gemma 3 4B or DeepSeek-V3.2 reduces the Overall penalty relative to direct translation with the same model, indicating that the gain is not explained solely by the strength of Claude Sonnet 4.6.

Changing only the evaluator from Claude Sonnet 4.6 to GPT-5.5 produces substantially smaller differences than changing the translation scaffold or backbone. For example, on en$\rightarrow$zh the Overall penalty changes from $0.48$ to $0.47$, while direct Gemma 3 4B and SMART with Gemma 3 4B differ by $1.46$ versus $0.67$. The same pattern appears in the reverse direction, where zh$\rightarrow$en changes from $0.44$ to $0.44$ under the alternative judge, compared with $1.64$ versus $0.62$ for direct and SMART-wrapped Gemma 3 4B. This separates the effect of the translation framework from small evaluator-specific score variation.

\begingroup
\scriptsize
\setlength{\tabcolsep}{3.0pt}
\renewcommand{\arraystretch}{0.96}
\setlength\LTleft{0pt}
\setlength\LTright{0pt}

\begin{longtable}{llccccccccc}
\caption{Fine-grained backbone and judge analysis on Subtitle Arena for translation from English into 15 target locales: Terminology, Accuracy, and Fluency. All values are penalties (lower is better). Best and second-best values within each language direction and metric are highlighted with blue-gray and warm-beige backgrounds, respectively. These results provide the full breakdown corresponding to Table~\ref{tab:backbone_judge_main}.}
\label{tab:backbone_out_en_semantic}\\
\toprule
\rowcolor{tablegray}
\textbf{Dir.} &
\textbf{Method / Setting} &
\multicolumn{2}{c}{\textbf{Terminology}} &
\multicolumn{3}{c}{\textbf{Accuracy}} &
\multicolumn{3}{c}{\textbf{Fluency}} &
\textbf{Overall} \\
\rowcolor{tablegray}
& &
NameInc & TermInc &
MisTrans & UndTrans & OvrTrans &
Coher & Natur & Vivid &
\\
\midrule
\endfirsthead

\multicolumn{11}{c}{\tablename~\thetable\ (continued)}\\
\toprule
\rowcolor{tablegray}
\textbf{Dir.} &
\textbf{Method / Setting} &
\multicolumn{2}{c}{\textbf{Terminology}} &
\multicolumn{3}{c}{\textbf{Accuracy}} &
\multicolumn{3}{c}{\textbf{Fluency}} &
\textbf{Overall} \\
\rowcolor{tablegray}
& &
NameInc & TermInc &
MisTrans & UndTrans & OvrTrans &
Coher & Natur & Vivid &
\\
\midrule
\endhead

\midrule
\multicolumn{11}{r}{\scriptsize Continued on next page}\\
\endfoot

\bottomrule
\multicolumn{11}{p{\textwidth}}{\scriptsize
\textbf{Abbreviations.}
\textbf{Terminology} (NameInc = Name Inconsistency; TermInc = Term Inconsistency).
\textbf{Accuracy} (MisTrans = Mistranslation; UndTrans = Undertranslation; OvrTrans = Overtranslation).
\textbf{Fluency} (Coher = Coherence; Natur = Naturalness; Vivid = Vividness).
} \\
\endlastfoot
\multirow{6}{*}{\centering en$\rightarrow$zh} & Gemma 3 4B & 0.52 & 0.46 & 3.45 & 2.68 & 2.15 & 2.65 & 2.12 & 1.15 & 1.46 \\
 & DeepSeek-V3.2 & 0.31 & 0.25 & 2.52 & 1.76 & 1.30 & 1.88 & 1.29 & \bestcell{0.58} & 0.93 \\
\cdashline{2-11}
 & SMART (Gemma 3 4B) & 0.30 & 0.16 & 1.20 & 1.08 & 1.14 & 1.39 & 1.28 & 1.14 & 0.67 \\
 & SMART (DeepSeek-V3.2) & \secondcell{0.22} & \secondcell{0.13} & 0.96 & 0.88 & 0.84 & 1.14 & 1.00 & 0.98 & 0.53 \\
 & SMART (Judge: GPT-5.5) & \bestcell{0.18} & \bestcell{0.12} & \secondcell{0.90} & \secondcell{0.83} & \bestcell{0.68} & \bestcell{1.03} & \bestcell{0.89} & \secondcell{0.78} & \bestcell{0.47} \\
\rowcolor{gray!4}  & \textbf{SMART (Claude 4.6)} & \bestcell{0.18} & \bestcell{0.12} & \bestcell{0.86} & \bestcell{0.81} & \secondcell{0.73} & \secondcell{1.08} & \secondcell{0.94} & 0.83 & \secondcell{0.48} \\
\midrule
\multirow{6}{*}{\centering en$\rightarrow$de} & Gemma 3 4B & 0.78 & 0.62 & 4.85 & 4.12 & 3.56 & 3.45 & 2.89 & 1.88 & 2.21 \\
 & DeepSeek-V3.2 & 0.50 & 0.37 & 3.52 & 2.81 & 2.38 & 2.54 & 1.98 & \bestcell{1.25} & 1.50 \\
\cdashline{2-11}
 & SMART (Gemma 3 4B) & 0.34 & 0.32 & 1.80 & 1.68 & 1.55 & 2.46 & 2.28 & 1.80 & 1.23 \\
 & SMART (DeepSeek-V3.2) & \secondcell{0.30} & 0.23 & 1.54 & 1.50 & 1.19 & 1.87 & 1.76 & 1.42 & \secondcell{0.99} \\
 & SMART (Judge: GPT-5.5) & \bestcell{0.29} & \bestcell{0.20} & \secondcell{1.43} & \bestcell{1.23} & \bestcell{1.00} & \secondcell{1.65} & \secondcell{1.48} & 1.39 & \bestcell{0.87} \\
\rowcolor{gray!4}  & \textbf{SMART (Claude 4.6)} & \bestcell{0.29} & \secondcell{0.21} & \bestcell{1.39} & \secondcell{1.28} & \secondcell{1.05} & \bestcell{1.62} & \bestcell{1.46} & \secondcell{1.30} & \bestcell{0.87} \\
\midrule
\multirow{6}{*}{\centering en$\rightarrow$ko} & Gemma 3 4B & 0.75 & 0.61 & 3.82 & 3.12 & 2.74 & 2.98 & 2.54 & 1.78 & 1.82 \\
 & DeepSeek-V3.2 & 0.48 & 0.35 & 2.58 & 1.98 & 1.72 & 2.08 & 1.66 & \bestcell{1.14} & 1.17 \\
\cdashline{2-11}
 & SMART (Gemma 3 4B) & 0.45 & 0.38 & 2.16 & 2.06 & 1.86 & 1.98 & 2.12 & 1.77 & 1.26 \\
 & SMART (DeepSeek-V3.2) & 0.38 & \secondcell{0.28} & 1.93 & 1.68 & 1.44 & 1.70 & 1.68 & 1.58 & 1.03 \\
 & SMART (Judge: GPT-5.5) & \secondcell{0.36} & \bestcell{0.24} & \secondcell{1.79} & \bestcell{1.51} & \bestcell{1.28} & \secondcell{1.60} & \bestcell{1.36} & \secondcell{1.32} & \bestcell{0.91} \\
\rowcolor{gray!4}  & \textbf{SMART (Claude 4.6)} & \bestcell{0.34} & \bestcell{0.24} & \bestcell{1.76} & \secondcell{1.56} & \secondcell{1.33} & \bestcell{1.56} & \secondcell{1.41} & \secondcell{1.32} & \secondcell{0.92} \\
\midrule
\multirow{6}{*}{\centering en$\rightarrow$it} & Gemma 3 4B & 0.78 & 0.62 & 4.95 & 4.15 & 3.65 & 3.98 & 3.42 & 2.35 & 2.34 \\
 & DeepSeek-V3.2 & 0.50 & 0.36 & 3.60 & 2.79 & 2.44 & 3.08 & 2.59 & \bestcell{1.71} & 1.62 \\
\cdashline{2-11}
 & SMART (Gemma 3 4B) & 0.40 & 0.25 & 2.16 & 1.86 & 1.89 & 3.25 & 2.99 & 2.84 & 1.56 \\
 & SMART (DeepSeek-V3.2) & 0.30 & 0.19 & 1.89 & 1.55 & 1.62 & 2.76 & 2.50 & 2.23 & \secondcell{1.30} \\
 & SMART (Judge: GPT-5.5) & \secondcell{0.28} & \secondcell{0.17} & \bestcell{1.61} & \bestcell{1.45} & \bestcell{1.30} & \bestcell{2.34} & \bestcell{2.07} & 2.16 & \bestcell{1.13} \\
\rowcolor{gray!4}  & \textbf{SMART (Claude 4.6)} & \bestcell{0.26} & \bestcell{0.16} & \secondcell{1.66} & \secondcell{1.46} & \secondcell{1.35} & \secondcell{2.36} & \secondcell{2.12} & \secondcell{1.97} & \bestcell{1.13} \\
\midrule
\multirow{6}{*}{\centering en$\rightarrow$es-ES} & Gemma 3 4B & 0.82 & 0.65 & 5.12 & 4.28 & 3.76 & 3.82 & 3.28 & 2.25 & 2.36 \\
 & DeepSeek-V3.2 & 0.52 & 0.38 & 3.70 & 2.89 & 2.51 & 2.86 & 2.38 & \bestcell{1.60} & 1.62 \\
\cdashline{2-11}
 & SMART (Gemma 3 4B) & 0.69 & 0.52 & 2.76 & 2.78 & 2.25 & 3.13 & 2.18 & 2.27 & 1.71 \\
 & SMART (DeepSeek-V3.2) & 0.50 & 0.38 & 2.38 & 2.05 & 1.69 & 2.51 & 1.93 & 1.88 & 1.36 \\
 & SMART (Judge: GPT-5.5) & \secondcell{0.46} & \bestcell{0.32} & \bestcell{1.91} & \bestcell{1.71} & \bestcell{1.54} & \bestcell{2.01} & \bestcell{1.75} & 1.77 & \bestcell{1.17} \\
\rowcolor{gray!4}  & \textbf{SMART (Claude 4.6)} & \bestcell{0.44} & \secondcell{0.34} & \secondcell{1.96} & \secondcell{1.76} & \secondcell{1.59} & \secondcell{2.06} & \secondcell{1.80} & \secondcell{1.66} & \secondcell{1.18} \\
\midrule
\multirow{6}{*}{\centering en$\rightarrow$fr} & Gemma 3 4B & 0.72 & 0.56 & 4.90 & 4.08 & 3.58 & 3.95 & 3.38 & 2.32 & 2.31 \\
 & DeepSeek-V3.2 & 0.42 & 0.28 & 3.55 & 2.80 & 2.40 & 3.04 & 2.50 & \bestcell{1.69} & 1.60 \\
\cdashline{2-11}
 & SMART (Gemma 3 4B) & 0.33 & 0.13 & 2.86 & 1.99 & 1.94 & 3.84 & 3.17 & 2.90 & 1.74 \\
 & SMART (DeepSeek-V3.2) & \secondcell{0.25} & \secondcell{0.11} & 2.22 & 1.72 & 1.69 & 2.87 & 2.60 & 2.33 & \secondcell{1.40} \\
 & SMART (Judge: GPT-5.5) & \bestcell{0.20} & \bestcell{0.10} & \bestcell{1.80} & \bestcell{1.53} & \bestcell{1.38} & \secondcell{2.52} & \secondcell{2.30} & \secondcell{1.92} & \bestcell{1.19} \\
\rowcolor{gray!4}  & \textbf{SMART (Claude 4.6)} & \bestcell{0.20} & \bestcell{0.10} & \secondcell{1.82} & \secondcell{1.58} & \secondcell{1.43} & \bestcell{2.41} & \bestcell{2.13} & 1.97 & \bestcell{1.19} \\
\midrule
\multirow{6}{*}{\centering en$\rightarrow$pt-PT} & Gemma 3 4B & 1.38 & 1.04 & 6.80 & 5.06 & 5.47 & 6.44 & 4.75 & 5.14 & 3.72 \\
 & DeepSeek-V3.2 & 1.01 & 0.84 & 5.05 & 3.62 & 3.70 & 4.73 & 3.70 & 4.07 & 2.73 \\
\cdashline{2-11}
 & SMART (Gemma 3 4B) & 0.76 & 0.49 & 3.91 & 2.48 & 2.20 & 2.68 & 2.88 & 2.90 & 1.89 \\
 & SMART (DeepSeek-V3.2) & \secondcell{0.60} & 0.42 & 2.81 & 2.24 & 1.87 & \secondcell{2.43} & \secondcell{2.17} & 2.08 & 1.50 \\
 & SMART (Judge: GPT-5.5) & \bestcell{0.53} & \bestcell{0.35} & \bestcell{2.21} & \bestcell{1.76} & \bestcell{1.67} & \bestcell{2.38} & \bestcell{1.99} & \bestcell{2.00} & \bestcell{1.30} \\
\rowcolor{gray!4}  & \textbf{SMART (Claude 4.6)} & \bestcell{0.53} & \secondcell{0.37} & \secondcell{2.41} & \secondcell{1.93} & \secondcell{1.83} & \secondcell{2.43} & 2.18 & \secondcell{2.03} & \secondcell{1.39} \\
\midrule
\multirow{6}{*}{\centering en$\rightarrow$pt-BR} & Gemma 3 4B & 1.04 & 0.78 & 4.24 & 3.72 & 3.26 & 4.30 & 3.69 & 3.60 & 2.56 \\
 & DeepSeek-V3.2 & 0.71 & 0.53 & 3.27 & 2.88 & 2.52 & 3.29 & 2.73 & 2.52 & 1.91 \\
\cdashline{2-11}
 & SMART (Gemma 3 4B) & 0.53 & 0.38 & 2.79 & 1.80 & 1.52 & 2.68 & 2.17 & 2.24 & 1.42 \\
 & SMART (DeepSeek-V3.2) & 0.39 & 0.34 & 2.03 & 1.61 & 1.36 & 2.17 & \secondcell{1.68} & 1.68 & 1.14 \\
 & SMART (Judge: GPT-5.5) & \bestcell{0.31} & \bestcell{0.30} & \bestcell{1.69} & \bestcell{1.40} & \bestcell{1.25} & \bestcell{1.71} & \bestcell{1.53} & \bestcell{1.32} & \bestcell{0.97} \\
\rowcolor{gray!4}  & \textbf{SMART (Claude 4.6)} & \secondcell{0.36} & \secondcell{0.32} & \secondcell{1.79} & \secondcell{1.48} & \secondcell{1.26} & \secondcell{1.85} & 1.69 & \secondcell{1.45} & \secondcell{1.04} \\
\midrule
\multirow{6}{*}{\centering en$\rightarrow$no} & Gemma 3 4B & 1.30 & 0.97 & 4.32 & 4.99 & 3.96 & 5.22 & 5.42 & 3.57 & 2.99 \\
 & DeepSeek-V3.2 & 0.93 & 0.76 & 3.33 & 3.37 & 3.04 & 3.96 & 3.64 & 2.54 & 2.18 \\
\cdashline{2-11}
 & SMART (Gemma 3 4B) & 0.58 & 0.41 & 2.29 & 2.49 & 2.32 & 3.07 & 2.60 & 2.09 & 1.60 \\
 & SMART (DeepSeek-V3.2) & 0.49 & 0.35 & 2.02 & 1.93 & 1.96 & 2.26 & 2.10 & 1.65 & 1.29 \\
 & SMART (Judge: GPT-5.5) & \bestcell{0.43} & \bestcell{0.31} & \bestcell{1.64} & \bestcell{1.70} & \bestcell{1.59} & \bestcell{1.90} & \bestcell{1.82} & \bestcell{1.39} & \bestcell{1.10} \\
\rowcolor{gray!4}  & \textbf{SMART (Claude 4.6)} & \secondcell{0.46} & \secondcell{0.33} & \secondcell{1.78} & \secondcell{1.86} & \secondcell{1.74} & \secondcell{2.05} & \secondcell{1.99} & \secondcell{1.53} & \secondcell{1.19} \\
\midrule
\multirow{6}{*}{\centering en$\rightarrow$da} & Gemma 3 4B & 1.10 & 0.88 & 6.12 & 4.72 & 3.83 & 6.45 & 5.29 & 4.23 & 3.30 \\
 & DeepSeek-V3.2 & 0.84 & 0.60 & 4.31 & 3.22 & 3.00 & 4.70 & 3.56 & 2.93 & 2.34 \\
\cdashline{2-11}
 & SMART (Gemma 3 4B) & 0.63 & 0.50 & 3.23 & 2.55 & 2.62 & 2.84 & 2.61 & 2.51 & 1.79 \\
 & SMART (DeepSeek-V3.2) & 0.55 & \secondcell{0.39} & 2.54 & 2.02 & 1.88 & 2.39 & 1.92 & 1.89 & 1.39 \\
 & SMART (Judge: GPT-5.5) & \secondcell{0.52} & \bestcell{0.34} & \bestcell{2.11} & \bestcell{1.72} & \bestcell{1.65} & \bestcell{2.09} & \bestcell{1.64} & \bestcell{1.47} & \bestcell{1.19} \\
\rowcolor{gray!4}  & \textbf{SMART (Claude 4.6)} & \bestcell{0.50} & \bestcell{0.34} & \secondcell{2.30} & \secondcell{1.88} & \secondcell{1.74} & \secondcell{2.12} & \secondcell{1.80} & \secondcell{1.61} & \secondcell{1.26} \\
\midrule
\multirow{6}{*}{\centering en$\rightarrow$nl} & Gemma 3 4B & 1.21 & 1.08 & 5.75 & 4.02 & 5.25 & 5.54 & 5.17 & 5.14 & 3.41 \\
 & DeepSeek-V3.2 & 0.82 & 0.77 & 4.67 & 3.17 & 3.74 & 4.17 & 3.99 & 3.71 & 2.57 \\
\cdashline{2-11}
 & SMART (Gemma 3 4B) & 0.66 & 0.56 & 3.74 & 2.22 & 2.62 & 3.01 & 2.46 & 2.63 & 1.80 \\
 & SMART (DeepSeek-V3.2) & 0.59 & 0.46 & 2.73 & 1.92 & 2.19 & 2.47 & 2.17 & 2.20 & 1.48 \\
 & SMART (Judge: GPT-5.5) & \bestcell{0.46} & \secondcell{0.41} & \bestcell{2.17} & \bestcell{1.85} & \bestcell{1.72} & \bestcell{2.30} & \bestcell{2.08} & \bestcell{1.89} & \bestcell{1.29} \\
\rowcolor{gray!4}  & \textbf{SMART (Claude 4.6)} & \secondcell{0.52} & \bestcell{0.40} & \secondcell{2.36} & \secondcell{1.91} & \secondcell{1.88} & \secondcell{2.44} & \secondcell{2.14} & \secondcell{2.02} & \secondcell{1.38} \\
\midrule
\multirow{6}{*}{\centering en$\rightarrow$ro} & Gemma 3 4B & 1.12 & 0.94 & 5.01 & 4.20 & 4.23 & 5.61 & 3.79 & 4.67 & 2.96 \\
 & DeepSeek-V3.2 & 0.81 & 0.65 & 3.50 & 3.13 & 3.24 & 3.80 & 2.95 & 3.51 & 2.17 \\
\cdashline{2-11}
 & SMART (Gemma 3 4B) & 0.55 & 0.48 & 2.39 & 2.08 & 1.92 & 2.53 & 2.04 & 2.23 & 1.45 \\
 & SMART (DeepSeek-V3.2) & \secondcell{0.46} & 0.37 & 1.86 & 1.81 & \secondcell{1.56} & 1.88 & 1.49 & 1.66 & 1.14 \\
 & SMART (Judge: GPT-5.5) & \bestcell{0.43} & \bestcell{0.32} & \bestcell{1.68} & \bestcell{1.44} & \bestcell{1.42} & \bestcell{1.80} & \bestcell{1.34} & \bestcell{1.48} & \bestcell{1.01} \\
\rowcolor{gray!4}  & \textbf{SMART (Claude 4.6)} & \bestcell{0.43} & \secondcell{0.33} & \secondcell{1.84} & \secondcell{1.58} & \secondcell{1.56} & \secondcell{1.84} & \secondcell{1.46} & \secondcell{1.51} & \secondcell{1.07} \\
\midrule
\multirow{6}{*}{\centering en$\rightarrow$sv} & Gemma 3 4B & 0.93 & 0.78 & 3.86 & 4.33 & 3.62 & 3.99 & 3.16 & 4.01 & 2.57 \\
 & DeepSeek-V3.2 & 0.77 & 0.56 & 3.14 & 2.91 & 2.59 & 3.31 & 2.44 & 2.75 & 1.91 \\
\cdashline{2-11}
 & SMART (Gemma 3 4B) & 0.48 & 0.36 & 2.76 & 2.19 & 1.68 & 2.50 & 2.27 & 1.79 & 1.43 \\
 & SMART (DeepSeek-V3.2) & 0.37 & 0.32 & 2.03 & 1.72 & 1.47 & 2.07 & 1.65 & 1.55 & 1.14 \\
 & SMART (Judge: GPT-5.5) & \bestcell{0.32} & \bestcell{0.25} & \bestcell{1.66} & \bestcell{1.46} & \bestcell{1.21} & \bestcell{1.61} & \bestcell{1.48} & \bestcell{1.23} & \bestcell{0.95} \\
\rowcolor{gray!4}  & \textbf{SMART (Claude 4.6)} & \secondcell{0.36} & \secondcell{0.28} & \secondcell{1.81} & \secondcell{1.60} & \secondcell{1.34} & \secondcell{1.76} & \secondcell{1.61} & \secondcell{1.36} & \secondcell{1.04} \\
\midrule
\multirow{6}{*}{\centering en$\rightarrow$tr} & Gemma 3 4B & 1.18 & 0.85 & 5.78 & 4.39 & 3.64 & 5.10 & 4.48 & 4.30 & 3.04 \\
 & DeepSeek-V3.2 & 0.93 & 0.62 & 4.18 & 3.14 & 2.58 & 3.94 & 3.46 & 3.45 & 2.26 \\
\cdashline{2-11}
 & SMART (Gemma 3 4B) & 0.57 & 0.52 & 2.77 & 2.76 & 1.92 & 2.43 & 2.50 & 2.76 & 1.66 \\
 & SMART (DeepSeek-V3.2) & 0.47 & 0.39 & 2.30 & 2.15 & 1.71 & 2.12 & 2.09 & 2.04 & 1.35 \\
 & SMART (Judge: GPT-5.5) & \bestcell{0.41} & \bestcell{0.37} & \bestcell{2.01} & \bestcell{1.82} & \bestcell{1.41} & \bestcell{1.85} & \secondcell{1.84} & \bestcell{1.78} & \bestcell{1.17} \\
\rowcolor{gray!4}  & \textbf{SMART (Claude 4.6)} & \secondcell{0.45} & \secondcell{0.38} & \secondcell{2.19} & \secondcell{1.98} & \secondcell{1.55} & \secondcell{2.01} & \bestcell{1.83} & \secondcell{1.80} & \secondcell{1.25} \\
\midrule
\multirow{6}{*}{\centering en$\rightarrow$es-MX} & Gemma 3 4B & 1.17 & 1.00 & 5.30 & 6.87 & 4.85 & 5.40 & 4.58 & 4.67 & 3.43 \\
 & DeepSeek-V3.2 & 0.94 & 0.69 & 4.05 & 4.73 & 3.28 & 3.72 & 3.39 & 3.86 & 2.51 \\
\cdashline{2-11}
 & SMART (Gemma 3 4B) & 0.67 & 0.50 & 2.50 & 2.40 & 2.31 & 3.24 & 2.85 & 2.53 & 1.72 \\
 & SMART (DeepSeek-V3.2) & 0.52 & 0.41 & \secondcell{2.15} & 2.14 & 1.86 & 2.38 & 2.13 & 2.10 & 1.39 \\
 & SMART (Judge: GPT-5.5) & \bestcell{0.45} & \bestcell{0.36} & \bestcell{2.09} & \bestcell{1.88} & \bestcell{1.58} & \bestcell{1.96} & \bestcell{1.97} & \bestcell{1.81} & \bestcell{1.23} \\
\rowcolor{gray!4}  & \textbf{SMART (Claude 4.6)} & \secondcell{0.50} & \secondcell{0.38} & \secondcell{2.15} & \secondcell{2.05} & \secondcell{1.73} & \secondcell{2.08} & \secondcell{2.01} & \secondcell{1.97} & \secondcell{1.31} \\
\end{longtable}
\endgroup

\begingroup
\tiny
\setlength{\tabcolsep}{1.25pt}
\renewcommand{\arraystretch}{1}
\setlength\LTleft{0pt}
\setlength\LTright{0pt}

\begin{longtable}{llcccccccccccc}
\caption{Fine-grained backbone and judge analysis on Subtitle Arena for translation from English into 15 target locales: Linguistic Conventions, Technical, Locale Conventions, and Audience Appropriateness. All values are penalties (lower is better). Best and second-best values within each language direction and metric are highlighted with blue-gray and warm-beige backgrounds, respectively. These results provide the full breakdown corresponding to Table~\ref{tab:backbone_judge_main}.}
\label{tab:backbone_out_en_form}\\
\toprule
\rowcolor{tablegray}
\textbf{Dir.} &
\textbf{Method / Setting} &
\multicolumn{4}{c}{\textbf{Linguistic Conventions}} &
\multicolumn{3}{c}{\textbf{Technical}} &
\multicolumn{2}{c}{\textbf{Locale Conventions}} &
\multicolumn{2}{c}{\textbf{Audience Appropriateness}} &
\textbf{Overall} \\
\rowcolor{tablegray}
& &
MisPunc & MisCap & Gram & Space &
LineBrk & CharLim & LineLim &
LocErr & LangDet &
Profan & Formal &
\\
\midrule
\endfirsthead

\multicolumn{14}{c}{\tablename~\thetable\ (continued)}\\
\toprule
\rowcolor{tablegray}
\textbf{Dir.} &
\textbf{Method / Setting} &
\multicolumn{4}{c}{\textbf{Linguistic Conventions}} &
\multicolumn{3}{c}{\textbf{Technical}} &
\multicolumn{2}{c}{\textbf{Locale Conventions}} &
\multicolumn{2}{c}{\textbf{Audience Appropriateness}} &
\textbf{Overall} \\
\rowcolor{tablegray}
& &
MisPunc & MisCap & Gram & Space &
LineBrk & CharLim & LineLim &
LocErr & LangDet &
Profan & Formal &
\\
\midrule
\endhead

\midrule
\multicolumn{14}{r}{\scriptsize Continued on next page}\\
\endfoot

\bottomrule
\multicolumn{14}{p{\textwidth}}{\scriptsize
\textbf{Abbreviations.}
\textbf{Linguistic Conventions} (MisPunc = Mispunctuation; MisCap = Miscapitalization; Gram = Grammar; Space = Spacing Error).
\textbf{Technical} (LineBrk = Incorrect Line Breaking; CharLim = Exceeding Characters per Line; LineLim = Exceeding Lines per Box).
\textbf{Locale Conventions} (LocErr = Localization Error; LangDet = Language Detection Error).
\textbf{Audience Appropriateness} (Profan = Profanity; Formal = Formality Error).
} \\
\endlastfoot
\multirow{6}{*}{\centering en$\rightarrow$zh} & Gemma 3 4B & 1.05 & \bestcell{0.00} & 0.58 & 0.72 & 1.18 & 0.84 & 0.35 & 0.18 & 0.12 & 0.38 & 0.31 & 1.46 \\
 & DeepSeek-V3.2 & 0.61 & \bestcell{0.00} & 0.19 & 0.34 & 0.82 & 0.52 & 0.14 & \bestcell{0.04} & \bestcell{0.01} & \bestcell{0.15} & \secondcell{0.10} & 0.93 \\
\cdashline{2-14}
 & SMART (Gemma 3 4B) & 0.15 & 0.03 & 0.08 & 0.17 & 0.03 & 0.03 & 0.03 & 0.08 & 0.06 & 0.24 & 0.13 & 0.67 \\
 & SMART (DeepSeek-V3.2) & 0.11 & \secondcell{0.01} & \secondcell{0.06} & 0.15 & \secondcell{0.01} & \secondcell{0.01} & \secondcell{0.01} & 0.06 & 0.04 & 0.17 & \secondcell{0.10} & 0.53 \\
 & SMART (Judge: GPT-5.5) & \bestcell{0.09} & \secondcell{0.01} & \bestcell{0.05} & \bestcell{0.12} & \bestcell{0.00} & \bestcell{0.00} & \secondcell{0.01} & \secondcell{0.05} & \secondcell{0.03} & \secondcell{0.16} & \bestcell{0.08} & \bestcell{0.47} \\
\rowcolor{gray!4}  & \textbf{SMART (Claude 4.6)} & \secondcell{0.10} & \bestcell{0.00} & \bestcell{0.05} & \secondcell{0.13} & \bestcell{0.00} & \bestcell{0.00} & \bestcell{0.00} & \secondcell{0.05} & \secondcell{0.03} & \secondcell{0.16} & \bestcell{0.08} & \secondcell{0.48} \\
\midrule
\multirow{6}{*}{\centering en$\rightarrow$de} & Gemma 3 4B & 1.24 & 0.65 & 0.86 & 0.68 & 2.45 & 1.98 & \secondcell{0.88} & 0.68 & 0.28 & 0.64 & 0.52 & 2.21 \\
 & DeepSeek-V3.2 & 0.70 & 0.27 & 0.41 & 0.26 & 2.02 & \secondcell{1.61} & \bestcell{0.52} & 0.38 & 0.05 & \bestcell{0.36} & \bestcell{0.25} & 1.50 \\
\cdashline{2-14}
 & SMART (Gemma 3 4B) & 0.44 & 0.18 & 0.44 & 0.25 & 2.51 & 2.33 & 1.61 & 0.07 & 0.05 & 0.69 & 0.50 & 1.23 \\
 & SMART (DeepSeek-V3.2) & 0.34 & \secondcell{0.16} & \secondcell{0.32} & 0.21 & 2.23 & 1.78 & 1.32 & \secondcell{0.05} & \secondcell{0.03} & 0.51 & 0.39 & \secondcell{0.99} \\
 & SMART (Judge: GPT-5.5) & \bestcell{0.27} & \bestcell{0.14} & \bestcell{0.27} & \secondcell{0.20} & \bestcell{1.80} & \bestcell{1.59} & 1.13 & \bestcell{0.04} & \bestcell{0.02} & \secondcell{0.45} & \secondcell{0.33} & \bestcell{0.87} \\
\rowcolor{gray!4}  & \textbf{SMART (Claude 4.6)} & \secondcell{0.28} & \bestcell{0.14} & \bestcell{0.27} & \bestcell{0.19} & \secondcell{1.85} & 1.64 & 1.16 & \bestcell{0.04} & \bestcell{0.02} & 0.48 & 0.34 & \bestcell{0.87} \\
\midrule
\multirow{6}{*}{\centering en$\rightarrow$ko} & Gemma 3 4B & 1.02 & \bestcell{0.00} & 0.82 & 0.71 & 1.14 & 0.82 & 0.45 & 0.28 & 0.15 & 1.15 & 0.92 & 1.82 \\
 & DeepSeek-V3.2 & 0.55 & \bestcell{0.00} & 0.38 & 0.30 & 0.74 & 0.49 & 0.18 & \secondcell{0.06} & \bestcell{0.02} & \bestcell{0.76} & \bestcell{0.56} & 1.17 \\
\cdashline{2-14}
 & SMART (Gemma 3 4B) & 0.39 & 0.03 & 0.28 & 0.28 & 0.24 & 0.11 & 0.08 & 0.08 & 0.06 & 1.53 & 1.04 & 1.26 \\
 & SMART (DeepSeek-V3.2) & 0.28 & \secondcell{0.01} & 0.25 & 0.22 & \secondcell{0.20} & \secondcell{0.09} & \secondcell{0.06} & \secondcell{0.06} & 0.04 & 1.11 & 0.78 & 1.03 \\
 & SMART (Judge: GPT-5.5) & \secondcell{0.27} & \bestcell{0.00} & \secondcell{0.23} & \secondcell{0.21} & \bestcell{0.17} & \bestcell{0.08} & \bestcell{0.05} & \bestcell{0.05} & \secondcell{0.03} & \secondcell{0.88} & \secondcell{0.60} & \bestcell{0.91} \\
\rowcolor{gray!4}  & \textbf{SMART (Claude 4.6)} & \bestcell{0.26} & \bestcell{0.00} & \bestcell{0.22} & \bestcell{0.20} & \bestcell{0.17} & \bestcell{0.08} & \bestcell{0.05} & \bestcell{0.05} & \secondcell{0.03} & 0.93 & 0.65 & \secondcell{0.92} \\
\midrule
\multirow{6}{*}{\centering en$\rightarrow$it} & Gemma 3 4B & 1.38 & 0.78 & 1.05 & 0.82 & \secondcell{2.38} & \secondcell{1.88} & \secondcell{0.82} & 0.62 & 0.25 & 0.68 & 0.56 & 2.34 \\
 & DeepSeek-V3.2 & 0.86 & 0.39 & 0.58 & 0.36 & \bestcell{1.91} & \bestcell{1.48} & \bestcell{0.48} & 0.34 & \bestcell{0.06} & \bestcell{0.34} & \bestcell{0.23} & 1.62 \\
\cdashline{2-14}
 & SMART (Gemma 3 4B) & 0.52 & 0.24 & 0.43 & 0.39 & 3.70 & 3.82 & 2.82 & 0.27 & 0.09 & 0.58 & 0.36 & 1.56 \\
 & SMART (DeepSeek-V3.2) & 0.44 & 0.20 & 0.36 & 0.30 & 2.92 & 2.90 & 2.28 & \secondcell{0.20} & \secondcell{0.07} & 0.48 & 0.30 & \secondcell{1.30} \\
 & SMART (Judge: GPT-5.5) & \secondcell{0.40} & \bestcell{0.16} & \secondcell{0.35} & \secondcell{0.26} & 2.94 & 2.44 & 1.87 & \bestcell{0.16} & \bestcell{0.06} & 0.46 & \secondcell{0.24} & \bestcell{1.13} \\
\rowcolor{gray!4}  & \textbf{SMART (Claude 4.6)} & \bestcell{0.37} & \secondcell{0.17} & \bestcell{0.33} & \bestcell{0.25} & 2.76 & 2.44 & 1.91 & \bestcell{0.16} & \bestcell{0.06} & \secondcell{0.43} & 0.27 & \bestcell{1.13} \\
\midrule
\multirow{6}{*}{\centering en$\rightarrow$es-ES} & Gemma 3 4B & 1.42 & 0.88 & 1.12 & 0.86 & 2.32 & \secondcell{1.82} & \secondcell{0.78} & 0.58 & 0.22 & 0.71 & 0.58 & 2.36 \\
 & DeepSeek-V3.2 & 0.91 & 0.49 & 0.64 & 0.41 & \bestcell{1.85} & \bestcell{1.42} & \bestcell{0.44} & 0.26 & \secondcell{0.05} & \bestcell{0.36} & \bestcell{0.25} & 1.62 \\
\cdashline{2-14}
 & SMART (Gemma 3 4B) & 0.91 & 0.37 & 0.83 & 0.62 & 3.09 & 2.97 & 2.34 & 0.15 & 0.07 & 0.71 & 0.57 & 1.71 \\
 & SMART (DeepSeek-V3.2) & 0.74 & 0.29 & 0.61 & 0.45 & 2.66 & 2.31 & 1.77 & 0.11 & \secondcell{0.05} & 0.59 & 0.42 & 1.36 \\
 & SMART (Judge: GPT-5.5) & \bestcell{0.64} & \secondcell{0.28} & \bestcell{0.47} & \bestcell{0.37} & \secondcell{2.27} & 2.02 & 1.44 & \bestcell{0.09} & \bestcell{0.04} & \secondcell{0.51} & 0.38 & \bestcell{1.17} \\
\rowcolor{gray!4}  & \textbf{SMART (Claude 4.6)} & \secondcell{0.66} & \bestcell{0.26} & \secondcell{0.52} & \secondcell{0.40} & 2.31 & 1.99 & 1.49 & \secondcell{0.10} & \bestcell{0.04} & \secondcell{0.51} & \secondcell{0.35} & \secondcell{1.18} \\
\midrule
\multirow{6}{*}{\centering en$\rightarrow$fr} & Gemma 3 4B & 1.40 & 0.82 & 1.08 & 0.84 & \secondcell{2.48} & \secondcell{1.95} & \secondcell{0.85} & 0.48 & 0.24 & 0.74 & 0.60 & 2.31 \\
 & DeepSeek-V3.2 & 0.90 & 0.40 & 0.60 & \secondcell{0.35} & \bestcell{2.01} & \bestcell{1.55} & \bestcell{0.52} & 0.16 & \secondcell{0.03} & \bestcell{0.40} & \bestcell{0.27} & 1.60 \\
\cdashline{2-14}
 & SMART (Gemma 3 4B) & 0.80 & 0.23 & 0.63 & 0.45 & 4.77 & 3.89 & 2.76 & 0.11 & 0.05 & 0.81 & 0.52 & 1.74 \\
 & SMART (DeepSeek-V3.2) & 0.61 & 0.21 & 0.46 & \secondcell{0.35} & 3.69 & 3.02 & 2.47 & \secondcell{0.09} & \secondcell{0.03} & 0.66 & 0.42 & \secondcell{1.40} \\
 & SMART (Judge: GPT-5.5) & \bestcell{0.51} & \secondcell{0.19} & \bestcell{0.41} & \secondcell{0.35} & 2.96 & 2.60 & 2.11 & \secondcell{0.09} & \bestcell{0.02} & 0.58 & \secondcell{0.35} & \bestcell{1.19} \\
\rowcolor{gray!4}  & \textbf{SMART (Claude 4.6)} & \secondcell{0.52} & \bestcell{0.18} & \secondcell{0.42} & \bestcell{0.32} & 3.01 & 2.65 & 2.08 & \bestcell{0.08} & \bestcell{0.02} & \secondcell{0.53} & \secondcell{0.35} & \bestcell{1.19} \\
\midrule
\multirow{6}{*}{\centering en$\rightarrow$pt-PT} & Gemma 3 4B & 2.36 & 0.91 & 1.14 & 1.43 & 7.11 & 6.92 & 4.75 & 0.38 & 0.14 & 1.32 & 1.07 & 3.72 \\
 & DeepSeek-V3.2 & 1.66 & 0.68 & 0.91 & 0.96 & 5.43 & 5.24 & 3.38 & 0.25 & 0.11 & 1.01 & 0.82 & 2.73 \\
\cdashline{2-14}
 & SMART (Gemma 3 4B) & 1.11 & 0.41 & 0.69 & 0.71 & 3.54 & 3.81 & 2.25 & 0.18 & 0.07 & 1.00 & 0.55 & 1.89 \\
 & SMART (DeepSeek-V3.2) & 0.83 & 0.34 & \bestcell{0.58} & 0.54 & 2.66 & 2.88 & \bestcell{1.61} & \secondcell{0.13} & \secondcell{0.05} & 0.72 & 0.46 & 1.50 \\
 & SMART (Judge: GPT-5.5) & \bestcell{0.75} & \bestcell{0.32} & \secondcell{0.59} & \bestcell{0.43} & \bestcell{2.39} & \bestcell{2.27} & \secondcell{1.62} & \bestcell{0.11} & \bestcell{0.04} & \bestcell{0.57} & \bestcell{0.36} & \bestcell{1.30} \\
\rowcolor{gray!4}  & \textbf{SMART (Claude 4.6)} & \secondcell{0.82} & \secondcell{0.33} & \bestcell{0.58} & \secondcell{0.46} & \secondcell{2.60} & \secondcell{2.47} & \bestcell{1.61} & \secondcell{0.13} & \bestcell{0.04} & \secondcell{0.64} & \secondcell{0.41} & \secondcell{1.39} \\
\midrule
\multirow{6}{*}{\centering en$\rightarrow$pt-BR} & Gemma 3 4B & 1.86 & 0.52 & 1.26 & 0.83 & 4.47 & 5.61 & 3.70 & 0.21 & 0.11 & 0.98 & 0.80 & 2.56 \\
 & DeepSeek-V3.2 & 1.25 & 0.41 & 0.90 & 0.68 & 3.25 & 4.05 & 2.71 & 0.15 & 0.10 & 0.81 & 0.56 & 1.91 \\
\cdashline{2-14}
 & SMART (Gemma 3 4B) & 0.84 & 0.32 & 0.65 & 0.45 & 2.70 & 2.15 & 1.82 & 0.12 & 0.06 & 0.66 & 0.43 & 1.42 \\
 & SMART (DeepSeek-V3.2) & 0.61 & 0.27 & 0.56 & 0.40 & 2.27 & 1.95 & 1.32 & 0.10 & \secondcell{0.04} & 0.51 & 0.33 & 1.14 \\
 & SMART (Judge: GPT-5.5) & \bestcell{0.52} & \bestcell{0.22} & \bestcell{0.48} & \bestcell{0.36} & \bestcell{1.82} & \secondcell{1.92} & \bestcell{1.18} & \bestcell{0.07} & \bestcell{0.03} & \bestcell{0.39} & \bestcell{0.30} & \bestcell{0.97} \\
\rowcolor{gray!4}  & \textbf{SMART (Claude 4.6)} & \secondcell{0.60} & \secondcell{0.23} & \secondcell{0.50} & \secondcell{0.37} & \secondcell{2.00} & \bestcell{1.90} & \secondcell{1.26} & \secondcell{0.09} & \bestcell{0.03} & \secondcell{0.45} & \secondcell{0.31} & \secondcell{1.04} \\
\midrule
\multirow{6}{*}{\centering en$\rightarrow$no} & Gemma 3 4B & 1.86 & 0.69 & 1.53 & 0.99 & 4.99 & 4.38 & 3.28 & 0.27 & 0.12 & 1.25 & 0.89 & 2.99 \\
 & DeepSeek-V3.2 & 1.45 & 0.52 & 1.11 & 0.70 & 3.61 & 3.49 & 2.32 & 0.21 & 0.11 & 0.84 & 0.68 & 2.18 \\
\cdashline{2-14}
 & SMART (Gemma 3 4B) & 0.88 & 0.35 & 0.68 & 0.59 & 2.86 & 2.44 & 1.84 & 0.15 & 0.07 & 0.71 & 0.51 & 1.60 \\
 & SMART (DeepSeek-V3.2) & 0.73 & \secondcell{0.29} & \secondcell{0.57} & 0.46 & 2.27 & \secondcell{1.92} & 1.53 & \secondcell{0.11} & \secondcell{0.05} & 0.52 & 0.40 & 1.29 \\
 & SMART (Judge: GPT-5.5) & \bestcell{0.69} & \bestcell{0.26} & \bestcell{0.51} & \bestcell{0.37} & \bestcell{1.97} & \secondcell{1.92} & \bestcell{1.34} & \bestcell{0.10} & \bestcell{0.04} & \secondcell{0.47} & \bestcell{0.36} & \bestcell{1.10} \\
\rowcolor{gray!4}  & \textbf{SMART (Claude 4.6)} & \secondcell{0.70} & \bestcell{0.26} & \secondcell{0.57} & \secondcell{0.42} & \secondcell{2.15} & \bestcell{1.89} & \secondcell{1.36} & \bestcell{0.10} & \bestcell{0.04} & \bestcell{0.46} & \secondcell{0.37} & \secondcell{1.19} \\
\midrule
\multirow{6}{*}{\centering en$\rightarrow$da} & Gemma 3 4B & 2.29 & 0.77 & 1.47 & 0.97 & 6.96 & 4.70 & 4.01 & 0.25 & 0.12 & 1.41 & 0.93 & 3.30 \\
 & DeepSeek-V3.2 & 1.54 & 0.54 & 1.07 & 0.69 & 4.75 & 3.45 & 2.98 & 0.18 & 0.11 & 1.01 & 0.66 & 2.34 \\
\cdashline{2-14}
 & SMART (Gemma 3 4B) & 0.95 & 0.39 & 0.79 & 0.55 & 3.09 & 3.10 & 2.29 & 0.14 & 0.07 & 0.77 & 0.50 & 1.79 \\
 & SMART (DeepSeek-V3.2) & 0.81 & 0.28 & 0.59 & 0.42 & 2.50 & \secondcell{2.28} & 1.77 & \secondcell{0.10} & \secondcell{0.05} & 0.55 & 0.40 & 1.39 \\
 & SMART (Judge: GPT-5.5) & \bestcell{0.73} & \bestcell{0.26} & \bestcell{0.49} & \bestcell{0.33} & \secondcell{2.34} & \bestcell{1.99} & \secondcell{1.68} & \bestcell{0.09} & \bestcell{0.04} & \bestcell{0.45} & \bestcell{0.33} & \bestcell{1.19} \\
\rowcolor{gray!4}  & \textbf{SMART (Claude 4.6)} & \secondcell{0.74} & \secondcell{0.27} & \secondcell{0.54} & \secondcell{0.39} & \bestcell{2.32} & \bestcell{1.99} & \bestcell{1.64} & \secondcell{0.10} & \bestcell{0.04} & \secondcell{0.52} & \secondcell{0.39} & \secondcell{1.26} \\
\midrule
\multirow{6}{*}{\centering en$\rightarrow$nl} & Gemma 3 4B & 1.93 & 0.70 & 1.49 & 1.16 & 5.79 & 6.82 & 4.40 & 0.32 & 0.12 & 1.70 & 1.23 & 3.41 \\
 & DeepSeek-V3.2 & 1.40 & 0.52 & 1.04 & 0.88 & 4.71 & 5.00 & 3.22 & 0.21 & 0.11 & 1.14 & 0.88 & 2.57 \\
\cdashline{2-14}
 & SMART (Gemma 3 4B) & 0.90 & 0.35 & 0.78 & 0.54 & 3.23 & 2.77 & 1.77 & \secondcell{0.13} & 0.07 & 0.70 & 0.54 & 1.80 \\
 & SMART (DeepSeek-V3.2) & \secondcell{0.75} & \secondcell{0.31} & \secondcell{0.59} & 0.48 & \secondcell{2.43} & 2.49 & \secondcell{1.59} & \bestcell{0.11} & \secondcell{0.05} & \secondcell{0.54} & 0.47 & 1.48 \\
 & SMART (Judge: GPT-5.5) & \bestcell{0.66} & \bestcell{0.29} & \bestcell{0.55} & \bestcell{0.43} & \bestcell{2.30} & \bestcell{2.15} & \bestcell{1.44} & \bestcell{0.11} & \bestcell{0.04} & \secondcell{0.54} & \bestcell{0.38} & \bestcell{1.29} \\
\rowcolor{gray!4}  & \textbf{SMART (Claude 4.6)} & \secondcell{0.75} & \bestcell{0.29} & \bestcell{0.55} & \secondcell{0.47} & \secondcell{2.43} & \secondcell{2.33} & 1.68 & \bestcell{0.11} & \bestcell{0.04} & \bestcell{0.53} & \secondcell{0.43} & \secondcell{1.38} \\
\midrule
\multirow{6}{*}{\centering en$\rightarrow$ro} & Gemma 3 4B & 1.62 & 0.66 & 1.64 & 0.94 & 5.74 & 3.76 & 3.44 & 0.23 & 0.11 & 1.04 & 0.80 & 2.96 \\
 & DeepSeek-V3.2 & 1.17 & 0.53 & 1.16 & 0.63 & 4.43 & 3.06 & 2.48 & 0.18 & 0.10 & 0.81 & 0.60 & 2.17 \\
\cdashline{2-14}
 & SMART (Gemma 3 4B) & 0.70 & 0.36 & 0.79 & 0.58 & 2.84 & 2.50 & 1.70 & 0.13 & 0.06 & 0.65 & 0.32 & 1.45 \\
 & SMART (DeepSeek-V3.2) & \secondcell{0.61} & 0.28 & 0.58 & 0.42 & 2.32 & 1.85 & 1.35 & 0.10 & \secondcell{0.04} & 0.55 & \bestcell{0.28} & 1.14 \\
 & SMART (Judge: GPT-5.5) & \bestcell{0.57} & \secondcell{0.26} & \bestcell{0.48} & \bestcell{0.31} & \bestcell{1.96} & \secondcell{1.68} & \bestcell{1.18} & \bestcell{0.08} & \bestcell{0.03} & \secondcell{0.48} & \secondcell{0.29} & \bestcell{1.01} \\
\rowcolor{gray!4}  & \textbf{SMART (Claude 4.6)} & \bestcell{0.57} & \bestcell{0.25} & \secondcell{0.51} & \secondcell{0.36} & \secondcell{2.14} & \bestcell{1.63} & \secondcell{1.27} & \secondcell{0.09} & \bestcell{0.03} & \bestcell{0.47} & \bestcell{0.28} & \secondcell{1.07} \\
\midrule
\multirow{6}{*}{\centering en$\rightarrow$sv} & Gemma 3 4B & 1.40 & 0.44 & 1.08 & 0.87 & 6.18 & 4.54 & 3.39 & 0.21 & 0.11 & 1.00 & 0.87 & 2.57 \\
 & DeepSeek-V3.2 & 0.95 & 0.37 & 0.79 & 0.63 & 4.22 & 3.53 & 2.50 & 0.17 & 0.10 & 0.77 & 0.58 & 1.91 \\
\cdashline{2-14}
 & SMART (Gemma 3 4B) & 0.65 & 0.27 & 0.68 & 0.41 & 3.04 & 2.45 & 1.81 & 0.13 & 0.06 & 0.52 & 0.47 & 1.43 \\
 & SMART (DeepSeek-V3.2) & 0.57 & 0.22 & 0.52 & 0.36 & 2.21 & 1.86 & 1.41 & \secondcell{0.11} & \secondcell{0.04} & 0.46 & 0.37 & 1.14 \\
 & SMART (Judge: GPT-5.5) & \bestcell{0.50} & \bestcell{0.18} & \bestcell{0.42} & \bestcell{0.33} & \bestcell{1.82} & \secondcell{1.83} & \bestcell{1.35} & \bestcell{0.09} & \bestcell{0.03} & \bestcell{0.35} & \secondcell{0.33} & \bestcell{0.95} \\
\rowcolor{gray!4}  & \textbf{SMART (Claude 4.6)} & \secondcell{0.52} & \secondcell{0.20} & \secondcell{0.44} & \secondcell{0.35} & \secondcell{1.98} & \bestcell{1.76} & \secondcell{1.38} & \bestcell{0.09} & \bestcell{0.03} & \secondcell{0.41} & \bestcell{0.32} & \secondcell{1.04} \\
\midrule
\multirow{6}{*}{\centering en$\rightarrow$tr} & Gemma 3 4B & 1.63 & 0.58 & 1.34 & 0.92 & 5.51 & 5.80 & 3.73 & 0.26 & 0.11 & 1.37 & 0.79 & 3.04 \\
 & DeepSeek-V3.2 & 1.15 & 0.41 & 0.94 & 0.61 & 4.16 & 3.96 & 2.97 & 0.21 & 0.10 & 1.01 & 0.63 & 2.26 \\
\cdashline{2-14}
 & SMART (Gemma 3 4B) & 0.89 & 0.33 & 0.66 & 0.48 & 3.31 & 2.42 & 2.31 & 0.17 & 0.07 & 0.81 & 0.56 & 1.66 \\
 & SMART (DeepSeek-V3.2) & 0.74 & 0.29 & 0.52 & 0.42 & 2.51 & \secondcell{2.02} & 1.71 & \secondcell{0.13} & \secondcell{0.05} & 0.60 & 0.47 & 1.35 \\
 & SMART (Judge: GPT-5.5) & \bestcell{0.67} & \bestcell{0.22} & \bestcell{0.50} & \bestcell{0.34} & \secondcell{2.32} & \bestcell{1.87} & \secondcell{1.50} & \bestcell{0.11} & \bestcell{0.04} & \bestcell{0.45} & \bestcell{0.35} & \bestcell{1.17} \\
\rowcolor{gray!4}  & \textbf{SMART (Claude 4.6)} & \secondcell{0.71} & \secondcell{0.25} & \secondcell{0.51} & \secondcell{0.38} & \bestcell{2.27} & 2.04 & \bestcell{1.47} & \bestcell{0.11} & \bestcell{0.04} & \secondcell{0.52} & \secondcell{0.40} & \secondcell{1.25} \\
\midrule
\multirow{6}{*}{\centering en$\rightarrow$es-MX} & Gemma 3 4B & 1.68 & 0.74 & 1.37 & 1.22 & 6.58 & 5.71 & 3.91 & 0.28 & 0.14 & 1.07 & 0.79 & 3.43 \\
 & DeepSeek-V3.2 & 1.31 & 0.50 & 1.03 & 0.87 & 5.19 & 4.27 & 2.92 & 0.20 & 0.11 & 0.88 & 0.64 & 2.51 \\
\cdashline{2-14}
 & SMART (Gemma 3 4B) & 0.98 & 0.39 & 0.71 & 0.61 & 3.45 & 3.18 & 2.40 & 0.14 & 0.08 & 0.66 & 0.49 & 1.72 \\
 & SMART (DeepSeek-V3.2) & 0.78 & 0.33 & 0.64 & 0.47 & 2.88 & 2.48 & 1.72 & \secondcell{0.10} & \secondcell{0.06} & 0.55 & 0.41 & 1.39 \\
 & SMART (Judge: GPT-5.5) & \bestcell{0.67} & \secondcell{0.30} & \bestcell{0.59} & \bestcell{0.42} & \secondcell{2.49} & \bestcell{2.06} & \bestcell{1.49} & \bestcell{0.09} & \bestcell{0.05} & \bestcell{0.50} & \bestcell{0.34} & \bestcell{1.23} \\
\rowcolor{gray!4}  & \textbf{SMART (Claude 4.6)} & \secondcell{0.76} & \bestcell{0.29} & \secondcell{0.63} & \secondcell{0.45} & \bestcell{2.47} & \secondcell{2.21} & \secondcell{1.59} & \bestcell{0.09} & \bestcell{0.05} & \secondcell{0.52} & \secondcell{0.36} & \secondcell{1.31} \\
\end{longtable}
\endgroup

\begingroup
\scriptsize
\setlength{\tabcolsep}{3.0pt}
\renewcommand{\arraystretch}{0.96}
\setlength\LTleft{0pt}
\setlength\LTright{0pt}

\begin{longtable}{llccccccccc}
\caption{Fine-grained backbone and judge analysis on Subtitle Arena for translation from 15 source locales into English: Terminology, Accuracy, and Fluency. All values are penalties (lower is better). Best and second-best values within each language direction and metric are highlighted with blue-gray and warm-beige backgrounds, respectively. These results provide the full breakdown corresponding to Table~\ref{tab:backbone_judge_main}.}
\label{tab:backbone_into_en_semantic}\\
\toprule
\rowcolor{tablegray}
\textbf{Dir.} &
\textbf{Method / Setting} &
\multicolumn{2}{c}{\textbf{Terminology}} &
\multicolumn{3}{c}{\textbf{Accuracy}} &
\multicolumn{3}{c}{\textbf{Fluency}} &
\textbf{Overall} \\
\rowcolor{tablegray}
& &
NameInc & TermInc &
MisTrans & UndTrans & OvrTrans &
Coher & Natur & Vivid &
\\
\midrule
\endfirsthead

\multicolumn{11}{c}{\tablename~\thetable\ (continued)}\\
\toprule
\rowcolor{tablegray}
\textbf{Dir.} &
\textbf{Method / Setting} &
\multicolumn{2}{c}{\textbf{Terminology}} &
\multicolumn{3}{c}{\textbf{Accuracy}} &
\multicolumn{3}{c}{\textbf{Fluency}} &
\textbf{Overall} \\
\rowcolor{tablegray}
& &
NameInc & TermInc &
MisTrans & UndTrans & OvrTrans &
Coher & Natur & Vivid &
\\
\midrule
\endhead

\midrule
\multicolumn{11}{r}{\scriptsize Continued on next page}\\
\endfoot

\bottomrule
\multicolumn{11}{p{\textwidth}}{\scriptsize
\textbf{Abbreviations.}
\textbf{Terminology} (NameInc = Name Inconsistency; TermInc = Term Inconsistency).
\textbf{Accuracy} (MisTrans = Mistranslation; UndTrans = Undertranslation; OvrTrans = Overtranslation).
\textbf{Fluency} (Coher = Coherence; Natur = Naturalness; Vivid = Vividness).
} \\
\endlastfoot
\multirow{6}{*}{\centering zh$\rightarrow$en} & Gemma 3 4B & 0.58 & 0.45 & 3.91 & 2.89 & 2.51 & 2.91 & 2.24 & 1.34 & 1.64 \\
 & DeepSeek-V3.2 & 0.36 & 0.26 & 2.74 & 1.86 & 1.50 & 2.11 & 1.39 & 0.78 & 1.07 \\
\cdashline{2-11}
 & SMART (Gemma 3 4B) & 0.25 & 0.13 & 1.25 & 1.09 & 1.02 & 1.44 & 1.04 & 0.69 & 0.62 \\
 & SMART (DeepSeek-V3.2) & \secondcell{0.18} & 0.11 & 1.05 & 0.87 & 0.85 & 1.17 & 0.94 & 0.61 & \secondcell{0.51} \\
 & SMART (Judge: GPT-5.5) & \bestcell{0.15} & \bestcell{0.09} & \bestcell{0.90} & \bestcell{0.72} & \bestcell{0.67} & \secondcell{1.04} & \secondcell{0.84} & \secondcell{0.57} & \bestcell{0.44} \\
\rowcolor{gray!4}  & \textbf{SMART (Claude 4.6)} & \bestcell{0.15} & \secondcell{0.10} & \secondcell{0.95} & \secondcell{0.74} & \secondcell{0.71} & \bestcell{1.02} & \bestcell{0.78} & \bestcell{0.56} & \bestcell{0.44} \\
\midrule
\multirow{6}{*}{\centering de$\rightarrow$en} & Gemma 3 4B & 0.53 & 0.39 & 3.58 & 2.68 & 2.32 & 2.60 & 1.99 & 1.18 & 1.49 \\
 & DeepSeek-V3.2 & 0.35 & 0.22 & 2.50 & 1.72 & 1.41 & 1.88 & 1.26 & 0.71 & 1.02 \\
\cdashline{2-11}
 & SMART (Gemma 3 4B) & 0.19 & 0.12 & 1.33 & 0.88 & 1.08 & 1.39 & 0.99 & 0.71 & 0.60 \\
 & SMART (DeepSeek-V3.2) & 0.17 & \secondcell{0.10} & 1.06 & 0.78 & 0.80 & 1.11 & 0.77 & 0.53 & \secondcell{0.47} \\
 & SMART (Judge: GPT-5.5) & \bestcell{0.13} & \bestcell{0.09} & \secondcell{0.94} & \secondcell{0.73} & \secondcell{0.67} & \bestcell{0.89} & \bestcell{0.65} & \bestcell{0.49} & \bestcell{0.41} \\
\rowcolor{gray!4}  & \textbf{SMART (Claude 4.6)} & \secondcell{0.14} & \bestcell{0.09} & \bestcell{0.89} & \bestcell{0.68} & \bestcell{0.65} & \secondcell{0.93} & \secondcell{0.70} & \secondcell{0.50} & \bestcell{0.41} \\
\midrule
\multirow{6}{*}{\centering ko$\rightarrow$en} & Gemma 3 4B & 0.61 & 0.48 & 4.11 & 3.09 & 2.71 & 3.08 & 2.39 & 1.41 & 1.74 \\
 & DeepSeek-V3.2 & 0.38 & 0.28 & 2.88 & 1.99 & 1.61 & 2.21 & 1.49 & 0.82 & 1.13 \\
\cdashline{2-11}
 & SMART (Gemma 3 4B) & 0.28 & 0.15 & 1.31 & 1.15 & 1.03 & 1.71 & 1.13 & 0.85 & 0.68 \\
 & SMART (DeepSeek-V3.2) & 0.21 & 0.12 & 1.18 & 0.90 & 0.85 & 1.31 & 0.96 & 0.68 & \secondcell{0.55} \\
 & SMART (Judge: GPT-5.5) & \bestcell{0.16} & \secondcell{0.11} & \bestcell{0.96} & \secondcell{0.80} & \secondcell{0.77} & \secondcell{1.12} & \bestcell{0.78} & \secondcell{0.62} & \bestcell{0.47} \\
\rowcolor{gray!4}  & \textbf{SMART (Claude 4.6)} & \secondcell{0.17} & \bestcell{0.10} & \secondcell{1.01} & \bestcell{0.78} & \bestcell{0.74} & \bestcell{1.06} & \secondcell{0.81} & \bestcell{0.59} & \bestcell{0.47} \\
\midrule
\multirow{6}{*}{\centering it$\rightarrow$en} & Gemma 3 4B & 0.49 & 0.35 & 3.41 & 2.52 & 2.18 & 2.44 & 1.86 & 1.11 & 1.39 \\
 & DeepSeek-V3.2 & 0.31 & 0.20 & 2.34 & 1.59 & 1.32 & 1.76 & 1.18 & 0.67 & 0.91 \\
\cdashline{2-11}
 & SMART (Gemma 3 4B) & 0.19 & 0.10 & 1.13 & 0.96 & 0.94 & 1.40 & 0.89 & 0.60 & 0.55 \\
 & SMART (DeepSeek-V3.2) & 0.16 & \secondcell{0.08} & 1.00 & 0.73 & 0.73 & 1.04 & 0.70 & 0.54 & \secondcell{0.44} \\
 & SMART (Judge: GPT-5.5) & \bestcell{0.13} & \bestcell{0.07} & \secondcell{0.90} & \bestcell{0.59} & \bestcell{0.59} & \secondcell{0.91} & \bestcell{0.60} & \secondcell{0.49} & \bestcell{0.38} \\
\rowcolor{gray!4}  & \textbf{SMART (Claude 4.6)} & \secondcell{0.14} & \bestcell{0.07} & \bestcell{0.84} & \secondcell{0.62} & \secondcell{0.60} & \bestcell{0.86} & \secondcell{0.64} & \bestcell{0.46} & \bestcell{0.38} \\
\midrule
\multirow{6}{*}{\centering es-ES$\rightarrow$en} & Gemma 3 4B & 0.51 & 0.36 & 3.31 & 2.42 & 2.11 & 2.36 & 1.79 & 1.07 & 1.35 \\
 & DeepSeek-V3.2 & 0.32 & 0.21 & 2.28 & 1.52 & 1.26 & 1.70 & 1.12 & 0.64 & 0.87 \\
\cdashline{2-11}
 & SMART (Gemma 3 4B) & 0.23 & 0.10 & 1.18 & 0.74 & 0.98 & 1.05 & 0.87 & 0.57 & 0.52 \\
 & SMART (DeepSeek-V3.2) & \secondcell{0.17} & 0.08 & 0.90 & 0.65 & 0.71 & 0.90 & 0.68 & 0.52 & 0.41 \\
 & SMART (Judge: GPT-5.5) & \bestcell{0.14} & \bestcell{0.06} & \bestcell{0.77} & \bestcell{0.57} & \bestcell{0.54} & \bestcell{0.77} & \bestcell{0.58} & \bestcell{0.40} & \bestcell{0.34} \\
\rowcolor{gray!4}  & \textbf{SMART (Claude 4.6)} & \bestcell{0.14} & \secondcell{0.07} & \secondcell{0.80} & \secondcell{0.58} & \secondcell{0.57} & \secondcell{0.82} & \secondcell{0.61} & \secondcell{0.44} & \secondcell{0.36} \\
\midrule
\multirow{6}{*}{\centering fr$\rightarrow$en} & Gemma 3 4B & 0.52 & 0.38 & 3.51 & 2.59 & 2.24 & 2.51 & 1.92 & 1.14 & 1.44 \\
 & DeepSeek-V3.2 & 0.33 & 0.22 & 2.41 & 1.64 & 1.36 & 1.81 & 1.22 & 0.68 & 0.94 \\
\cdashline{2-11}
 & SMART (Gemma 3 4B) & 0.24 & 0.12 & 1.31 & 0.95 & 0.90 & 1.29 & 0.95 & 0.72 & 0.58 \\
 & SMART (DeepSeek-V3.2) & \secondcell{0.19} & \secondcell{0.09} & 1.05 & 0.73 & 0.66 & 0.96 & 0.78 & 0.57 & \secondcell{0.45} \\
 & SMART (Judge: GPT-5.5) & \bestcell{0.15} & \bestcell{0.08} & \secondcell{0.90} & \secondcell{0.65} & \secondcell{0.65} & \bestcell{0.84} & \secondcell{0.71} & \bestcell{0.47} & \bestcell{0.39} \\
\rowcolor{gray!4}  & \textbf{SMART (Claude 4.6)} & \bestcell{0.15} & \bestcell{0.08} & \bestcell{0.86} & \bestcell{0.64} & \bestcell{0.62} & \secondcell{0.89} & \bestcell{0.67} & \secondcell{0.48} & \bestcell{0.39} \\
\midrule
\multirow{6}{*}{\centering pt-PT$\rightarrow$en} & Gemma 3 4B & 0.57 & 0.23 & 2.68 & 1.39 & 1.72 & 2.58 & 1.83 & 1.26 & 1.09 \\
 & DeepSeek-V3.2 & 0.38 & 0.18 & 2.00 & 1.09 & 1.21 & 1.95 & 1.43 & 0.95 & 0.82 \\
\cdashline{2-11}
 & SMART (Gemma 3 4B) & 0.21 & 0.15 & 1.52 & 0.84 & 0.87 & 1.65 & 1.08 & 0.86 & 0.63 \\
 & SMART (DeepSeek-V3.2) & \secondcell{0.16} & \secondcell{0.11} & 1.18 & 0.72 & 0.75 & 1.27 & 0.79 & 0.62 & 0.50 \\
 & SMART (Judge: GPT-5.5) & \bestcell{0.14} & \bestcell{0.08} & \secondcell{1.03} & \secondcell{0.63} & \secondcell{0.67} & \secondcell{1.17} & \secondcell{0.71} & \secondcell{0.53} & \secondcell{0.44} \\
\rowcolor{gray!4}  & \textbf{SMART (Claude 4.6)} & 0.18 & \bestcell{0.08} & \bestcell{0.97} & \bestcell{0.62} & \bestcell{0.61} & \bestcell{0.92} & \bestcell{0.69} & \bestcell{0.49} & \bestcell{0.41} \\
\midrule
\multirow{6}{*}{\centering pt-BR$\rightarrow$en} & Gemma 3 4B & 0.33 & 0.15 & 1.83 & 0.98 & 1.19 & 1.73 & 1.26 & 0.82 & 0.75 \\
 & DeepSeek-V3.2 & 0.25 & 0.12 & 1.33 & 0.72 & \secondcell{0.81} & 1.33 & 0.90 & 0.68 & 0.55 \\
\cdashline{2-11}
 & SMART (Gemma 3 4B) & 0.19 & 0.14 & 1.52 & 1.12 & 1.02 & 1.39 & 1.13 & 0.91 & 0.66 \\
 & SMART (DeepSeek-V3.2) & 0.17 & 0.10 & 1.13 & 0.86 & 0.83 & 1.09 & 0.91 & 0.68 & 0.51 \\
 & SMART (Judge: GPT-5.5) & \secondcell{0.15} & \secondcell{0.09} & \secondcell{0.91} & \secondcell{0.70} & \secondcell{0.81} & \secondcell{0.86} & \secondcell{0.80} & \secondcell{0.63} & \secondcell{0.44} \\
\rowcolor{gray!4}  & \textbf{SMART (Claude 4.6)} & \bestcell{0.13} & \bestcell{0.06} & \bestcell{0.71} & \bestcell{0.44} & \bestcell{0.49} & \bestcell{0.74} & \bestcell{0.51} & \bestcell{0.35} & \bestcell{0.31} \\
\midrule
\multirow{6}{*}{\centering no$\rightarrow$en} & Gemma 3 4B & 0.30 & 0.20 & 1.71 & 1.44 & 1.61 & 1.99 & 1.25 & 1.13 & 0.86 \\
 & DeepSeek-V3.2 & 0.24 & 0.14 & 1.20 & 1.00 & 1.11 & 1.46 & 1.04 & 0.78 & 0.62 \\
\cdashline{2-11}
 & SMART (Gemma 3 4B) & 0.17 & 0.15 & 1.25 & 1.02 & 1.10 & 1.46 & 0.89 & 0.61 & 0.60 \\
 & SMART (DeepSeek-V3.2) & \secondcell{0.15} & 0.12 & 0.94 & 0.86 & 0.85 & 1.19 & \secondcell{0.76} & 0.55 & 0.48 \\
 & SMART (Judge: GPT-5.5) & \bestcell{0.13} & \secondcell{0.11} & \secondcell{0.81} & \secondcell{0.68} & \secondcell{0.74} & \secondcell{1.02} & \secondcell{0.76} & \secondcell{0.46} & \secondcell{0.42} \\
\rowcolor{gray!4}  & \textbf{SMART (Claude 4.6)} & \bestcell{0.13} & \bestcell{0.07} & \bestcell{0.68} & \bestcell{0.50} & \bestcell{0.53} & \bestcell{0.75} & \bestcell{0.54} & \bestcell{0.39} & \bestcell{0.32} \\
\midrule
\multirow{6}{*}{\centering da$\rightarrow$en} & Gemma 3 4B & 0.29 & 0.12 & 1.68 & 1.32 & 1.10 & 1.70 & 1.15 & 1.07 & 0.76 \\
 & DeepSeek-V3.2 & 0.21 & \secondcell{0.11} & 1.16 & 1.00 & 0.91 & 1.34 & 0.90 & 0.74 & 0.57 \\
\cdashline{2-11}
 & SMART (Gemma 3 4B) & 0.22 & 0.13 & 1.56 & 1.00 & 1.11 & 1.49 & 1.06 & 0.88 & 0.66 \\
 & SMART (DeepSeek-V3.2) & 0.18 & \secondcell{0.11} & 1.21 & 0.78 & 0.95 & 1.24 & 0.91 & 0.72 & 0.54 \\
 & SMART (Judge: GPT-5.5) & \secondcell{0.16} & \secondcell{0.11} & \secondcell{1.01} & \secondcell{0.69} & \secondcell{0.77} & \secondcell{1.02} & \secondcell{0.82} & \secondcell{0.58} & \secondcell{0.46} \\
\rowcolor{gray!4}  & \textbf{SMART (Claude 4.6)} & \bestcell{0.11} & \bestcell{0.05} & \bestcell{0.64} & \bestcell{0.50} & \bestcell{0.48} & \bestcell{0.60} & \bestcell{0.46} & \bestcell{0.36} & \bestcell{0.29} \\
\midrule
\multirow{6}{*}{\centering nl$\rightarrow$en} & Gemma 3 4B & 0.38 & 0.17 & 1.66 & 1.50 & 1.15 & 1.80 & 1.36 & 1.07 & 0.81 \\
 & DeepSeek-V3.2 & 0.26 & 0.14 & 1.33 & 1.06 & \secondcell{0.85} & 1.32 & \secondcell{0.93} & 0.77 & 0.60 \\
\cdashline{2-11}
 & SMART (Gemma 3 4B) & 0.30 & 0.17 & 1.51 & 1.03 & 1.33 & 1.51 & 1.30 & 0.98 & 0.73 \\
 & SMART (DeepSeek-V3.2) & 0.22 & 0.12 & 1.28 & \secondcell{0.86} & 1.00 & 1.33 & 1.01 & 0.84 & 0.59 \\
 & SMART (Judge: GPT-5.5) & \secondcell{0.19} & \secondcell{0.11} & \secondcell{1.18} & \secondcell{0.86} & \secondcell{0.85} & \secondcell{1.08} & 0.94 & \secondcell{0.68} & \secondcell{0.52} \\
\rowcolor{gray!4}  & \textbf{SMART (Claude 4.6)} & \bestcell{0.12} & \bestcell{0.06} & \bestcell{0.70} & \bestcell{0.52} & \bestcell{0.47} & \bestcell{0.71} & \bestcell{0.52} & \bestcell{0.36} & \bestcell{0.31} \\
\midrule
\multirow{6}{*}{\centering ro$\rightarrow$en} & Gemma 3 4B & 0.36 & 0.17 & 2.11 & 1.41 & 1.22 & 2.27 & 1.22 & 1.04 & 0.87 \\
 & DeepSeek-V3.2 & 0.28 & \secondcell{0.13} & 1.65 & 1.14 & 1.01 & 1.55 & \secondcell{0.96} & 0.81 & 0.68 \\
\cdashline{2-11}
 & SMART (Gemma 3 4B) & 0.25 & 0.19 & 1.64 & 1.31 & 1.05 & 1.93 & 1.37 & 0.88 & 0.76 \\
 & SMART (DeepSeek-V3.2) & 0.20 & 0.14 & 1.48 & 1.04 & 0.92 & 1.64 & 1.17 & 0.80 & 0.65 \\
 & SMART (Judge: GPT-5.5) & \secondcell{0.17} & 0.14 & \secondcell{1.24} & \secondcell{0.84} & \secondcell{0.80} & \secondcell{1.32} & 0.99 & \secondcell{0.66} & \secondcell{0.55} \\
\rowcolor{gray!4}  & \textbf{SMART (Claude 4.6)} & \bestcell{0.14} & \bestcell{0.07} & \bestcell{0.80} & \bestcell{0.58} & \bestcell{0.54} & \bestcell{0.77} & \bestcell{0.54} & \bestcell{0.39} & \bestcell{0.34} \\
\midrule
\multirow{6}{*}{\centering sv$\rightarrow$en} & Gemma 3 4B & 0.41 & 0.18 & 2.18 & 1.29 & 1.62 & 2.24 & 1.83 & 1.36 & 0.98 \\
 & DeepSeek-V3.2 & 0.33 & 0.15 & 1.80 & 0.96 & 1.31 & 1.59 & 1.38 & 1.02 & 0.76 \\
\cdashline{2-11}
 & SMART (Gemma 3 4B) & 0.17 & 0.13 & 1.30 & 0.81 & 0.85 & 1.14 & 0.81 & 0.61 & 0.53 \\
 & SMART (DeepSeek-V3.2) & \secondcell{0.15} & 0.11 & 1.05 & 0.71 & 0.69 & \secondcell{0.92} & \secondcell{0.65} & 0.53 & 0.44 \\
 & SMART (Judge: GPT-5.5) & \bestcell{0.13} & \secondcell{0.09} & \bestcell{0.81} & \secondcell{0.66} & \secondcell{0.62} & \bestcell{0.88} & \bestcell{0.55} & \bestcell{0.44} & \bestcell{0.38} \\
\rowcolor{gray!4}  & \textbf{SMART (Claude 4.6)} & 0.17 & \bestcell{0.08} & \secondcell{0.92} & \bestcell{0.59} & \bestcell{0.61} & \bestcell{0.88} & 0.66 & \secondcell{0.52} & \secondcell{0.39} \\
\midrule
\multirow{6}{*}{\centering tr$\rightarrow$en} & Gemma 3 4B & 0.37 & 0.20 & 2.17 & 1.76 & 1.70 & 2.03 & 1.60 & 1.52 & 1.02 \\
 & DeepSeek-V3.2 & 0.29 & 0.14 & 1.52 & 1.22 & 1.20 & 1.63 & 1.21 & 1.10 & 0.74 \\
\cdashline{2-11}
 & SMART (Gemma 3 4B) & 0.19 & 0.15 & 1.60 & 1.19 & 1.05 & 1.68 & 1.18 & 0.80 & 0.70 \\
 & SMART (DeepSeek-V3.2) & \secondcell{0.17} & 0.12 & 1.21 & 0.94 & \secondcell{0.86} & 1.22 & 1.06 & \secondcell{0.60} & 0.55 \\
 & SMART (Judge: GPT-5.5) & \bestcell{0.14} & \secondcell{0.10} & \secondcell{1.01} & \secondcell{0.89} & 0.87 & \secondcell{1.05} & \secondcell{0.90} & \secondcell{0.60} & \secondcell{0.49} \\
\rowcolor{gray!4}  & \textbf{SMART (Claude 4.6)} & \bestcell{0.14} & \bestcell{0.07} & \bestcell{0.83} & \bestcell{0.66} & \bestcell{0.58} & \bestcell{0.86} & \bestcell{0.64} & \bestcell{0.51} & \bestcell{0.38} \\
\midrule
\multirow{6}{*}{\centering es-MX$\rightarrow$en} & Gemma 3 4B & 0.35 & 0.18 & 2.44 & 1.45 & 1.22 & 1.61 & 1.46 & 1.22 & 0.90 \\
 & DeepSeek-V3.2 & 0.27 & 0.14 & 1.85 & 1.01 & 0.96 & 1.32 & 1.01 & 0.85 & 0.68 \\
\cdashline{2-11}
 & SMART (Gemma 3 4B) & 0.24 & 0.13 & 1.42 & 1.17 & 1.14 & 1.44 & 1.03 & 0.83 & 0.67 \\
 & SMART (DeepSeek-V3.2) & 0.18 & 0.11 & 1.14 & 0.90 & 0.85 & 1.29 & 0.87 & 0.64 & 0.53 \\
 & SMART (Judge: GPT-5.5) & \secondcell{0.15} & \secondcell{0.10} & \secondcell{1.05} & \secondcell{0.70} & \secondcell{0.77} & \secondcell{1.02} & \secondcell{0.72} & \secondcell{0.58} & \secondcell{0.45} \\
\rowcolor{gray!4}  & \textbf{SMART (Claude 4.6)} & \bestcell{0.13} & \bestcell{0.07} & \bestcell{0.88} & \bestcell{0.60} & \bestcell{0.55} & \bestcell{0.79} & \bestcell{0.61} & \bestcell{0.43} & \bestcell{0.36} \\
\end{longtable}
\endgroup

\begingroup
\tiny
\setlength{\tabcolsep}{1.25pt}
\renewcommand{\arraystretch}{1}
\setlength\LTleft{0pt}
\setlength\LTright{0pt}

\begin{longtable}{llcccccccccccc}
\caption{Fine-grained backbone and judge analysis on Subtitle Arena for translation from 15 source locales into English: Linguistic Conventions, Technical, Locale Conventions, and Audience Appropriateness. All values are penalties (lower is better). Best and second-best values within each language direction and metric are highlighted with blue-gray and warm-beige backgrounds, respectively. These results provide the full breakdown corresponding to Table~\ref{tab:backbone_judge_main}.}
\label{tab:backbone_into_en_form}\\
\toprule
\rowcolor{tablegray}
\textbf{Dir.} &
\textbf{Method / Setting} &
\multicolumn{4}{c}{\textbf{Linguistic Conventions}} &
\multicolumn{3}{c}{\textbf{Technical}} &
\multicolumn{2}{c}{\textbf{Locale Conventions}} &
\multicolumn{2}{c}{\textbf{Audience Appropriateness}} &
\textbf{Overall} \\
\rowcolor{tablegray}
& &
MisPunc & MisCap & Gram & Space &
LineBrk & CharLim & LineLim &
LocErr & LangDet &
Profan & Formal &
\\
\midrule
\endfirsthead

\multicolumn{14}{c}{\tablename~\thetable\ (continued)}\\
\toprule
\rowcolor{tablegray}
\textbf{Dir.} &
\textbf{Method / Setting} &
\multicolumn{4}{c}{\textbf{Linguistic Conventions}} &
\multicolumn{3}{c}{\textbf{Technical}} &
\multicolumn{2}{c}{\textbf{Locale Conventions}} &
\multicolumn{2}{c}{\textbf{Audience Appropriateness}} &
\textbf{Overall} \\
\rowcolor{tablegray}
& &
MisPunc & MisCap & Gram & Space &
LineBrk & CharLim & LineLim &
LocErr & LangDet &
Profan & Formal &
\\
\midrule
\endhead

\midrule
\multicolumn{14}{r}{\scriptsize Continued on next page}\\
\endfoot

\bottomrule
\multicolumn{14}{p{\textwidth}}{\scriptsize
\textbf{Abbreviations.}
\textbf{Linguistic Conventions} (MisPunc = Mispunctuation; MisCap = Miscapitalization; Gram = Grammar; Space = Spacing Error).
\textbf{Technical} (LineBrk = Incorrect Line Breaking; CharLim = Exceeding Characters per Line; LineLim = Exceeding Lines per Box).
\textbf{Locale Conventions} (LocErr = Localization Error; LangDet = Language Detection Error).
\textbf{Audience Appropriateness} (Profan = Profanity; Formal = Formality Error).
} \\
\endlastfoot
\multirow{6}{*}{\centering zh$\rightarrow$en} & Gemma 3 4B & 1.21 & 0.56 & 0.88 & 0.59 & 1.51 & 1.04 & 0.44 & 0.26 & 0.05 & 0.46 & 0.31 & 1.64 \\
 & DeepSeek-V3.2 & 0.64 & 0.32 & 0.46 & 0.24 & 1.41 & 0.99 & 0.24 & 0.15 & \bestcell{0.01} & 0.25 & 0.10 & 1.07 \\
\cdashline{2-14}
 & SMART (Gemma 3 4B) & 0.29 & 0.05 & 0.21 & 0.07 & 0.21 & 0.05 & 0.08 & 0.04 & 0.06 & 0.18 & 0.06 & 0.62 \\
 & SMART (DeepSeek-V3.2) & 0.22 & \secondcell{0.03} & 0.17 & \secondcell{0.05} & \secondcell{0.17} & \secondcell{0.03} & \secondcell{0.06} & \secondcell{0.02} & 0.04 & \secondcell{0.15} & \secondcell{0.04} & \secondcell{0.51} \\
 & SMART (Judge: GPT-5.5) & \bestcell{0.18} & \bestcell{0.02} & \secondcell{0.16} & \bestcell{0.04} & \bestcell{0.16} & \bestcell{0.02} & \bestcell{0.05} & \bestcell{0.01} & \secondcell{0.03} & \bestcell{0.14} & \bestcell{0.03} & \bestcell{0.44} \\
\rowcolor{gray!4}  & \textbf{SMART (Claude 4.6)} & \secondcell{0.19} & \bestcell{0.02} & \bestcell{0.15} & \bestcell{0.04} & \bestcell{0.16} & \bestcell{0.02} & \bestcell{0.05} & \bestcell{0.01} & \secondcell{0.03} & \bestcell{0.14} & \bestcell{0.03} & \bestcell{0.44} \\
\midrule
\multirow{6}{*}{\centering de$\rightarrow$en} & Gemma 3 4B & 1.01 & 0.46 & 0.74 & 0.46 & 1.41 & 0.92 & 0.38 & 0.32 & 0.05 & 0.39 & 0.24 & 1.49 \\
 & DeepSeek-V3.2 & 0.62 & 0.16 & 0.38 & 0.18 & 1.28 & 0.86 & 0.31 & 0.20 & 0.22 & \bestcell{0.08} & 0.96 & 1.02 \\
\cdashline{2-14}
 & SMART (Gemma 3 4B) & 0.22 & 0.03 & 0.20 & 0.06 & 0.17 & 0.04 & 0.10 & 0.10 & 0.03 & 0.15 & 0.05 & 0.60 \\
 & SMART (DeepSeek-V3.2) & \secondcell{0.20} & \secondcell{0.01} & 0.15 & \secondcell{0.04} & 0.15 & \secondcell{0.02} & \secondcell{0.08} & \secondcell{0.08} & \secondcell{0.01} & 0.13 & \secondcell{0.03} & \secondcell{0.47} \\
 & SMART (Judge: GPT-5.5) & \bestcell{0.17} & \bestcell{0.00} & \secondcell{0.14} & \bestcell{0.03} & \bestcell{0.13} & \bestcell{0.01} & \bestcell{0.07} & \secondcell{0.08} & \bestcell{0.00} & 0.13 & \bestcell{0.02} & \bestcell{0.41} \\
\rowcolor{gray!4}  & \textbf{SMART (Claude 4.6)} & \bestcell{0.17} & \bestcell{0.00} & \bestcell{0.13} & \bestcell{0.03} & \secondcell{0.14} & \bestcell{0.01} & \bestcell{0.07} & \bestcell{0.07} & \bestcell{0.00} & \secondcell{0.12} & \bestcell{0.02} & \bestcell{0.41} \\
\midrule
\multirow{6}{*}{\centering ko$\rightarrow$en} & Gemma 3 4B & 1.26 & 0.62 & 0.94 & 0.64 & 1.56 & 1.09 & 0.46 & 0.29 & 0.05 & 0.50 & 0.34 & 1.74 \\
 & DeepSeek-V3.2 & 0.78 & 0.24 & 0.49 & 0.26 & 1.44 & 1.02 & 0.36 & 0.16 & \bestcell{0.00} & 0.27 & 0.11 & 1.13 \\
\cdashline{2-14}
 & SMART (Gemma 3 4B) & 0.31 & 0.05 & 0.22 & 0.08 & 0.29 & 0.04 & 0.12 & 0.10 & 0.03 & 0.23 & 0.06 & 0.68 \\
 & SMART (DeepSeek-V3.2) & 0.24 & \secondcell{0.03} & \secondcell{0.17} & \secondcell{0.06} & 0.21 & \secondcell{0.02} & \secondcell{0.10} & \secondcell{0.08} & \secondcell{0.01} & \secondcell{0.20} & \secondcell{0.04} & \secondcell{0.55} \\
 & SMART (Judge: GPT-5.5) & \secondcell{0.22} & \bestcell{0.02} & \bestcell{0.16} & \bestcell{0.05} & \secondcell{0.20} & \bestcell{0.01} & \bestcell{0.09} & \secondcell{0.08} & \bestcell{0.00} & \bestcell{0.16} & \bestcell{0.03} & \bestcell{0.47} \\
\rowcolor{gray!4}  & \textbf{SMART (Claude 4.6)} & \bestcell{0.21} & \bestcell{0.02} & \bestcell{0.16} & \bestcell{0.05} & \bestcell{0.19} & \bestcell{0.01} & \bestcell{0.09} & \bestcell{0.07} & \bestcell{0.00} & \bestcell{0.16} & \bestcell{0.03} & \bestcell{0.47} \\
\midrule
\multirow{6}{*}{\centering it$\rightarrow$en} & Gemma 3 4B & 0.94 & 0.42 & 0.68 & 0.42 & 1.34 & 0.86 & 0.34 & 0.26 & 0.03 & 0.36 & 0.20 & 1.39 \\
 & DeepSeek-V3.2 & 0.57 & 0.14 & 0.37 & 0.14 & 1.22 & 0.80 & 0.28 & 0.18 & \bestcell{0.00} & 0.21 & 0.07 & 0.91 \\
\cdashline{2-14}
 & SMART (Gemma 3 4B) & 0.23 & 0.03 & 0.16 & 0.05 & 0.22 & 0.03 & 0.10 & 0.09 & 0.03 & 0.17 & 0.03 & 0.55 \\
 & SMART (DeepSeek-V3.2) & \secondcell{0.18} & \secondcell{0.01} & \secondcell{0.13} & \secondcell{0.03} & \secondcell{0.18} & \secondcell{0.01} & \secondcell{0.08} & \secondcell{0.07} & \secondcell{0.01} & 0.14 & \secondcell{0.01} & \secondcell{0.44} \\
 & SMART (Judge: GPT-5.5) & \bestcell{0.16} & \bestcell{0.00} & \bestcell{0.12} & \bestcell{0.02} & \bestcell{0.15} & \bestcell{0.00} & \bestcell{0.07} & \bestcell{0.06} & \bestcell{0.00} & \bestcell{0.11} & \bestcell{0.00} & \bestcell{0.38} \\
\rowcolor{gray!4}  & \textbf{SMART (Claude 4.6)} & \bestcell{0.16} & \bestcell{0.00} & \bestcell{0.12} & \bestcell{0.02} & \bestcell{0.15} & \bestcell{0.00} & \bestcell{0.07} & \bestcell{0.06} & \bestcell{0.00} & \secondcell{0.12} & \bestcell{0.00} & \bestcell{0.38} \\
\midrule
\multirow{6}{*}{\centering es-ES$\rightarrow$en} & Gemma 3 4B & 0.91 & 0.39 & 0.64 & 0.39 & 1.31 & 0.84 & 0.32 & 0.24 & 0.02 & 0.34 & 0.18 & 1.35 \\
 & DeepSeek-V3.2 & 0.54 & 0.12 & 0.34 & 0.12 & 1.20 & 0.78 & 0.27 & 0.16 & \bestcell{0.00} & 0.20 & 0.06 & 0.87 \\
\cdashline{2-14}
 & SMART (Gemma 3 4B) & 0.18 & 0.03 & 0.18 & 0.04 & 0.23 & 0.03 & 0.09 & 0.09 & 0.03 & 0.16 & 0.03 & 0.52 \\
 & SMART (DeepSeek-V3.2) & 0.16 & \secondcell{0.01} & \secondcell{0.13} & \secondcell{0.02} & 0.17 & \secondcell{0.01} & \secondcell{0.07} & \secondcell{0.07} & \secondcell{0.01} & 0.13 & \secondcell{0.01} & 0.41 \\
 & SMART (Judge: GPT-5.5) & \bestcell{0.14} & \secondcell{0.01} & \bestcell{0.11} & \bestcell{0.01} & \bestcell{0.13} & \bestcell{0.00} & \bestcell{0.06} & \bestcell{0.06} & \bestcell{0.00} & \bestcell{0.10} & \secondcell{0.01} & \bestcell{0.34} \\
\rowcolor{gray!4}  & \textbf{SMART (Claude 4.6)} & \secondcell{0.15} & \bestcell{0.00} & \bestcell{0.11} & \bestcell{0.01} & \secondcell{0.14} & \bestcell{0.00} & \bestcell{0.06} & \bestcell{0.06} & \bestcell{0.00} & \secondcell{0.11} & \bestcell{0.00} & \secondcell{0.36} \\
\midrule
\multirow{6}{*}{\centering fr$\rightarrow$en} & Gemma 3 4B & 0.96 & 0.44 & 0.71 & 0.44 & 1.36 & 0.89 & 0.36 & 0.28 & 0.04 & 0.38 & 0.22 & 1.44 \\
 & DeepSeek-V3.2 & 0.58 & 0.14 & 0.38 & 0.14 & 1.24 & 0.82 & 0.30 & 0.19 & \bestcell{0.00} & 0.22 & 0.07 & 0.94 \\
\cdashline{2-14}
 & SMART (Gemma 3 4B) & 0.27 & 0.03 & 0.16 & 0.04 & 0.20 & 0.04 & 0.11 & 0.10 & 0.03 & 0.20 & 0.03 & 0.58 \\
 & SMART (DeepSeek-V3.2) & 0.20 & \secondcell{0.01} & \secondcell{0.13} & \secondcell{0.02} & \secondcell{0.16} & \secondcell{0.02} & \secondcell{0.09} & \secondcell{0.08} & \secondcell{0.01} & 0.15 & \secondcell{0.01} & \secondcell{0.45} \\
 & SMART (Judge: GPT-5.5) & \bestcell{0.16} & \bestcell{0.00} & \bestcell{0.12} & \bestcell{0.01} & \bestcell{0.15} & \bestcell{0.01} & \bestcell{0.08} & \bestcell{0.07} & \secondcell{0.01} & \bestcell{0.12} & \secondcell{0.01} & \bestcell{0.39} \\
\rowcolor{gray!4}  & \textbf{SMART (Claude 4.6)} & \secondcell{0.17} & \bestcell{0.00} & \bestcell{0.12} & \bestcell{0.01} & \bestcell{0.15} & \bestcell{0.01} & \bestcell{0.08} & \bestcell{0.07} & \bestcell{0.00} & \secondcell{0.13} & \bestcell{0.00} & \bestcell{0.39} \\
\midrule
\multirow{6}{*}{\centering pt-PT$\rightarrow$en} & Gemma 3 4B & 0.52 & 0.07 & 0.31 & 0.08 & 0.38 & 0.07 & 0.16 & 0.19 & 0.08 & 0.31 & 0.08 & 1.09 \\
 & DeepSeek-V3.2 & 0.37 & 0.06 & 0.25 & 0.07 & 0.27 & 0.06 & 0.13 & 0.13 & 0.07 & 0.24 & 0.07 & 0.82 \\
\cdashline{2-14}
 & SMART (Gemma 3 4B) & 0.33 & 0.05 & 0.21 & 0.07 & 0.25 & 0.05 & 0.08 & 0.04 & 0.06 & 0.19 & 0.06 & 0.63 \\
 & SMART (DeepSeek-V3.2) & 0.26 & 0.03 & 0.17 & 0.05 & 0.21 & 0.03 & \secondcell{0.06} & \secondcell{0.02} & 0.04 & \secondcell{0.16} & 0.04 & 0.50 \\
 & SMART (Judge: GPT-5.5) & \secondcell{0.21} & \secondcell{0.02} & \secondcell{0.16} & \secondcell{0.04} & \secondcell{0.19} & \secondcell{0.02} & \bestcell{0.05} & \bestcell{0.01} & \secondcell{0.03} & \secondcell{0.16} & \secondcell{0.03} & \secondcell{0.44} \\
\rowcolor{gray!4}  & \textbf{SMART (Claude 4.6)} & \bestcell{0.17} & \bestcell{0.00} & \bestcell{0.14} & \bestcell{0.01} & \bestcell{0.17} & \bestcell{0.00} & 0.07 & 0.07 & \bestcell{0.01} & \bestcell{0.14} & \bestcell{0.01} & \bestcell{0.41} \\
\midrule
\multirow{6}{*}{\centering pt-BR$\rightarrow$en} & Gemma 3 4B & 0.34 & 0.07 & 0.19 & 0.08 & 0.31 & 0.07 & 0.15 & 0.12 & 0.08 & 0.26 & 0.08 & 0.75 \\
 & DeepSeek-V3.2 & 0.25 & 0.06 & 0.15 & 0.07 & 0.23 & 0.06 & 0.11 & 0.11 & 0.07 & 0.20 & 0.07 & 0.55 \\
\cdashline{2-14}
 & SMART (Gemma 3 4B) & 0.21 & 0.05 & 0.17 & 0.07 & 0.25 & 0.05 & 0.08 & 0.04 & 0.06 & 0.22 & 0.06 & 0.66 \\
 & SMART (DeepSeek-V3.2) & \secondcell{0.17} & 0.03 & 0.15 & 0.05 & 0.19 & 0.03 & \secondcell{0.06} & \secondcell{0.02} & 0.04 & 0.17 & 0.04 & 0.51 \\
 & SMART (Judge: GPT-5.5) & \secondcell{0.17} & \secondcell{0.02} & \secondcell{0.13} & \secondcell{0.04} & \secondcell{0.15} & \secondcell{0.02} & \bestcell{0.05} & \bestcell{0.01} & \secondcell{0.03} & \secondcell{0.15} & \secondcell{0.03} & \secondcell{0.44} \\
\rowcolor{gray!4}  & \textbf{SMART (Claude 4.6)} & \bestcell{0.13} & \bestcell{0.00} & \bestcell{0.09} & \bestcell{0.01} & \bestcell{0.12} & \bestcell{0.00} & \bestcell{0.05} & 0.05 & \bestcell{0.01} & \bestcell{0.10} & \bestcell{0.01} & \bestcell{0.31} \\
\midrule
\multirow{6}{*}{\centering no$\rightarrow$en} & Gemma 3 4B & 0.28 & 0.08 & 0.28 & 0.08 & 0.32 & 0.08 & 0.12 & 0.12 & 0.08 & 0.25 & 0.08 & 0.86 \\
 & DeepSeek-V3.2 & 0.23 & 0.07 & 0.19 & 0.07 & 0.23 & 0.07 & 0.11 & 0.11 & 0.07 & 0.20 & 0.07 & 0.62 \\
\cdashline{2-14}
 & SMART (Gemma 3 4B) & 0.22 & 0.05 & 0.21 & 0.07 & 0.19 & 0.05 & 0.08 & 0.04 & 0.06 & 0.18 & 0.06 & 0.60 \\
 & SMART (DeepSeek-V3.2) & \secondcell{0.19} & 0.03 & 0.17 & 0.05 & 0.16 & 0.03 & \secondcell{0.06} & \secondcell{0.02} & 0.04 & 0.15 & 0.04 & 0.48 \\
 & SMART (Judge: GPT-5.5) & \secondcell{0.19} & \secondcell{0.02} & \secondcell{0.15} & \secondcell{0.04} & \secondcell{0.14} & \secondcell{0.02} & \bestcell{0.05} & \bestcell{0.01} & \secondcell{0.03} & \secondcell{0.14} & \secondcell{0.03} & \secondcell{0.42} \\
\rowcolor{gray!4}  & \textbf{SMART (Claude 4.6)} & \bestcell{0.13} & \bestcell{0.01} & \bestcell{0.09} & \bestcell{0.01} & \bestcell{0.13} & \bestcell{0.01} & \bestcell{0.05} & 0.05 & \bestcell{0.00} & \bestcell{0.10} & \bestcell{0.01} & \bestcell{0.32} \\
\midrule
\multirow{6}{*}{\centering da$\rightarrow$en} & Gemma 3 4B & 0.30 & 0.08 & 0.20 & 0.08 & 0.25 & 0.08 & 0.16 & 0.16 & 0.08 & 0.22 & 0.07 & 0.76 \\
 & DeepSeek-V3.2 & 0.22 & 0.07 & 0.16 & 0.07 & \secondcell{0.18} & 0.07 & 0.11 & 0.11 & 0.07 & 0.15 & 0.06 & 0.57 \\
\cdashline{2-14}
 & SMART (Gemma 3 4B) & 0.30 & 0.05 & 0.21 & 0.07 & 0.23 & 0.05 & 0.10 & 0.04 & 0.06 & 0.17 & 0.06 & 0.66 \\
 & SMART (DeepSeek-V3.2) & 0.23 & 0.03 & 0.17 & 0.05 & \secondcell{0.18} & 0.03 & 0.07 & \secondcell{0.02} & 0.04 & 0.15 & 0.04 & 0.54 \\
 & SMART (Judge: GPT-5.5) & \secondcell{0.18} & \secondcell{0.02} & \secondcell{0.14} & \secondcell{0.04} & \secondcell{0.18} & \secondcell{0.02} & \secondcell{0.06} & \bestcell{0.01} & \secondcell{0.03} & \secondcell{0.14} & \secondcell{0.03} & \secondcell{0.46} \\
\rowcolor{gray!4}  & \textbf{SMART (Claude 4.6)} & \bestcell{0.12} & \bestcell{0.01} & \bestcell{0.09} & \bestcell{0.01} & \bestcell{0.11} & \bestcell{0.01} & \bestcell{0.05} & 0.05 & \bestcell{0.00} & \bestcell{0.09} & \bestcell{0.00} & \bestcell{0.29} \\
\midrule
\multirow{6}{*}{\centering nl$\rightarrow$en} & Gemma 3 4B & 0.34 & 0.08 & 0.22 & 0.08 & 0.28 & 0.08 & 0.12 & 0.14 & 0.07 & 0.19 & 0.08 & 0.81 \\
 & DeepSeek-V3.2 & 0.23 & 0.07 & \secondcell{0.18} & 0.07 & 0.21 & 0.07 & 0.11 & 0.11 & 0.06 & \secondcell{0.15} & 0.07 & 0.60 \\
\cdashline{2-14}
 & SMART (Gemma 3 4B) & 0.36 & 0.06 & 0.26 & 0.07 & 0.30 & 0.05 & 0.09 & 0.04 & 0.06 & 0.23 & 0.06 & 0.73 \\
 & SMART (DeepSeek-V3.2) & 0.28 & 0.04 & 0.20 & 0.05 & 0.24 & 0.03 & 0.07 & \secondcell{0.02} & 0.04 & 0.20 & 0.04 & 0.59 \\
 & SMART (Judge: GPT-5.5) & \secondcell{0.21} & \secondcell{0.03} & \secondcell{0.18} & \secondcell{0.04} & \secondcell{0.19} & \secondcell{0.02} & \secondcell{0.06} & \bestcell{0.01} & \secondcell{0.03} & \secondcell{0.15} & \secondcell{0.03} & \secondcell{0.52} \\
\rowcolor{gray!4}  & \textbf{SMART (Claude 4.6)} & \bestcell{0.12} & \bestcell{0.01} & \bestcell{0.09} & \bestcell{0.01} & \bestcell{0.12} & \bestcell{0.01} & \bestcell{0.05} & 0.05 & \bestcell{0.00} & \bestcell{0.09} & \bestcell{0.01} & \bestcell{0.31} \\
\midrule
\multirow{6}{*}{\centering ro$\rightarrow$en} & Gemma 3 4B & 0.33 & 0.07 & 0.22 & 0.08 & 0.31 & 0.07 & 0.19 & 0.14 & 0.08 & 0.25 & 0.07 & 0.87 \\
 & DeepSeek-V3.2 & 0.23 & 0.06 & \secondcell{0.15} & 0.07 & 0.24 & 0.06 & 0.13 & 0.11 & 0.07 & \secondcell{0.17} & 0.06 & 0.68 \\
\cdashline{2-14}
 & SMART (Gemma 3 4B) & 0.33 & 0.06 & 0.25 & 0.08 & 0.27 & 0.05 & 0.09 & 0.04 & 0.07 & 0.25 & 0.07 & 0.76 \\
 & SMART (DeepSeek-V3.2) & 0.24 & 0.04 & 0.19 & 0.06 & 0.22 & 0.03 & 0.07 & \secondcell{0.02} & 0.05 & 0.22 & 0.05 & 0.65 \\
 & SMART (Judge: GPT-5.5) & \secondcell{0.21} & \secondcell{0.03} & 0.16 & \secondcell{0.05} & \secondcell{0.19} & \secondcell{0.02} & \secondcell{0.06} & \bestcell{0.01} & \secondcell{0.04} & 0.19 & \secondcell{0.04} & \secondcell{0.55} \\
\rowcolor{gray!4}  & \textbf{SMART (Claude 4.6)} & \bestcell{0.13} & \bestcell{0.00} & \bestcell{0.09} & \bestcell{0.01} & \bestcell{0.14} & \bestcell{0.00} & \bestcell{0.05} & 0.05 & \bestcell{0.01} & \bestcell{0.10} & \bestcell{0.00} & \bestcell{0.34} \\
\midrule
\multirow{6}{*}{\centering sv$\rightarrow$en} & Gemma 3 4B & 0.45 & 0.08 & 0.27 & 0.08 & 0.32 & 0.07 & 0.16 & 0.17 & 0.08 & 0.25 & 0.08 & 0.98 \\
 & DeepSeek-V3.2 & 0.31 & 0.07 & 0.21 & 0.07 & 0.25 & 0.06 & 0.12 & 0.13 & 0.07 & 0.19 & 0.07 & 0.76 \\
\cdashline{2-14}
 & SMART (Gemma 3 4B) & 0.30 & 0.05 & 0.18 & 0.07 & 0.18 & 0.05 & 0.07 & 0.04 & 0.06 & 0.16 & 0.06 & 0.53 \\
 & SMART (DeepSeek-V3.2) & 0.22 & 0.03 & \secondcell{0.14} & 0.05 & \secondcell{0.16} & 0.03 & \secondcell{0.05} & \secondcell{0.02} & 0.04 & 0.14 & 0.04 & 0.44 \\
 & SMART (Judge: GPT-5.5) & \secondcell{0.17} & \secondcell{0.02} & \bestcell{0.13} & \secondcell{0.04} & \bestcell{0.12} & \secondcell{0.02} & \bestcell{0.04} & \bestcell{0.01} & \secondcell{0.03} & \secondcell{0.13} & \secondcell{0.03} & \bestcell{0.38} \\
\rowcolor{gray!4}  & \textbf{SMART (Claude 4.6)} & \bestcell{0.16} & \bestcell{0.01} & \bestcell{0.13} & \bestcell{0.01} & \secondcell{0.16} & \bestcell{0.00} & 0.06 & 0.07 & \bestcell{0.01} & \bestcell{0.11} & \bestcell{0.01} & \secondcell{0.39} \\
\midrule
\multirow{6}{*}{\centering tr$\rightarrow$en} & Gemma 3 4B & 0.37 & 0.08 & 0.31 & 0.08 & 0.31 & 0.08 & 0.14 & 0.21 & 0.08 & 0.28 & 0.08 & 1.02 \\
 & DeepSeek-V3.2 & 0.27 & 0.07 & 0.23 & 0.07 & 0.23 & 0.07 & 0.12 & 0.15 & 0.07 & 0.19 & 0.07 & 0.74 \\
\cdashline{2-14}
 & SMART (Gemma 3 4B) & 0.30 & 0.05 & 0.24 & 0.07 & 0.26 & 0.05 & 0.08 & 0.04 & 0.06 & 0.21 & 0.06 & 0.70 \\
 & SMART (DeepSeek-V3.2) & 0.25 & 0.03 & 0.20 & 0.05 & \secondcell{0.20} & 0.03 & \secondcell{0.06} & \secondcell{0.02} & 0.04 & 0.18 & 0.04 & 0.55 \\
 & SMART (Judge: GPT-5.5) & \secondcell{0.19} & \secondcell{0.02} & \secondcell{0.16} & \secondcell{0.04} & 0.21 & \secondcell{0.02} & \bestcell{0.05} & \bestcell{0.01} & \secondcell{0.03} & \secondcell{0.14} & \secondcell{0.03} & \secondcell{0.49} \\
\rowcolor{gray!4}  & \textbf{SMART (Claude 4.6)} & \bestcell{0.15} & \bestcell{0.01} & \bestcell{0.13} & \bestcell{0.01} & \bestcell{0.14} & \bestcell{0.01} & \secondcell{0.06} & 0.07 & \bestcell{0.01} & \bestcell{0.11} & \bestcell{0.01} & \bestcell{0.38} \\
\midrule
\multirow{6}{*}{\centering es-MX$\rightarrow$en} & Gemma 3 4B & 0.33 & 0.07 & 0.24 & 0.08 & 0.35 & 0.08 & 0.14 & 0.21 & 0.08 & 0.34 & 0.08 & 0.90 \\
 & DeepSeek-V3.2 & 0.26 & 0.06 & 0.19 & 0.07 & 0.26 & 0.07 & 0.12 & 0.16 & 0.07 & 0.23 & 0.07 & 0.68 \\
\cdashline{2-14}
 & SMART (Gemma 3 4B) & 0.31 & 0.05 & 0.22 & 0.07 & 0.23 & 0.05 & 0.08 & 0.04 & 0.06 & 0.23 & 0.06 & 0.67 \\
 & SMART (DeepSeek-V3.2) & 0.23 & 0.03 & 0.17 & 0.05 & 0.17 & 0.03 & \secondcell{0.06} & \secondcell{0.02} & 0.04 & 0.18 & 0.04 & 0.53 \\
 & SMART (Judge: GPT-5.5) & \secondcell{0.21} & \secondcell{0.02} & \secondcell{0.13} & \secondcell{0.04} & \secondcell{0.15} & \secondcell{0.02} & \bestcell{0.05} & \bestcell{0.01} & \secondcell{0.03} & \secondcell{0.16} & \secondcell{0.03} & \secondcell{0.45} \\
\rowcolor{gray!4}  & \textbf{SMART (Claude 4.6)} & \bestcell{0.15} & \bestcell{0.00} & \bestcell{0.12} & \bestcell{0.01} & \bestcell{0.14} & \bestcell{0.00} & \secondcell{0.06} & 0.07 & \bestcell{0.01} & \bestcell{0.13} & \bestcell{0.00} & \bestcell{0.36} \\
\end{longtable}
\endgroup

\subsection{Temporal Generalization to Newly Released TV Series}
\label{sec:temporal-generalization}

To test whether SMART's gains extend beyond the content represented in Subtitle Arena, we construct a separate evaluation set of 200 TV series released between 2025 and 2026. This set was collected independently and is disjoint from Subtitle Arena in title. Anthropic reports a May 2025 knowledge cutoff for Claude Sonnet 4.6~\citep{anthropic2025claudesonnet46}; therefore, the collection includes titles released beyond the backbone's reported knowledge horizon. We use this evaluation to test transfer to newly released television content rather than differences in translation difficulty across historical eras.

Table~\ref{tab:temporal_generalization} reports SubMQM results on the same new-series split for Online subtitles, TransAgent, and SMART. SMART achieves a lower Overall penalty than TransAgent in all six directions. Its average Overall penalty is $0.73$, compared with $0.77$ for TransAgent, with absolute improvements of $0.04$, $0.07$, $0.02$, $0.03$, $0.02$, and $0.01$ for en$\rightarrow$zh, en$\rightarrow$de, en$\rightarrow$ko, en$\rightarrow$it, en$\rightarrow$es, and en$\rightarrow$fr, respectively.

The largest difference appears in semantic fidelity: averaged across the six directions, the Accuracy penalty decreases from $1.29$ with TransAgent to $1.16$ with SMART. Terminology and Locale Conventions remain comparable, while Fluency, Technical, and Audience Appropriateness show smaller mixed differences. These results show that the performance pattern observed on Subtitle Arena persists on a temporally shifted set containing newly released television content.

\begin{table*}[t]
\centering
\caption{
SubMQM results on 200 newly collected TV series released in 2025--2026.
All values are penalties (lower is better). The best and second-best values
for each metric within each direction are highlighted in blue-gray and
warm beige, respectively.
}
\vspace{-8pt}
\label{tab:temporal_generalization}

\scriptsize
\setlength{\tabcolsep}{3.2pt}
\renewcommand{\arraystretch}{0.98}

\resizebox{0.75\textwidth}{!}{%
\begin{tabular}{llcccccccc}
\toprule

\rowcolor{tablegray}
\textbf{Direction} &
\textbf{Method} &
\textbf{Term.} &
\textbf{Acc.} &
\textbf{Flu.} &
\textbf{Ling.} &
\textbf{Tech.} &
\textbf{Loc.} &
\textbf{Aud.} &
\textbf{Overall} \\
\midrule

\multirow{3}{*}{en$\rightarrow$zh}
& Online & 0.28 & 3.52 & 2.95 & 1.42 & 0.12 & 0.85 & \bestcell{0.08} & 1.87 \\
\cdashline{2-10}
& TransAgent & \secondcell{0.17} & \secondcell{0.88} & \secondcell{0.79} & \secondcell{0.04} & \secondcell{0.05} & \bestcell{0.04} & 0.11 & \secondcell{0.48} \\
\rowcolor{gray!4}
& \textbf{SMART} & \bestcell{0.16} & \bestcell{0.83} & \bestcell{0.71} & \bestcell{0.03} & \bestcell{0.00} & \secondcell{0.05} & \secondcell{0.10} & \bestcell{0.44} \\

\midrule

\multirow{3}{*}{en$\rightarrow$de}
& Online & 0.55 & 4.37 & \secondcell{2.63} & 0.51 & 1.45 & 0.67 & \bestcell{0.29} & 2.14 \\
\cdashline{2-10}
& TransAgent & \secondcell{0.28} & \secondcell{1.51} & \bestcell{1.46} & \secondcell{0.31} & \bestcell{1.07} & \secondcell{0.05} & \secondcell{0.40} & \secondcell{0.94} \\
\rowcolor{gray!4}
& \textbf{SMART} & \bestcell{0.27} & \bestcell{1.24} & \bestcell{1.46} & \bestcell{0.26} & \secondcell{1.27} & \bestcell{0.03} & 0.42 & \bestcell{0.87} \\

\midrule

\multirow{3}{*}{en$\rightarrow$ko}
& Online & 0.44 & 2.73 & 2.06 & 0.93 & 0.30 & \secondcell{0.15} & 0.56 & 1.48 \\
\cdashline{2-10}
& TransAgent & \secondcell{0.17} & \secondcell{0.62} & \secondcell{0.81} & \secondcell{0.13} & \secondcell{0.12} & \bestcell{0.03} & \bestcell{0.35} & \secondcell{0.44} \\
\rowcolor{gray!4}
& \textbf{SMART} & \bestcell{0.15} & \bestcell{0.59} & \bestcell{0.74} & \bestcell{0.12} & \bestcell{0.09} & \bestcell{0.03} & \secondcell{0.42} & \bestcell{0.42} \\

\midrule

\multirow{3}{*}{en$\rightarrow$it}
& Online & \secondcell{0.45} & 4.21 & 3.15 & \secondcell{2.74} & \bestcell{0.98} & 0.79 & \secondcell{0.30} & 2.33 \\
\cdashline{2-10}
& TransAgent & \bestcell{0.20} & \secondcell{1.43} & \bestcell{1.35} & \bestcell{0.16} & \secondcell{1.01} & \secondcell{0.13} & \bestcell{0.07} & \secondcell{0.81} \\
\rowcolor{gray!4}
& \textbf{SMART} & \bestcell{0.20} & \bestcell{1.23} & \secondcell{1.43} & \bestcell{0.16} & 1.05 & \bestcell{0.11} & \bestcell{0.07} & \bestcell{0.78} \\

\midrule

\multirow{3}{*}{en$\rightarrow$es}
& Online & 0.41 & 5.97 & 3.58 & 1.30 & \secondcell{0.98} & 0.46 & 0.77 & 2.86 \\
\cdashline{2-10}
& TransAgent & \bestcell{0.19} & \secondcell{1.83} & \bestcell{1.09} & \bestcell{0.49} & 0.99 & \secondcell{0.06} & \bestcell{0.19} & \secondcell{0.92} \\
\rowcolor{gray!4}
& \textbf{SMART} & \secondcell{0.20} & \bestcell{1.61} & \secondcell{1.26} & \secondcell{0.53} & \bestcell{0.94} & \bestcell{0.05} & \secondcell{0.21} & \bestcell{0.90} \\

\midrule

\multirow{3}{*}{en$\rightarrow$fr}
& Online & 0.51 & 4.21 & 2.94 & 1.82 & 0.93 & \secondcell{0.66} & 0.75 & 2.27 \\
\cdashline{2-10}
& TransAgent & \secondcell{0.13} & \secondcell{1.49} & \secondcell{2.06} & \bestcell{0.40} & \secondcell{0.71} & \bestcell{0.09} & \secondcell{0.36} & \secondcell{1.00} \\
\rowcolor{gray!4}
& \textbf{SMART} & \bestcell{0.12} & \bestcell{1.47} & \bestcell{2.05} & \secondcell{0.41} & \bestcell{0.69} & \bestcell{0.09} & \bestcell{0.34} & \bestcell{0.99} \\

\bottomrule
\end{tabular}%
}

\vspace{3pt}
\begin{minipage}{0.96\textwidth}
\scriptsize
\textbf{Abbreviations.}
Term. = Terminology; Acc. = Accuracy; Flu. = Fluency;
Ling. = Linguistic Conventions; Tech. = Technical;
Loc. = Locale Conventions; Aud. = Audience Appropriateness.
\end{minipage}

\vspace{-10pt}
\end{table*}

\subsection{Ablation Settings}
\label{app:ablation-settings}

We provide detailed definitions of the ablation variants used in
Table~\ref{tab:ablation}. Unless otherwise specified, each variant changes only
the indicated component while retaining the remaining SMART configuration,
backbone models, prompts, tools, and evaluation protocol.

\textbf{w/o Dynamic Router.}
We replace the dynamic routing policy with a fixed translation workflow.
Instead of selecting translator roles and auxiliary tools according to the
current subtitle context, all inputs follow the same predefined execution
path.

\textbf{w/o Self-Evolution.}
We disable SMART's test-time evolution stage. The routing policy and agent
prompts remain at their initial configurations throughout evaluation rather
than being updated using feedback accumulated from the series-level
test-time-training subset.

\textbf{w/o Memory.}
We remove the persistent series memory shared across subtitle sentences and
episodes. Each translation is therefore produced without access to previously
accumulated series-level information, while the remaining retrieval, routing,
and refinement components are unchanged.

\textbf{w/o Contextual Retrieval \& Idiom Bank.}
We disable the retrieval mechanisms that provide contextually relevant
information and previously accumulated idiomatic or expression-level
knowledge to the translation agents. The agents still receive their ordinary
subtitle context and can use the remaining components of SMART.

\textbf{w/o Sliding Window.}
We remove the series-level sliding-window consistency stage. Translation and
refinement are performed without the additional overlapping-window pass used
to identify and correct inconsistencies across neighboring subtitle sentences.

\textbf{w/o MoA.}
We replace the Mixture-of-Agents translator with a single translator, removing
parallel specialist hypothesis generation and subsequent candidate selection.
All other routing, memory, retrieval, and refinement components are retained.

\textbf{w/ Combined Prompt.}
To distinguish the effect of multiple independently generated hypotheses from
the effect of specialist instructions themselves, we concatenate the
instructions used by the specialist translators into a single prompt and use
one translator to produce the translation. This variant therefore exposes the
model to the same high-level translation guidance while removing independent
candidate generation and selection.

\subsection{Computational Cost and Latency}
\label{app:cost-latency}

Table~\ref{tab:cost_analysis} reports the computational overhead of SMART. An episode contains 611.7 subtitle sentences on average and requires approximately 4,970 model calls, 14.03M input tokens, 487.7K output tokens, and 3,501 tool calls. Under serial execution, the average end-to-end runtime is 25.94 minutes per episode.

These results make the quality-compute trade-off explicit. SMART achieves stronger translation quality through substantially more test-time computation than a single-call translator. At the same time, runtime remains relatively stable across the six evaluated languages, with all directions completing in approximately 22 to 27 minutes per episode. SMART is therefore most suitable for long-form translation settings where semantic accuracy, contextual consistency, and subtitle quality are more important than minimizing inference-time computation.

\begin{table*}[htbp]
\centering
\scriptsize
\setlength{\tabcolsep}{5pt}
\caption{
Average computational cost per episode across six translation directions.
Runtime is measured under serial execution.
}
\renewcommand{\arraystretch}{1.10}
\resizebox{\textwidth}{!}{%
\begin{tabular}{lcccccc}
\toprule
\textbf{Direction} &
\textbf{sentences / Ep.} &
\textbf{API Calls / Ep.} &
\textbf{Input Tokens / Ep.} &
\textbf{Output Tokens / Ep.} &
\textbf{Tool Calls / Ep.} &
\textbf{Time / Ep. (min)} \\
\midrule

en$\rightarrow$zh
& 571.3
& 4545.9
& 12.83M
& 446.1K
& 2524.7
& 21.78 \\

en$\rightarrow$es
& 597.2
& 4881.1
& 13.78M
& 479.0K
& 3826.7
& 25.60 \\

en$\rightarrow$fr
& 625.4
& 5453.9
& 15.78M
& 535.2K
& 3275.3
& 27.34 \\

en$\rightarrow$de
& 625.4
& 5312.2
& 14.99M
& 521.3K
& 4151.4
& 26.97 \\

en$\rightarrow$it
& 625.4
& 4229.7
& 11.94M
& 415.1K
& 3025.7
& 26.75 \\

en$\rightarrow$ko
& 625.4
& 5394.9
& 15.23M
& 529.5K
& 4201.9
& 27.19 \\

\midrule

Average
& 611.7
& 4969.6
& 14.03M
& 487.7K
& 3500.9
& 25.94 \\

\bottomrule
\end{tabular}%
}

\label{tab:cost_analysis}
\end{table*}



\end{CJK}

\subsection{Human Evaluation Details}
\label{app:human-evaluation-details}

To reduce annotation burden, we do not ask human annotators to reproduce the full 19-error-type SubMQM rubric. Instead, we consolidate the fine-grained criteria into four high-level dimensions that are easier to assess consistently over continuous subtitle passages. \textbf{Fidelity} measures whether the translation preserves the meaning, intent, and relevant information of the source without omission, addition, or semantic distortion. \textbf{Consistency} measures whether recurring names, terminology, character references, and discourse choices remain consistent across the passage. \textbf{Language Quality} covers fluency, naturalness, stylistic appropriateness, and linguistic conventions in the target language. \textbf{Subtitle Quality} captures viewer-facing presentation quality, including readability, conciseness, segmentation, and compatibility with subtitle display constraints.

Twenty annotators participate in the study. The evaluation contains $20$ continuous passages comprising $267$ aligned subtitle lines from $13$ television series. Each passage is presented together with its corresponding video clip and English source subtitles. For every passage, annotators compare five anonymized candidate translations produced by Online, Claude Sonnet 4.6, GPT-5.5, TransAgent, and SMART. Candidate order is independently randomized for each annotation instance to reduce position bias.

For each of the four criteria, annotators rank the five candidate translations from best to worst. The highest-ranked candidate receives a score of $5$, followed by $4$, $3$, $2$, and $1$ for the lowest-ranked candidate. Annotators additionally provide an \textbf{Overall Preference} ranking after considering the translation as a whole. We report the average ranking score across annotations and passages, such that higher values indicate stronger human preference.

\begin{figure}[t]
    \centering
    \includegraphics[width=\textwidth]{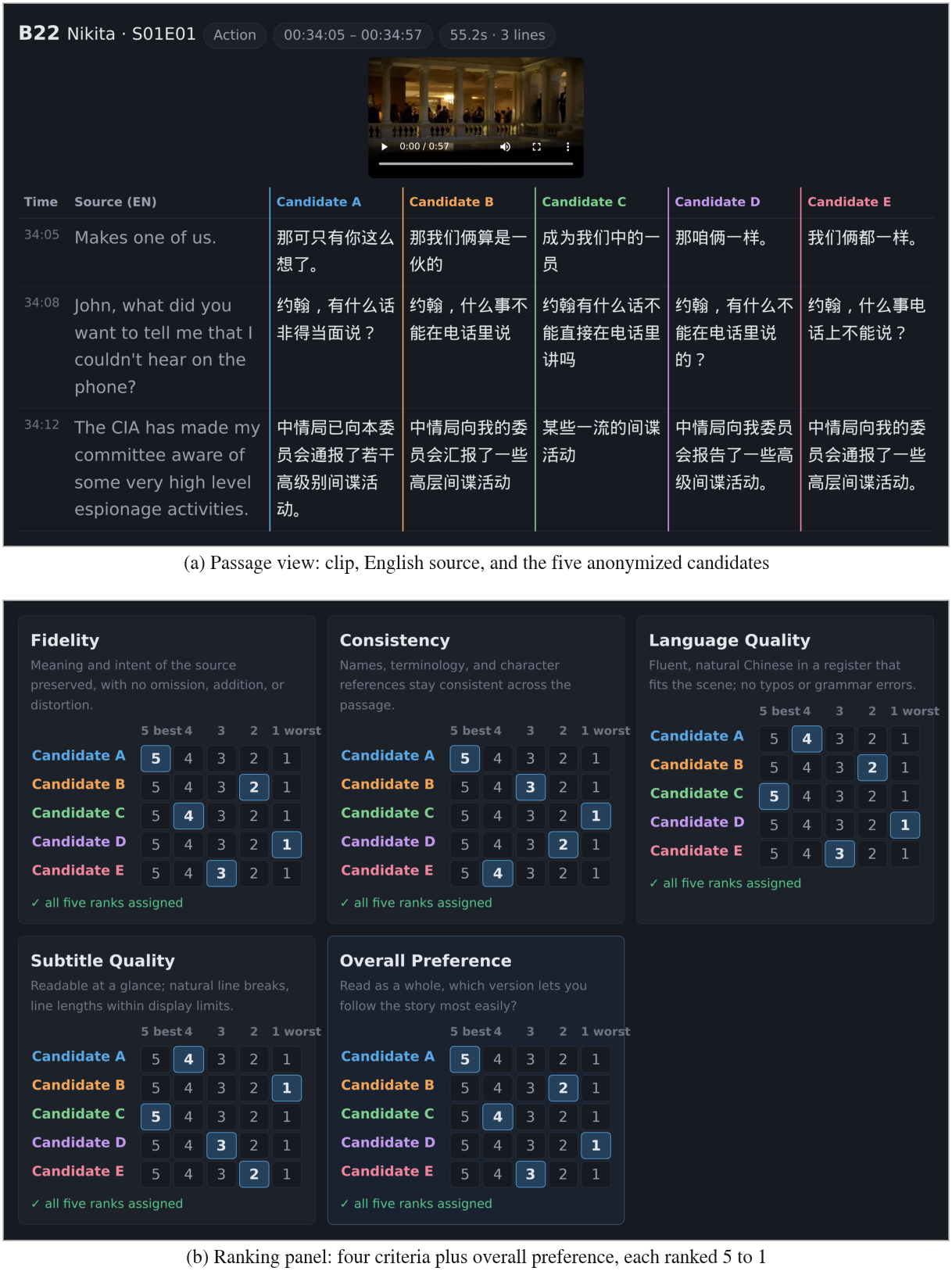}
    \caption{
    The annotation interface used in the human study.
    \textbf{(a)} Each passage is presented with its video clip, the English source lines, and the five candidate translations as anonymized columns whose order is randomized per annotation instance.
    \textbf{(b)} Annotators rank the five candidates from $5$ (best) to $1$ (worst) on Fidelity, Consistency, Language Quality, and Subtitle Quality, and then on Overall Preference; a passage can be submitted only once every panel has all five ranks assigned.
    The passage shown is from \emph{Nikita} S01E01.
    }
    \label{fig:human_eval_ui}
\end{figure}

The human study is intended as a complementary validation of viewer-facing translation quality rather than a replacement for SubMQM. Several fine-grained error categories, such as punctuation, capitalization, locale-specific formatting, language detection, and character-per-line violations, are localized or relatively sparse and are difficult to estimate reliably with a study of this scale. Human evaluation therefore focuses on broad perceptual quality, while the LLM-based SubMQM evaluation is retained for scalable and fine-grained diagnostic analysis. The annotation interface and representative evaluation examples are provided in Figure~\ref{fig:human_eval_ui}.

\section{Multimodal Expansion}
\label{app:multimodal}

SMART primarily operates on text-based SRT subtitles, but its tool-augmented design allows additional sources of contextual evidence to be incorporated without modifying the underlying translation workflow. To evaluate this extensibility, we introduce multimodal tools that expose information from the video and audio streams aligned with each subtitle segment.

Specifically, we use Qwen3-Omni~\citep{xu2025qwen3} as the multimodal backbone. Given the timestamp of a subtitle segment, the tools retrieve its corresponding video clip or audio span and return a textual description of the relevant audiovisual evidence. These tools are exposed to the MOA Translators together with the existing textual context and retrieval tools. Importantly, multimodal processing is invoked \emph{selectively}: translators may query audiovisual evidence when the subtitle alone leaves an ambiguity unresolved, rather than processing the entire video indiscriminately. Typical cases include identifying the active speaker, resolving visually grounded pronouns or references, interpreting emotional delivery, distinguishing literal from sarcastic utterances, and inferring scene-dependent expressions.

We compare four systems on six representative translation directions: text-only SMART, ViDove~\citep{lu2025vidove}, Hermes~\citep{cui2026hermes}, and our multimodal extension (\textbf{SMART + MM}). Table~\ref{tab:multimodal-expansion} reports the complete SubMQM breakdown.

\begin{table*}[t]
\centering
\caption{
Fine-grained SubMQM results for multimodal expansion on six representative translation directions.
We compare text-only SMART with ViDove, Hermes, and SMART augmented with video/audio context (\textbf{SMART + MM}).
All values are penalties (lower is better).
For clarity, only the best and second-best Overall scores within each translation direction are highlighted with blue-gray and warm-beige backgrounds, respectively.
}
\label{tab:multimodal-expansion}

\scriptsize
\setlength{\tabcolsep}{1.15pt}
\renewcommand{\arraystretch}{1.03}

\resizebox{\textwidth}{!}{%
\begin{tabular}{ll*{20}{c}}
\toprule

\rowcolor{tablegray}
& &
\multicolumn{2}{c}{\textbf{Terminology}} &
\multicolumn{3}{c}{\textbf{Accuracy}} &
\multicolumn{3}{c}{\textbf{Fluency}} &
\multicolumn{4}{c}{\textbf{Linguistic Conventions}} &
\multicolumn{3}{c}{\textbf{Technical}} &
\multicolumn{2}{c}{\textbf{Locale Conventions}} &
\multicolumn{2}{c}{\textbf{Audience Appropriateness}} &
\\

\rowcolor{tablegray}
\multirow{-2}{*}{\textbf{Dir.}} &
\multirow{-2}{*}{\textbf{Method}} &
\textbf{Name} &
\textbf{Term} &
\textbf{Mis.} &
\textbf{Under} &
\textbf{Over} &
\textbf{Coh.} &
\textbf{Nat.} &
\textbf{Viv.} &
\textbf{MisP.} &
\textbf{MisC.} &
\textbf{Gram.} &
\textbf{Space} &
\textbf{Brk.} &
\textbf{CPL} &
\textbf{Lines} &
\textbf{Loc.} &
\textbf{Lang.} &
\textbf{Prof.} &
\textbf{Form.} &
\multirow{-2}{*}{\textbf{Overall}} \\
\midrule

\multirow{4}{*}{en$\rightarrow$zh}
& SMART
& 0.18 & 0.12 & 0.86 & 0.81 & 0.73
& 1.08 & 0.94 & 0.83
& 0.10 & 0.00 & 0.05 & 0.13
& 0.00 & 0.00 & 0.00
& 0.05 & 0.03
& 0.16 & 0.08
& \secondcell{0.48} \\

& ViDove
& 0.18 & 0.14 & 0.72 & 0.87 & 0.72
& 1.36 & 1.12 & 0.62
& 0.11 & 0.00 & 0.06 & 0.15
& 0.00 & 0.00 & 0.00
& 0.05 & 0.03
& 0.17 & 0.08
& 0.49 \\

& Hermes
& 0.19 & 0.13 & 0.64 & 0.69 & 0.78
& 1.57 & 1.01 & 0.65
& 0.09 & 0.00 & 0.06 & 0.15
& 0.06 & 0.00 & 0.06
& 0.05 & 0.03
& 0.17 & 0.07
& \secondcell{0.48} \\

\rowcolor{gray!4}
& \textbf{SMART + MM}
& 0.18 & 0.12 & 0.72 & 0.70 & 0.73
& 1.08 & 0.94 & 0.59
& 0.10 & 0.00 & 0.05 & 0.13
& 0.00 & 0.00 & 0.00
& 0.03 & 0.03
& 0.16 & 0.08
& \bestcell{0.44} \\

\midrule

\multirow{4}{*}{en$\rightarrow$de}
& SMART
& 0.29 & 0.21 & 1.39 & 1.28 & 1.05
& 1.62 & 1.46 & 1.30
& 0.28 & 0.14 & 0.27 & 0.19
& 1.85 & 1.64 & 1.16
& 0.04 & 0.02
& 0.48 & 0.34
& 0.87 \\

& ViDove
& 0.33 & 0.20 & 1.19 & 1.23 & 1.07
& 2.44 & 1.64 & 1.05
& 0.29 & 0.14 & 0.34 & 0.19
& 2.05 & 1.58 & 1.33
& 0.04 & 0.02
& 0.46 & 0.38
& 0.91 \\

& Hermes
& 0.28 & 0.24 & 1.00 & 0.97 & 0.98
& 2.62 & 1.47 & 0.79
& 0.30 & 0.16 & 0.28 & 0.18
& 2.51 & 1.83 & 1.24
& 0.04 & 0.02
& 0.51 & 0.35
& \secondcell{0.85} \\

\rowcolor{gray!4}
& \textbf{SMART + MM}
& 0.29 & 0.21 & 0.97 & 0.99 & 1.05
& 1.62 & 1.46 & 0.95
& 0.28 & 0.14 & 0.27 & 0.19
& 1.85 & 1.64 & 1.16
& 0.03 & 0.02
& 0.48 & 0.34
& \bestcell{0.78} \\

\midrule

\multirow{4}{*}{en$\rightarrow$fr}
& SMART
& 0.20 & 0.10 & 1.82 & 1.58 & 1.43
& 2.41 & 2.13 & 1.97
& 0.52 & 0.18 & 0.42 & 0.32
& 3.01 & 2.65 & 2.08
& 0.08 & 0.02
& 0.53 & 0.35
& \secondcell{1.19} \\

& ViDove
& 0.20 & 0.10 & 1.37 & 1.55 & 1.62
& 3.38 & 2.61 & 1.63
& 0.58 & 0.18 & 0.50 & 0.34
& 3.35 & 2.99 & 2.11
& 0.08 & 0.02
& 0.55 & 0.39
& 1.25 \\

& Hermes
& 0.19 & 0.10 & 1.31 & 1.30 & 1.51
& 4.32 & 2.16 & 1.28
& 0.61 & 0.18 & 0.44 & 0.32
& 3.98 & 2.46 & 2.24
& 0.08 & 0.02
& 0.53 & 0.40
& 1.22 \\

\rowcolor{gray!4}
& \textbf{SMART + MM}
& 0.20 & 0.10 & 1.27 & 1.31 & 1.43
& 2.41 & 2.13 & 1.38
& 0.52 & 0.18 & 0.42 & 0.32
& 3.01 & 2.65 & 2.08
& 0.05 & 0.02
& 0.53 & 0.35
& \bestcell{1.06} \\

\midrule

\multirow{4}{*}{zh$\rightarrow$en}
& SMART
& 0.15 & 0.10 & 0.95 & 0.74 & 0.71
& 1.02 & 0.78 & 0.56
& 0.19 & 0.02 & 0.15 & 0.04
& 0.16 & 0.02 & 0.05
& 0.01 & 0.03
& 0.14 & 0.03
& \secondcell{0.44} \\

& ViDove
& 0.16 & 0.11 & 0.73 & 0.72 & 0.79
& 1.60 & 1.06 & 0.42
& 0.20 & 0.02 & 0.18 & 0.04
& 0.16 & 0.02 & 0.05
& 0.01 & 0.03
& 0.16 & 0.03
& 0.48 \\

& Hermes
& 0.16 & 0.11 & 0.67 & 0.57 & 0.70
& 1.66 & 0.90 & 0.40
& 0.21 & 0.02 & 0.15 & 0.04
& 0.19 & 0.02 & 0.08
& 0.01 & 0.03
& 0.13 & 0.03
& \secondcell{0.44} \\

\rowcolor{gray!4}
& \textbf{SMART + MM}
& 0.15 & 0.10 & 0.75 & 0.60 & 0.71
& 1.02 & 0.78 & 0.37
& 0.19 & 0.02 & 0.15 & 0.04
& 0.16 & 0.02 & 0.05
& 0.00 & 0.03
& 0.14 & 0.03
& \bestcell{0.40} \\

\midrule

\multirow{4}{*}{de$\rightarrow$en}
& SMART
& 0.14 & 0.09 & 0.89 & 0.68 & 0.65
& 0.93 & 0.70 & 0.50
& 0.17 & 0.00 & 0.13 & 0.03
& 0.14 & 0.01 & 0.07
& 0.07 & 0.00
& 0.12 & 0.02
& \secondcell{0.41} \\

& ViDove
& 0.15 & 0.09 & 0.69 & 0.75 & 0.70
& 1.38 & 0.83 & 0.39
& 0.16 & 0.00 & 0.14 & 0.03
& 0.16 & 0.01 & 0.08
& 0.08 & 0.00
& 0.13 & 0.02
& 0.43 \\

& Hermes
& 0.15 & 0.09 & 0.70 & 0.52 & 0.75
& 1.49 & 0.65 & 0.37
& 0.18 & 0.00 & 0.15 & 0.03
& 0.24 & 0.01 & 0.10
& 0.07 & 0.00
& 0.13 & 0.02
& \secondcell{0.41} \\

\rowcolor{gray!4}
& \textbf{SMART + MM}
& 0.14 & 0.09 & 0.73 & 0.57 & 0.65
& 0.93 & 0.70 & 0.33
& 0.17 & 0.00 & 0.13 & 0.03
& 0.14 & 0.01 & 0.07
& 0.05 & 0.00
& 0.12 & 0.02
& \bestcell{0.37} \\

\midrule

\multirow{4}{*}{fr$\rightarrow$en}
& SMART
& 0.15 & 0.08 & 0.86 & 0.64 & 0.62
& 0.89 & 0.67 & 0.48
& 0.17 & 0.00 & 0.12 & 0.01
& 0.15 & 0.01 & 0.08
& 0.07 & 0.00
& 0.13 & 0.00
& \secondcell{0.39} \\

& ViDove
& 0.16 & 0.08 & 0.72 & 0.71 & 0.69
& 1.17 & 0.86 & 0.37
& 0.18 & 0.00 & 0.13 & 0.01
& 0.17 & 0.01 & 0.08
& 0.07 & 0.00
& 0.14 & 0.00
& 0.42 \\

& Hermes
& 0.14 & 0.09 & 0.61 & 0.48 & 0.65
& 1.59 & 0.65 & 0.33
& 0.16 & 0.00 & 0.13 & 0.01
& 0.24 & 0.01 & 0.13
& 0.07 & 0.00
& 0.15 & 0.00
& \secondcell{0.39} \\

\rowcolor{gray!4}
& \textbf{SMART + MM}
& 0.15 & 0.08 & 0.59 & 0.47 & 0.62
& 0.89 & 0.67 & 0.26
& 0.17 & 0.00 & 0.12 & 0.01
& 0.15 & 0.01 & 0.08
& 0.04 & 0.00
& 0.13 & 0.00
& \bestcell{0.34} \\

\bottomrule
\end{tabular}%
}

\vspace{3pt}
\begin{minipage}{0.99\textwidth}
\tiny
\textbf{Abbreviations.}
\textbf{Terminology}: Name = Name Inconsistency; Term = Term Inconsistency.
\textbf{Accuracy}: Mis. = Mistranslation; Under = Undertranslation; Over = Overtranslation.
\textbf{Fluency}: Coh. = Coherence; Nat. = Naturalness; Viv. = Vividness.
\textbf{Linguistic Conventions}: MisP. = Mispunctuation; MisC. = Miscapitalization; Gram. = Grammar; Space = Spacing Error.
\textbf{Technical}: Brk. = Incorrect Line Breaking; CPL = Exceeding Characters per Line; Lines = Exceeding Lines per Box.
\textbf{Locale Conventions}: Loc. = Localization Error; Lang. = Language Detection Error.
\textbf{Audience Appropriateness}: Prof. = Profanity; Form. = Formality Error.
\end{minipage}

\end{table*}

Across all six directions, SMART + MM achieves the lowest overall SubMQM penalty. Averaged across directions, multimodal augmentation reduces the overall penalty from 0.63 for text-only SMART to 0.57, corresponding to a relative reduction of approximately 9.5\%. The largest improvements tend to occur in dimensions that depend strongly on contextual interpretation, particularly mistranslation, undertranslation, vividness, and locale-related errors. In contrast, dimensions governed largely by surface-form constraints, such as punctuation, capitalization, spacing, and subtitle line formatting, change little.

\begin{figure}[ht]
    \centering
    \includegraphics[width=\textwidth]{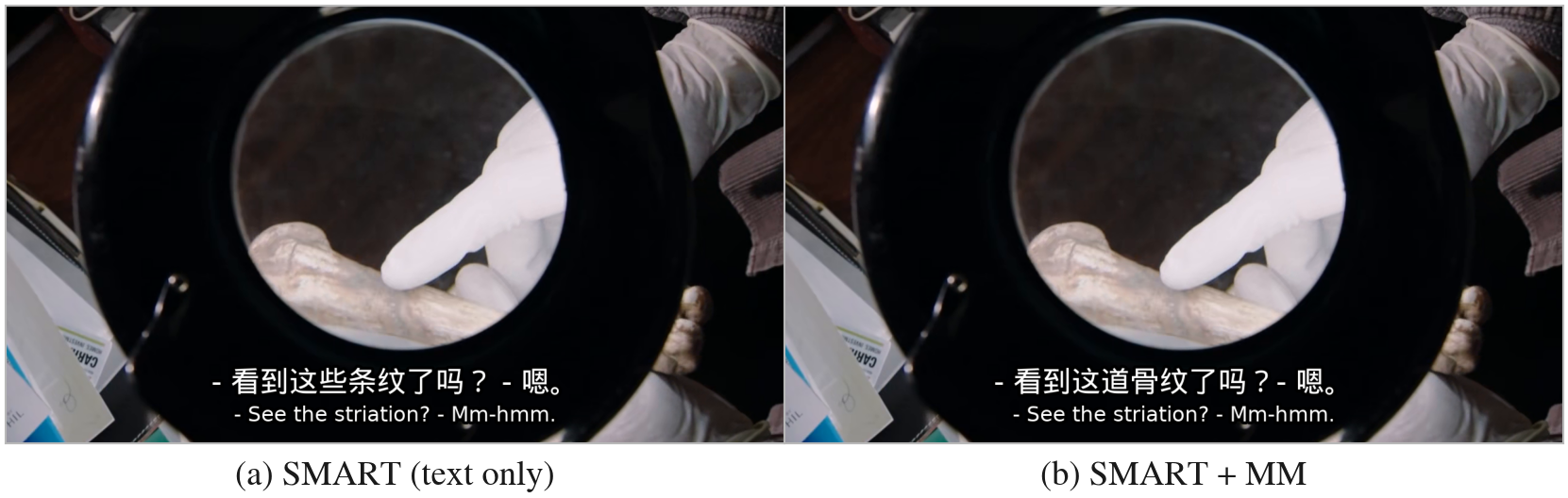}
    \caption{
    A case in which visual evidence changes the translation, on the same frame for both systems.
    The magnifier shows a single mark on a bone, which the subtitle alone does not specify.
    Text-only SMART renders \emph{striation} as a generic plural (\textbf{(a)}, \begin{CJK}{UTF8}{gkai}这些条纹\end{CJK}), while SMART + MM queries the aligned video clip, recovers that one mark on a bone is in view, and renders it as a single bone striation (\textbf{(b)}, \begin{CJK}{UTF8}{gkai}这道骨纹\end{CJK}).
    The English source is composited beneath the Chinese for reference.
    }
    \label{fig:mm_comp}
\end{figure}

Figure~\ref{fig:mm_comp} shows what the reduction looks like on a single line. In a forensic examination scene, the source line \emph{See the striation?} leaves both the number and the referent of the mark unspecified. Text-only SMART must choose from the text alone and produces a generic plural, whereas the multimodal tools report that a single mark on a bone is under the magnifier, and the translation names it accordingly. The error is one the text-only system has no evidence to avoid, and it falls in the Accuracy dimension where the table shows the largest average gain.

The results also illustrate an advantage of exposing multimodal information as \emph{tools} rather than making multimodal processing mandatory for every sentence. Most subtitle sentences can already be translated accurately from textual and long-range discourse context, while only a subset benefits materially from inspecting the underlying video or audio. SMART therefore retains its original translation workflow and invokes multimodal reasoning only when additional evidence is useful.

\section{Qualitative Analysis on Rendered Frames}
\label{app:rendered-frames}

\begin{figure}[htbp]
    \centering
    \includegraphics[width=\textwidth]{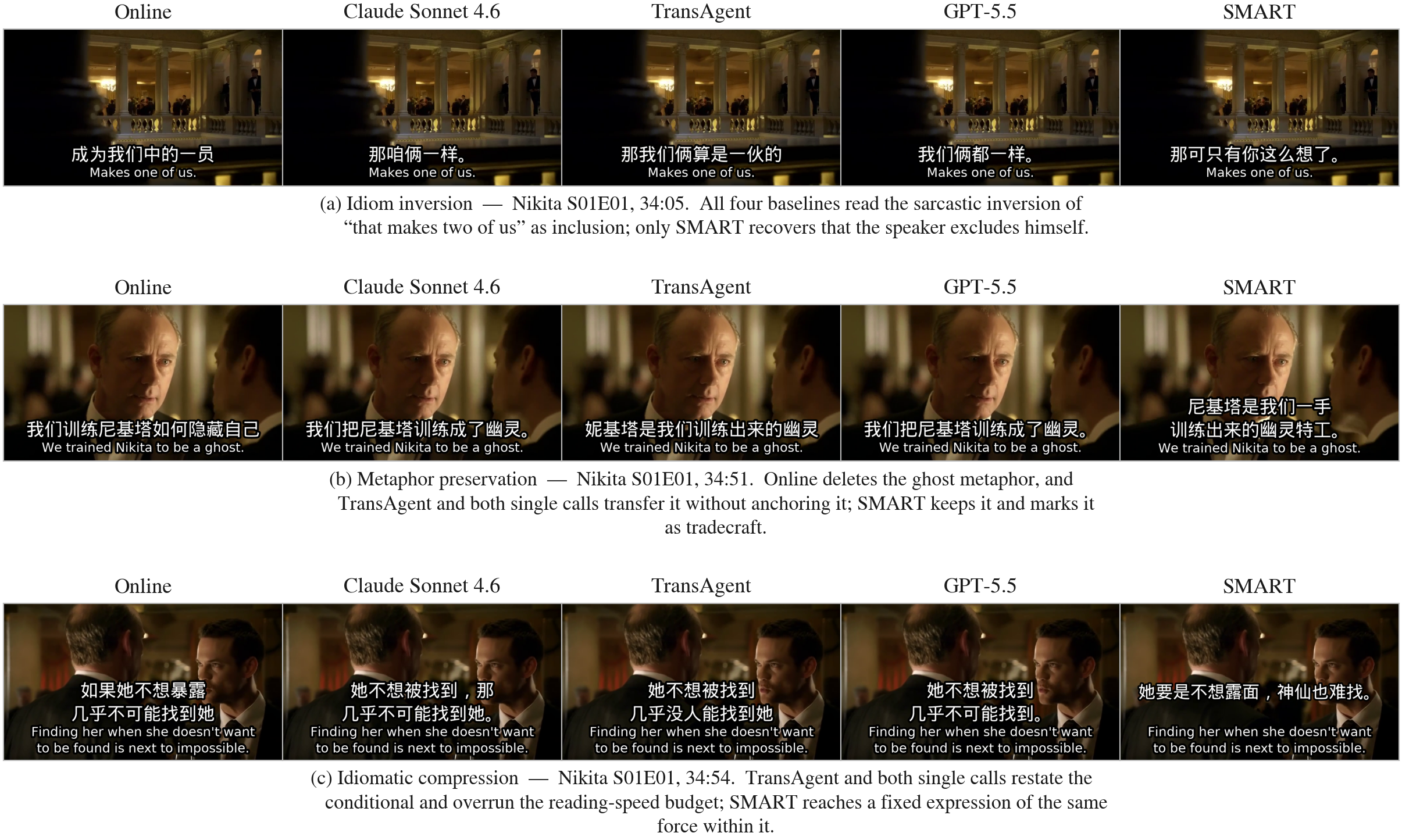}
    \caption{
    Rendered subtitle comparison on \emph{Nikita} S01E01, English$\rightarrow$Chinese, panels (a) and (b).
    Each group states the phenomenon, the timecode, and the reading the comparison turns on, so that the panels can be read without this caption.
    Two observations the panels do not state.
    In (a), \texttt{Makes one of us} is a sarcastic inversion of ``that makes two of us,'' and this is the one line in the set on which every system other than SMART agrees and is wrong.
    In (b), the TransAgent panel spells the protagonist's given name with a different character than TransAgent itself uses for that name in the other two passages of the study, whereas Online, Claude Opus 4.8, and SMART use one spelling throughout; cross-passage naming consistency is what series memory (Section~\ref{sec:series-memory}) is for, and here it is an agentic pipeline rather than a single call that fails it.
    }
    \label{fig:rendered-frames-1}
\end{figure}

\begin{figure}[ht]
    \centering
    \includegraphics[width=\textwidth]{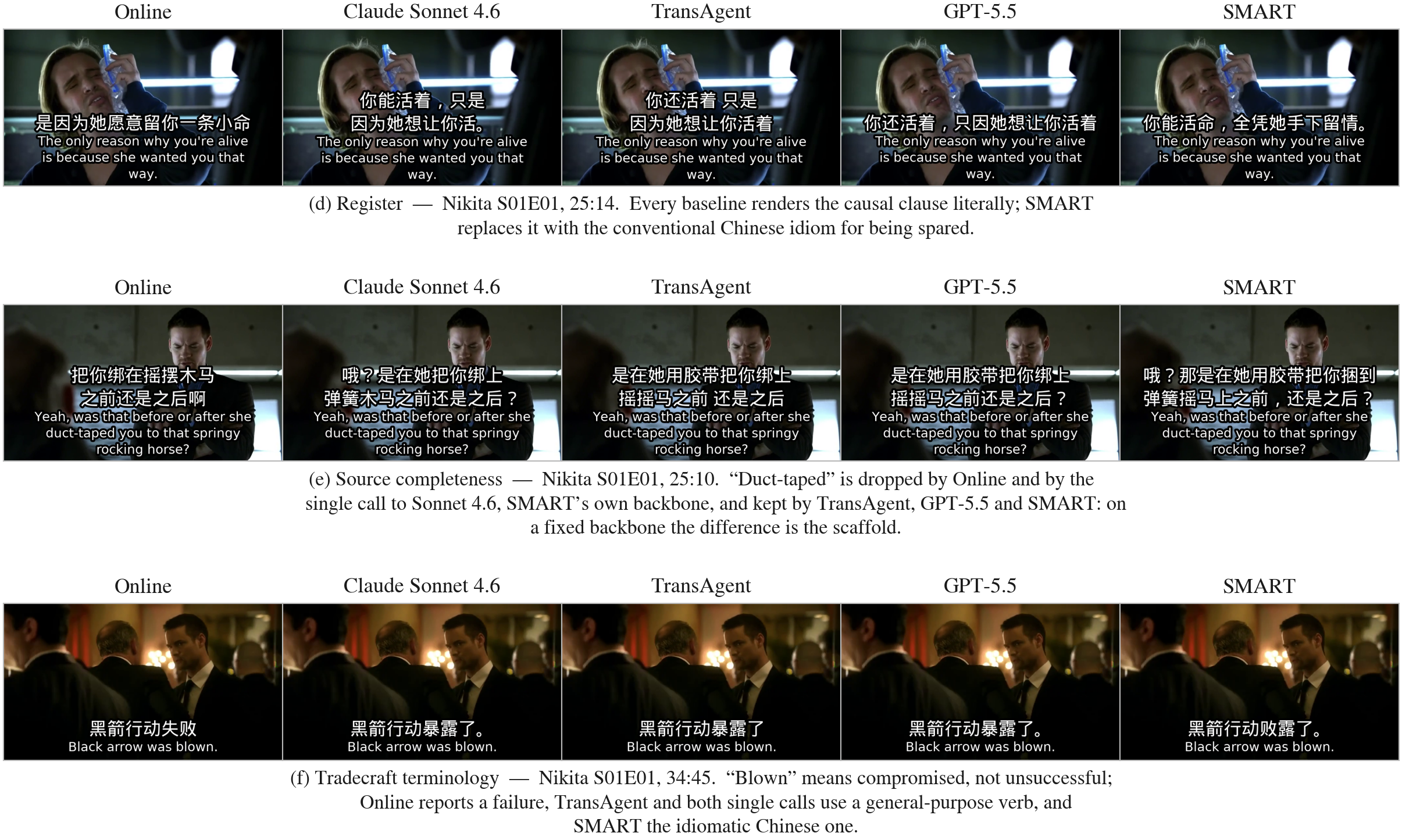}
    \caption{
    Rendered subtitle comparison on \emph{Nikita} S01E01, English$\rightarrow$Chinese, panels (c) and (d). Groups are annotated as in Figure~\ref{fig:rendered-frames-1}.
    These are the two panels in the set on which no system misreads the source. What separates them is how the Chinese is written: (c) is a line every baseline renders as a restated conditional that exceeds the reading-speed budget for its duration, and (d) a line every baseline renders as a literal causal clause.
    }
    \label{fig:rendered-frames-2}
\end{figure}

The SubMQM tables report penalties and the case studies compare passages as text, but neither shows what the differences look like where a viewer actually meets them: on the screen, for the two or three seconds the line is up. We therefore take six lines from the English$\rightarrow$Chinese passages of the human study (Appendix~\ref{app:human-evaluation-details}) and composite each system's subtitle onto the video frame at the moment the line is spoken, at the position, size, and two-line budget a player would use. The English source is burned into each panel directly beneath the Chinese, at a smaller size, so that every comparison can be read on its own. Within a row all panels are the identical frame, so the target-language subtitle is the only variable. Figures~\ref{fig:rendered-frames-1} and~\ref{fig:rendered-frames-2} show the result, and Table~\ref{tab:rendered-lines} indexes the six lines.

The lines are drawn from \emph{Nikita} S01E01, which contributes two of the study passages: one of ordinary dialogue ($00{:}25{:}04$--$00{:}25{:}40$) and one requiring external knowledge about the series ($00{:}34{:}05$--$00{:}34{:}57$). Five systems are shown: Online, single calls to Claude Sonnet 4.6 and Claude Opus 4.8, the agentic baseline TransAgent, and SMART. Claude Sonnet 4.6 is SMART's own backbone, so that column isolates the scaffold from the underlying model, and the TransAgent column separates SMART's margin over a fixed pipeline of agents from its margin over a single call. Two caveats apply. The panels are our own compositing rather than a screenshot of the annotation interface, in which candidate translations were presented as unlabeled text columns beside the clip rather than burned into the video. And six lines chosen to be legible in print illustrate the error types that the tables aggregate; they are not evidence about how often those errors occur.

\textbf{Construction.} The panels are built from the same material as the annotation interface: the clip files shown to annotators, the candidate texts exactly as they were presented, and the key that de-anonymizes the randomized columns. For each line we take the frame at the temporal midpoint of its subtitle interval, so that the panel falls inside the delivery of the line rather than on a cut. The subtitle is composited bottom-centered in white with a dark outline, wrapped to at most two lines at a break a player would take, with the English source beneath it. Glyph size is scaled inversely with the number of columns, so that the printed subtitle has the same physical size regardless of how many systems a figure shows. Five columns of legibly rendered subtitle do not fit across the page width, so both figures are set landscape. No target text is edited, repunctuated, or rewrapped by hand.

\textbf{Verification.} SMART's translation of this episode was revised after the human study was run. Every SMART line in the figures was therefore checked character-for-character against the released translation of the episode, and only lines that are identical in both are used. One candidate contrast was discarded on this basis, because a later revision changed the very word it turned on. The five columns are thus text that is fixed and citable rather than a moving target compared against frozen baselines.

\begin{table}[t]
\centering
\caption{The six lines of Figures~\ref{fig:rendered-frames-1} and~\ref{fig:rendered-frames-2}, in panel order. Timecodes are episode time in \emph{Nikita} S01E01; \emph{Passage} indicates the ordinary-dialogue (A) or external-knowledge (B) passage of the human study; \emph{Dimension} names the SubMQM dimension that the contrast bears on.}
\label{tab:rendered-lines}
\small
\setlength{\tabcolsep}{4pt}
\renewcommand{\arraystretch}{1.15}
\begin{tabular}{cccp{0.46\textwidth}l}
\toprule
\textbf{Panel} & \textbf{Timecode} & \textbf{Passage} & \textbf{English source line} & \textbf{Dimension} \\
\midrule
(a) & 34:05 & B & Makes one of us. & Accuracy \\
(b) & 34:51 & B & We trained Nikita to be a ghost. & Terminology \\
(c) & 34:54 & B & Finding her when she doesn't want to be found is next to impossible. & Fluency \\
(d) & 25:14 & A & The only reason why you're alive is because she wanted you that way. & Aud.\ Approp. \\
(e) & 25:10 & A & Yeah, was that before or after she duct-taped you to that springy rocking horse? & Accuracy \\
(f) & 34:45 & B & Black arrow was blown. & Terminology \\
\bottomrule
\end{tabular}
\end{table}

\textbf{What the six panels show.} Read together, the panels locate SMART's margin where the tables locate it. Four of them, (a), (b), (e), and (f), turn on reading the source correctly rather than on writing Chinese well, and two of those, (b) and (f), turn specifically on the material being tradecraft, which is the reading that content profiling supplies and that a single call has no reason to reach on a four-word line. The remaining two, (c) and (d), are lines that every system understands, where only SMART writes them the way a subtitle is written. The panels also show the limits of the illustration. They cover one direction, one genre, and one episode, and what they make visible are the semantic dimensions; they say nothing about the punctuation, capitalization, locale-formatting, and characters-per-line error types, which are localized and relatively sparse and for which SubMQM rather than inspection is the appropriate instrument.

\section{Limitations and Future Work}
\label{app:limitations}

\textbf{Evaluation.} SubMQM is an LLM-based protocol. We mitigate judge dependence by swapping the judge model (Appendix~\ref{app:backbone-judge-full}) and by running a human study (Appendix~\ref{app:human-evaluation-details}), but both checks cover a subset of directions, and the \emph{Online} subtitles we report alongside are community releases rather than a controlled human upper bound.

\textbf{Cost.} SMART spends substantially more test-time computation than a single-call translator (Appendix~\ref{app:cost-latency}), which limits its use in latency-bound or high-volume settings. Reducing the number of model calls without losing the consistency gains is the most direct extension.

\textbf{Coverage.} Subtitle Arena is derived from a single upstream corpus of pre-2024 subtitles and is English-pivoted, so non-English pairs and locales absent from that corpus remain untested; the temporal split of Appendix~\ref{sec:temporal-generalization} probes only the first of these. The qualitative analyses cover one direction and a small number of episodes.

\textbf{Modality.} Multimodal evidence is optional and is invoked per sentence (Appendix~\ref{app:multimodal}); speaker diarization, longer-horizon visual context, and audio prosody are not yet part of the persistent series memory.

\end{document}